\documentclass[letterpaper]{article} % DO NOT CHANGE THIS
\usepackage[submission]{aaai25}  % DO NOT CHANGE THIS
\usepackage{times}  % DO NOT CHANGE THIS
\usepackage{helvet}  % DO NOT CHANGE THIS
\usepackage{courier}  % DO NOT CHANGE THIS
\usepackage[hyphens]{url}  % DO NOT CHANGE THIS
\usepackage{graphicx} % DO NOT CHANGE THIS
\usepackage{natbib}  % DO NOT CHANGE THIS AND DO NOT ADD ANY OPTIONS TO IT
\usepackage{caption} % DO NOT CHANGE THIS AND DO NOT ADD ANY OPTIONS TO IT
\usepackage{algorithm}
\usepackage{algorithmic}

\usepackage{newfloat}
\usepackage{listings}
\DeclareCaptionStyle{ruled}{labelfont=normalfont,labelsep=colon,strut=off} % DO NOT CHANGE THIS
\floatstyle{ruled}
\newfloat{listing}{tb}{lst}{}
\floatname{listing}{Listing}
\newcommand{\yun}[1]{{\small\color{brown}{\bf\xspace#1 -yun}}}
\newcommand{\system}{\texttt{CoPreFL}\xspace}
\newcommand{\systemnott}{{CoPreFL}\xspace}

\title{Rethinking the Starting Point: \\ Collaborative Pre-Training for Federated Downstream Tasks}
\author {
    Yun-Wei Chu\textsuperscript{\rm 1},
    Dong-Jun Han\textsuperscript{\rm 2},
    Seyyedali Hosseinalipour\textsuperscript{\rm 3},
    Christopher G. Brinton\textsuperscript{\rm 1}
}
\affiliations {
     \textsuperscript{\rm 1}Purdue University, \textsuperscript{\rm 2}Yonsei University, \textsuperscript{\rm 3}University at Buffalo-SUNY\\
    chu198@purdue.edu, djh@yonsei.ac.kr,
    alipour@buffalo.edu,
    cgb@purdue.edu
}

\usepackage{bibentry}
\usepackage{enumitem}
\usepackage{amsmath}
\usepackage{xcolor}
\usepackage{multirow}
\usepackage{booktabs}
\usepackage{graphicx}
\usepackage{mwe}
\usepackage{lipsum}
\usepackage{subcaption}
\usepackage{amsfonts}

\begin{document}

\maketitle

\begin{abstract}
% A few recent studies have shown that leveraging centrally pre-trained models can offer advantageous initializations for federated learning (FL). However, existing pre-training methods do not generalize well when faced with an arbitrary set of downstream FL tasks. Specifically, they often (i) achieve limited average accuracy, particularly when there are unseen downstream labels, and (ii) result in significant accuracy variance, failing to provide a balanced performance across clients. To address these challenges, we propose {\system}, a collaborative/distributed pre-training approach which provides a robust initialization for downstream FL tasks. 
% The key idea of {\system} is a model-agnostic meta-learning (MAML) procedure that tailors the global model to closely mimic heterogeneous and unseen FL scenarios, resulting in a pre-trained model that is rapidly adaptable to arbitrary FL tasks. Our MAML procedure incorporates performance variance into the meta-objective function, balancing performance across clients rather than solely optimizing for accuracy. Through extensive experiments, we demonstrate that {\system} obtains significant improvements in both average accuracy and variance across arbitrary downstream FL tasks with unseen/seen labels, compared with various pre-training baselines. We also show how {\system} is compatible with different well-known FL algorithms applied by the downstream tasks, enhancing performance in each case.
A few recent studies have shown the benefits of using centrally pre-trained models for initializing federated learning (FL). However, existing pre-training methods do not generalize well when faced with an arbitrary set of downstream FL tasks. Specifically, they often (i) achieve limited average accuracy, particularly when there are unseen downstream labels, and (ii) result in significant accuracy variance, failing to provide a balanced performance across clients. To address these challenges, we propose {\system}, a collaborative/distributed pre-training approach which robustly initializes for downstream FL tasks. The key idea of {\system} is a model-agnostic meta-learning (MAML) procedure that tailors the global model to closely mimic heterogeneous and unseen FL scenarios, resulting in a pre-trained model that is rapidly adaptable to any FL task. Our MAML procedure integrates performance variance into the meta-objective function, balancing performance across clients rather than solely optimizing for accuracy. Extensive experiments show that {\system} significantly enhances both average accuracy and reduces variance in arbitrary downstream FL tasks with unseen/seen labels, outperforming various pre-training baselines. Additionally, {\system} proves compatible with different well-known FL algorithms applied by the downstream tasks, boosting performance in each case.

\end{abstract}

\vspace{-4mm}
\section{Introduction}
% \vspace{-1.5mm}

% Federated learning (FL) has emerged as a popular distributed machine learning paradigm, facilitating collaborative model training among sets of clients through periodic aggregations of local models by a server~\citep{McMahan2017CommunicationEfficientLO, Konecn2016FederatedLS}. 
% In recent years, significant research attention has been given to various components of the FL process, such as  aggregation schemes~\citep{Ji2019LearningPN, Wang2020FederatedLW} or local training techniques~\citep{Reddi2021AdaptiveFO, Sahu2018FederatedOI}. 
Federated learning (FL) has gained prominence as a distributed machine learning framework, enabling collaborative training among clients by periodic aggregations of local models on a server~\citep{McMahan2017CommunicationEfficientLO, Konecn2016FederatedLS}. 
Recent research has extensively explored various aspects of FL, such as aggregation schemes~\citep{Ji2019LearningPN, Wang2020FederatedLW} or local training techniques~\citep{Reddi2021AdaptiveFO, Sahu2018FederatedOI}. 
One aspect that remains understudied, however, is the impact of \textit{model initialization} in FL. 
While pre-training boosts performance in centralized AI/ML~\citep{Radford2019LanguageMA, Devlin2019BERTPO, Dosovitskiy2020AnII}, most FL works still rely on random weight initialization instead of well pre-trained models.
%Pre-training has already been well-studied in centralized AI/ML, e.g., through the practice of transfer learning in natural language processing~\citep{Radford2019LanguageMA, Devlin2019BERTPO} and computer vision~\citep{Dosovitskiy2020AnII}, where existing models are adapted to new tasks. 
%In this work, we consider the question of how to develop an effective pre-training strategy tailored to downstream FL tasks.

%which involve training centralized models from pre-trained models

\begin{figure*}
\vspace{-4mm}
 \centering
\includegraphics[width=0.93\textwidth]{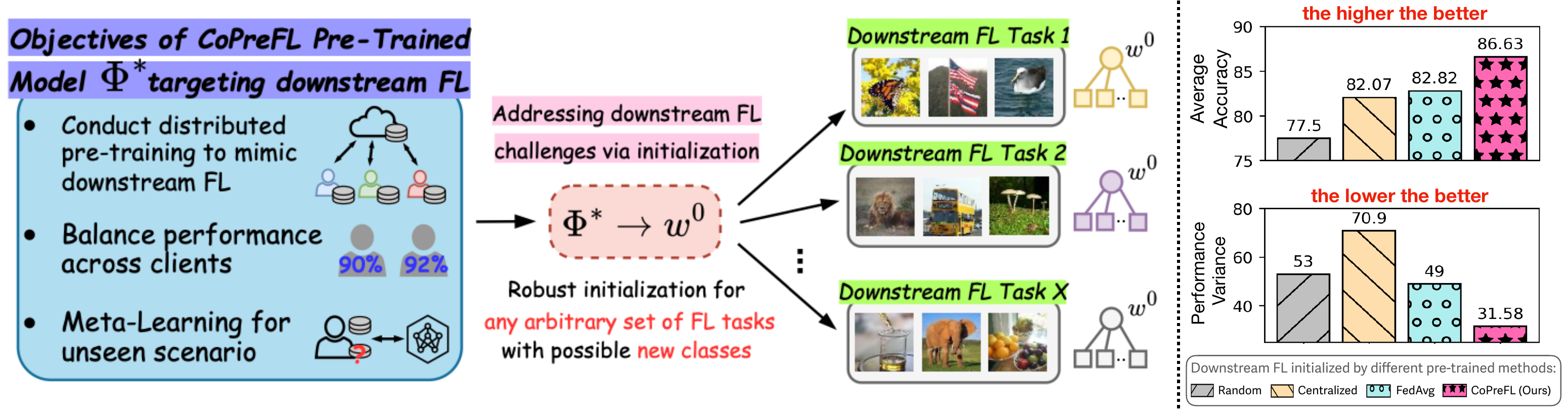}
\vspace{-4mm}
\caption{\small (Left): Overview of {\system}, aiming to provide a robust initialization for an arbitrary set of downstream FL tasks. (Right): Average accuracy and variance achieved by FL tasks (from Section~\ref{sec:experiment}) initialized by various pre-trained models. Centralized pre-training achieves limited performance as it is not able to capture the heterogeneous characteristics of unforseen FL settings. {\system} demonstrates improved  performance in terms of both average accuracy and variance %\yun{change to: both average accuracy and balanced performance across clients} 
by strategically mimicing downstream FL scenarios during pre-training. 
}
\label{fig:overview}
\vspace{-4.5mm}
\end{figure*}

\textbf{Motivation.}
% Recently, a few works \cite{nguyen2023where,chen2023on} have shown that initializing FL with \textit{centrally pre-trained models} can enhance the resulting average performance across clients. However, existing centralized pre-training methods face important drawbacks in practice, where they often have to handle \textit{newly emerging} and/or \textit{heterogeneous} downstream FL tasks unanticipated during pre-training.
Recently, a few works \cite{nguyen2023where,chen2023on} have shown that initializing FL with \textit{centrally pre-trained models} can enhance the average performance across clients. 
Yet, existing centralized pre-training methods face significant challenges, particularly when handling \textit{newly emerging} and/or \textit{heterogeneous} downstream FL tasks unanticipated during pre-training. These include: (i) limited average accuracy (despite outperforming random initialization), due to newly encountered data and labels, and (ii) large performance variance, resulting in unbalanced accuracy across clients.
% These drawbacks include: (i) limited average accuracy (despite outperforming random initialization), due to newly encountered data and labels, and (ii) large performance variance, failing to provide balanced accuracy across clients.
% The histograms in Figure~\ref{fig:overview}, which compare the performance of different pre-trained models in various downstream image classification tasks (see Section 4\ref{sec:experiment} for details), illustrate these limitations. 
% Compared to random initialization, although utilizing the centrally pre-trained model enhances average accuracy, it introduces substantial performance variance across clients, a well-cited concern in distributed AI/ML~\cite{Li2020Fair, Cho2022MaximizingGM}. 
% Furthermore, we see that the achievable average accuracy of centralized pre-training is also suboptimal. This indicates that such models struggle to mimic data heterogeneity and other diverse characteristics present in downstream FL tasks.
The histograms in Figure~\ref{fig:overview} show the performance of various pre-trained models in multiple downstream image classification tasks (see Section~\ref{sec:experiment} for details), illustrating these limitations. 
While using centrally pre-trained models improves average accuracy over random initialization, it introduces significant performance variance across clients, a well-cited concern in distributed AI/ML~\cite{Li2020Fair, Cho2022MaximizingGM}. 
Additionally, the achievable average accuracy of centralized pre-training remains suboptimal, indicating that such models struggle to mimic data heterogeneity and diversity in downstream FL tasks.

\textbf{Goals.}
% Motivated by these limitations, the goal of this paper is to develop a robust FL pre-training methodology that provides a model initialization which achieves two main objectives: (i) \textit{improved average accuracy}, and (ii) \textit{reduced performance variance} to achieve balanced accuracy across clients, in each downstream FL task encountered. 
% This is particularly challenging as it must be achieved across \textit{an arbitrary set} of downstream FL tasks, potentially facing data statistics and labels that were unseen during pre-training, due to time-varying environments (e.g., self-driving cars classifying previously unseen objects), new clients joining the system (e.g., face/speech recognition for new phone users), or other factors. Therefore, the pre-trained model must adapt well to unfamiliar classes and accommodate new data heterogeneity during downstream FL tasks, a challenge overlooked by existing methods \cite{nguyen2023where,chen2023on}.
Motivated by these limitations, we aim to develop a robust FL pre-training methodology that provides an initialization which achieves two main objectives: (i) \textit{improved average accuracy}, and (ii) \textit{reduced performance variance} for balanced accuracy across clients, in each downstream FL task. 
% This is particularly challenging as it must be achieved across \textit{an arbitrary set} of downstream FL tasks, which may include data and labels previously unseen due to factors like time-varying environments (e.g., self-driving cars classifying previously unseen objects) and new clients joining the system (e.g., face/speech recognition for new phone users).
This is challenging as it must work across \textit{any arbitrary set} of downstream FL tasks, which may include data and labels previously unseen due to factors like time-varying environments or new clients joining the system.
Thus, the pre-trained model must handle unfamiliar classes and data heterogeneity during downstream FL tasks, a challenge overlooked by existing methods \cite{nguyen2023where,chen2023on}. 
We summarize our research question as follows:
%. In practice, the data samples and labels available during pre-training may differ from those encountered during downstream tasks

% \vspace{-1.5mm}
\begin{itemize}[leftmargin=*]
\item[] \textit{How can we design a pre-training strategy that can simultaneously (i) enhance average accuracy and (ii) reduce performance variance across clients, for an arbitrary set of downstream FL tasks which possess heterogeneity in their data statistics as well as unseen labels?}
% \vspace{-1.5mm}
\end{itemize}

\textbf{Contributions.} We propose {\system}, a \underline{Co}llaborative \underline{Pre}-training approach for handling an arbitrary set of downstream \underline{FL} tasks, to address the above question. We make the following key contributions:  %{\system}   addresses the above challenges with the following key innovations:
\begin{itemize}
[leftmargin=6mm]
\vspace{-1.5mm}
% To handle the unseen data challenge, 
%closely mimics the goal of downstream FL tasks, namely, construction of a robust global model,
\item \textit{Distributed pre-training infused with meta-learning}: 
% The foundation of {\system} is an FL-inspired pre-training procedure which iteratively applies model-agnostic meta-learning (MAML)-based updates on the collaboratively-constructed global model. Through our developed MAML procedure, {\system} produces robust initializations in the sense of enabling the pre-trained model to easily adapt to unseen labels and various data distributions encountered in any downstream FL tasks. We note that this approach differs in purpose and method from prior works that use meta-learning for personalization in FL~\cite{Chen2018FederatedMW, Jiang2019ImprovingFL, Fallah2020PersonalizedFL}: since our downstream tasks aim to construct a global model rather than client-specific personalized models, we conduct meta-updates based on the \textit{global model} instead of directly using local models.
% The limitations of existing meta-learning methods for pre-training will be empirically demonstrated in Section 4.2\ref{sec:exp_result}.
The core of {\system} is an FL-inspired pre-training procedure which employs model-agnostic meta-learning (MAML)-based updates on the collaboratively-bulit global model. Through our developed MAML procedure, {\system} ensures robust initializations, enabling the pre-trained model to easily adapt to unseen labels and various data distributions in any downstream FL tasks. This approach differs in purpose and method from prior works that use meta-learning for personalization in FL~\cite{Chen2018FederatedMW, Jiang2019ImprovingFL, Fallah2020PersonalizedFL}: since our downstream tasks aim to construct a global model rather than client-specific personalized models, we conduct meta-updates based on the \textit{global model} instead of directly using local models.
% The limitations of existing meta-learning methods for pre-training will be empirically demonstrated in Section~\ref{sec:exp_result}.

\item \textit{Meta-objective function incorporating variance}: To enhance average accuracy while improving performance balance among clients in the downstream FL tasks, we explicitly incorporate both expected loss and performance variance into the meta-objective function during pre-training in {\system}. In doing so, we introduce a first-order approximation for efficiently computing the gradient of the proposed meta-objective function. %This new objective function is computed using (i) the query dataset of the clients and (i) the global model constructed based on the support datasets of the clients.%Adopt second order approximation to efficiently implement it with less complexity? (can remove this if we have enough materials here) \textcolor{blue}{To be updated}

\item \textit{Relaxing the assumption of centrally stored pre-training data}: 
% {\system} relaxes the assumption made by existing works that all pre-training data is stored centrally. We develop our pre-training algorithm for a range of hybrid client-server data storage settings, where (i) data is exclusively held by distributed clients, or (ii) the server also maintains a portion of data. This aligns with many practical distributed/FL applications that possess data privacy limitations and/or communication constraints. Our approach is applicable even when all pre-training data is centrally stored, which is also validated through our experiments.
{\system} relaxes the assumption made by existing works that all pre-training data is stored centrally. We develop our pre-training algorithm for various hybrid client-server data storage settings, where (i) data is exclusively held by distributed clients, or (ii) the server also maintains partial data. This is crucial for distributed/FL applications with data privacy limitations. Our approach is applicable even when all pre-training data is centrally stored, as confirmed by our experiments.

\item \textit{Extensive experiments across various downstream FL tasks and settings}:
We compare {\system} against various baselines across diverse downstream FL tasks with different data distributions, seen/unseen labels, and client-server data allocations. 
Results show notable improvements in accuracy and performance variance when downstream tasks are initialized with {\system}. We also show {\system}'s compatibility with various popular FL algorithms used downstream and its resilience to distributional shifts.

% \vspace{-2.5mm}
\end{itemize}
Our work is among the first to consider FL in both the pre-training and downstream stages of distributed learning tasks. We introduce several unique features tailored to FL, including meta-updating the global model during distributed pre-training, hybrid client-server learning, and balancing between average performance and variance across the clients.

\vspace{-3mm}

\section{Related Work}\label{sec:related}
% \vspace{-2.2mm}

\textbf{Pre-training for FL.} 
% While pre-training has been extensively studied in centralized AI/ML applications~\citep{Radford2019LanguageMA, Brown2020LanguageMA, Devlin2019BERTPO, Dosovitskiy2020AnII}, its effects on downstream FL tasks have remained relatively unexplored. A few recent works have studied the effect of model initialization in FL \cite{nguyen2023where,chen2023on}, showing that conducting FL initialized from centrally pre-trained models can significantly enhance average performance compared with random initializations.
% However, as observed in Figure~\ref{fig:overview}, such initialization strategies introduce large performance variance  and even limited average accuracy, since they are not able to  mimic multiple downstream FL settings. 
% To overcome this, we develop a MAML-based pre-training strategy tailored to distributed downstream settings, so that initializing FL with the pre-trained model can improve both average accuracy and performance variance while addressing the challenges of heterogeneous/unseen data encountered in each downstream FL task. We demonstrate the advantages of {\system} over existing pre-training schemes through extensive experiments in Section 4\ref{sec:experiment}. 
While pre-training has been extensively studied in centralized AI/ML applications~\citep{Radford2019LanguageMA, Brown2020LanguageMA, Devlin2019BERTPO, Dosovitskiy2020AnII}, its effects on downstream FL tasks remains less explored. A few recent works have shown that starting FL with centrally pre-trained models can improve performance over random initializations~\cite{nguyen2023where,chen2023on}.
However, as observed in Figure~\ref{fig:overview}, such initialization strategies often lead to large performance variance and even limited average accuracy, since they are not able to  mimic multiple downstream FL settings. 
To address this, we develop a MAML-based pre-training strategy, {\system}, tailored for distributed downstream settings, so that initializing FL with this pre-trained model can improve both average accuracy and performance variance while addressing the challenges of heterogeneous/unseen data encountered in downstream FL tasks.
 % We demonstrate the advantages of {\system} over existing pre-training schemes through extensive experiments in Section~\ref{sec:experiment}. 

\vspace{-0.2mm}

\textbf{Meta-learning in FL.}
% {\system} employs meta-learning, aiming to construct a global model that is adaptable to arbitrary downstream FL tasks while jointly optimizing performance and variance.
% This distinguishes it from other FL methods that have employed meta-learning, such as personalized FL~\cite{Chen2018FederatedMW, Jiang2019ImprovingFL, Fallah2020PersonalizedFL} and few-round FL \cite{Park2021FewRoundLF}.
{\system} utilizes meta-learning to construct a global model adaptable to any downstream FL tasks, optimizing both performance and variance.
This distinguishes it from other meta-learning-based FL, like personalized FL~\cite{Chen2018FederatedMW, Jiang2019ImprovingFL, Fallah2020PersonalizedFL, Chu2022MitigatingBI} and few-round FL \cite{Park2021FewRoundLF}.
% Specifically, \cite{Park2021FewRoundLF} employs a meta-learning-based episodic training strategy to adapt FL to any client group within only a few rounds of FL.
% However, \cite{Park2021FewRoundLF} does not consider performance imbalance across clients as an objective. Further, the practical scenario in which the server holds a proxy dataset for hybrid pre-training is not considered.
% Specifically, \citet{Park2021FewRoundLF} uses a meta-learning-based episodic training strategy to adapt FL to any client group within a few rounds.
% However, \cite{Park2021FewRoundLF} overlooks performance imbalance across clients and does not account for scenarios where the server holds a proxy dataset for hybrid pre-training.
Specifically, \citet{Park2021FewRoundLF} uses meta-learning-based episodic training for quick FL adaptation, but it overlooks performance imbalance across clients and fails to consider scenarios where the server holds a proxy dataset for hybrid pre-training.
% In personalized FL~\citep{Chen2018FederatedMW, Jiang2019ImprovingFL, Fallah2020PersonalizedFL}, meta-update is conducted based on the performance of each client's local model, aiming for better personalization to clients in downstream tasks.
% By contrast, our emphasis is on establishing a pre-trained model that results in high-accuracy and equitable \textit{global models} for downstream tasks \textit{that themselves are trained through FL}, requiring meta-updates based on the performance of the global model. 
In personalized FL, meta-update target individual client model performance, aiming for better personalization to clients in downstream tasks.
In contrast, our focus is on developing a pre-trained model that ensures high accuracy and fairness for \textit{global models} in downstream FL tasks.

\textbf{Performance imbalance in FL.}
% Several works in the FL literature have considered performance imbalance~\citep{Mohri2019AgnosticFL, Li2020Fair, Cho2022MaximizingGM}, typically aiming to construct a global model that satisfies as many clients as possible (e.g., achieving a uniform distribution of accuracy across participants). When properly architected, such models have a higher likelihood of satisfying new clients joining the FL system without requiring additional model training. 
Several works in FL address performance imbalance~\citep{Mohri2019AgnosticFL, Li2020Fair, Cho2022MaximizingGM} typically by creating a global model that satisfies as many clients as possible (e.g., achieving a uniform accuracy across clients). 
Such models are more likely to satisfy new clients joining the FL system.
% These methods can be applied downstream from our pre-training methodology, whose primary objective is to construct a robust \textit{initial} model that will lead to higher average and more balanced performance across clients after FL training. Our experiments in Section 4\ref{sec:experiment} will show how initializing these methods from {\system} further enhances their performance.
These methods can be applied downstream from our pre-training methodology, whose primary objective is to build a robust \textit{initial} model that will lead to higher average and more balanced performance across clients after FL training. 
% Our experiments in Section~\ref{sec:experiment} will show how starting from our {\system} further boosts their performance.

\vspace{-1mm}

  \section{Proposed {\system} Methodology}
% \vspace{-2.3mm}

\subsection{Problem Setup and Pre-Training Objectives}
 % \vspace{-2mm}

\textbf{Federated downstream tasks.}
Referring to Figure~\ref{fig:overview}, {\system} aims to provide a robust initialization for any downstream FL tasks. In each of these downstream tasks, we assume a central server is connected to a set of clients $G$. Starting from the initialized model $w^0$, each FL task iterates between (i) local training at the clients and (ii) global aggregations at the server, across multiple communication rounds. In each downstream round $r$, every client $g \in G$ downloads the previous global model $w^{r-1}$ from the server and subsequently updates it through multiple iterations of stochastic gradient descent (SGD) using its local dataset, denoted $D_g$. 
After completing their local updates, clients upload their updated models, denoted $w^{r}_{g}$, to the server for aggregation.
This aggregation results in a new global model $w^{r} = \sum_{g \in G}\frac{|D_g|}{|D|} w^{r}_g$ assuming FedAvg~\citep{McMahan2017CommunicationEfficientLO} is employed, where $|D|$ is the total data samples across all clients.
This entire process is iterated over $r = 1,..., R$ communication rounds for each task.

\textbf{Pre-training scenarios.} 
One of our contributions is relaxing the assumption that all pre-training data is stored centrally.
To this end, we consider two distributed pre-training scenarios for {\system}:
% \vspace{-4mm}
\begin{itemize}
\vspace{-1.5mm}
    \item \textbf{Scenario I:} Pre-training datasets are  exclusively available at distributed clients. %The conventional scenario encountered in FL where pre-training datasets are exclusively available at the clients.
    \vspace{-1mm}
    \item \textbf{Scenario II:} A hybrid scenario where the server also holds a small amount of pre-training data.
\end{itemize}
\vspace{-1.5mm}

% {\tt Scenario} I emulates the setting encountered by the downstream FL tasks for their training. A key challenge is that the labels and data that appear in the pre-training stage may be different from the ones that will appear in the downstream tasks. 
% {\tt Scenario} II emulates settings where the server providing the downstream model initializations holds some data that reflects the broader population distribution (e.g., a self-driving car manufacturer with a data store of images encountered on roadways). 
% Such hybrid FL settings that combine client data with a relatively small portion of server data are becoming   popular~\citep{Yang2023OnTC, Bian2023AcceleratingHF}, but have not been investigated for pre-training. Further, as we will discuss in \textbf{Remark 2}, our method is still applicable even in settings where all pre-training data is centralized.
{\tt Scenario} I simulates downstream FL tasks where pre-training labels and data may differ from those in downstream tasks.
{\tt Scenario} II represents settings where the server holds data reflecting the broader population distribution (e.g., a self-driving car manufacturer with database of roadway images). 
Such hybrid FL settings that combine client data with a relatively small portion of server data are becoming  popular~\citep{Yang2023OnTC, Bian2023AcceleratingHF}, but remain underexplored for pre-training. Further, as we will discuss in \textbf{Remark 2}, our method is still applicable even when all pre-training data is centralized.

\textbf{Pre-training objectives.}
Our goal is to design a pre-trained model $\Phi^*$ that serves as a robust starting point $w^0=\Phi^*$ for any arbitrary downstream FL task.
% \yun{emphasize we provide a initialization for "any downstream FL task" (A, P(A) should be defined here).}
% ${A}_t|_{t=1}^{T}$.
More precisely, letting $\mathcal{G}$ represents possible client sets comprising a downstream FL task, the goal of designing $\Phi^*$ is to minimize the following objective function:
 \vspace{-0.5mm}
\begin{equation}\label{eq:objtive1}
% F(\Phi) = \mathbb{E}_{{A}_t \sim p(\mathcal{A})} \left[ \frac{1}{K} \sum_{k \in {A}_t} f(\theta^R(\Phi), D^f_k) \right],
% F(\Phi) = \mathbb{E}_{A \sim p(\mathcal{A})} \left[ \frac{1}{K} \sum_{k \in K} f(w^R(\Phi), D_k) \right],
\small
A(\Phi) = \mathbb{E}_{G \sim p(\mathcal{G})} \biggl[ \frac{1}{|G|} \sum_{g \in G} f(w^R(\Phi, G), D_g) \biggr],
\end{equation}
% where ${A}_t$ represents a specific group drawn from the distribution $p(\mathcal{A})$ in the fine-tuning dataset ${D}^{f}$, $K$ denotes the total number of participants, $\theta^R(\Phi)$ represents the global model initialized through $\Phi$ after $R$ rounds of FL within group ${A}_t$, and $D^f_k$ represents the fine-tuning dataset of participant $k$ within group ${A}_t.
where $p(\mathcal{G})$ represents the probability distribution over $\mathcal{G}$, $G$ is a specific client group (i.e., a specific task) drawn from $p(\mathcal{G})$, $f(\cdot)$ is the per-client loss function for downstream training, $w^R(\Phi, G)$ symbolizes the final $R$-th round global model derived from client set $G$ when initialized by $\Phi$, and $D_g$ represents the local dataset of client $g$. $A(\Phi)$ denotes the average FL performance across all clients fir downstream tasks, with each group weighted by likelihood of occurrence.

On the other hand, FL settings can lead to significant performance variations among clients, especially when the aggregated models are biased towards those with larger datasets. This performance variation can be measured by the variance in testing accuracy across participants~\citep{Li2020Fair}.
Thus, besides improving performance for any FL task, we aim for the final global model $w^R(\Phi^*, G)$ initialized from our pre-trained model $\Phi^*$ to achieve balanced testing performance across client set $G$.
Specifically, our second objective for $\Phi^*$ is to minimize the variance of the loss distribution across participants in downstream FL tasks, i.e., 
\vspace{-1mm}
\begin{equation}\label{eq:objtive2}
\small
% \begin{split}
% F(\Phi) = \mathbb{E}_{{G}& \sim p(\mathcal{G})} \biggl[ \frac{1}{|G|} \sum_{g \in {G}} f^2(w^R(\Phi^*, G), D_g) \\
% &- \Big(\frac{1}{|G|} \sum_{g \in {G}} f(w^R(\Phi^*, G), D_g) \Big)^2 \biggr].
% \end{split}
\begin{split}
F(\Phi) = \mathbb{E}_{G \sim p(\mathcal{G})}& \biggl[ \frac{1}{|G|} \sum_{g \in G} f^2(w^R(\Phi^*, G), D_g) \\
&- \biggl(\frac{1}{|G|} \sum_{g \in G} f(w^R(\Phi^*, G), D_g) \biggr)^2 \biggr].
\end{split}
\end{equation}
\vspace{-3mm}

\textbf{Overview of approach.} 
% One of our key contributions is striking a balance between (\ref{eq:objtive1}) and (\ref{eq:objtive2}). 
% This is challenging as $D_g$, $\mathcal{G}$, and $p(\mathcal{G})$ are not known during pre-training, preventing us from directly optimizing $A(\Phi)$ and $F(\Phi)$.
% To address this, we develop a model-agnostic meta-learning (MAML) approach for {\system} to mimic statistical heterogeneity found in downstream FL. 
% This approach yields pre-trained models that offer robust initialization for unseen downstream tasks, considering (\ref{eq:objtive1}) and (\ref{eq:objtive2}) in the MAML formulation.
% Specifically, in Section 3.2\ref{sec:methodsc1} and 3.3\ref{sec:method_sec2}, we construct a pre-training environment for scenarios I and II mirroring arbitrary federated setups encountered downstream, facilitating the pre-trained model's ability to account for data heterogeneity across clients and tasks. Our meta-learning-based {\system} updates the pre-trained model iteratively over federated rounds using a support set, followed by a concluding adjustment (meta-update) using a query set. 
% By treating the query set as unseen knowledge, our pre-trained model has the capability to effectively handle unforeseen FL scenarios in downstream tasks while striking a balance between (\ref{eq:objtive1}) and (\ref{eq:objtive2}).
One of our key contributions is balancing (\ref{eq:objtive1}) and (\ref{eq:objtive2}). 
The challenge arises as $D_g$, $\mathcal{G}$, and $p(\mathcal{G})$ are unknown during pre-training, preventing us from directly optimizing $A(\Phi)$ and $F(\Phi)$.
To address this, we develop a model-agnostic meta-learning (MAML) approach in {\system} to mimic statistical heterogeneity of downstream FL tasks. 
This method yields pre-trained models that offer robust initialization for unseen downstream tasks, considering (\ref{eq:objtive1}) and (\ref{eq:objtive2}).
Detailed in Sections~\ref{sec:methodsc1} and \ref{sec:method_sec2}, we construct a pre-training environment for scenarios I and II that mirrors  downstream federated setups, enabling the pre-trained model to handle data heterogeneity across clients and tasks.
Our meta-learning-based {\system} updates the pre-trained model iteratively over federated rounds using a support set, with a concluding adjustment (meta-update) using a query set treated as unseen knowledge. 
This enables our pre-trained model to the effectively handle unforeseen FL scenarios downstream while balancing between (\ref{eq:objtive1}) and (\ref{eq:objtive2}).

% \vspace{-10mm}
 \subsection{{\system} in Scenario I (Pre-training with Distributed Clients)}
 \label{sec:methodsc1}
\begin{algorithm}[t]
\small
% \begin{algorithm}
\caption{Our Pre-training Method {\system} (Pre-training Phase in {\tt Scenario} I)}
\label{alg:scenario1}
\begin{algorithmic}[1]
\STATE \textbf{Input:} A set of  clients $M$ in the pre-training phase, with each client $i$ holding its pre-training dataset $D^p_i$.
% participate in CoPreFL , each partipant clients each holds dataset $D^P_$
% Client dataset $D^C$ 
% \State Split $D^C$ to $M$ clients based on IID or non-IID distribution, and select $m \in M$ participants.
% \State Each participant $j \in m$ partitions their own dataset $D_j$ into support set $S_j$ and query set $Q_j$
\FOR {Each pre-training round $t = 1, 2, ..., T$} 
\STATE Randomly select a set of clients $m \subset M$  to participate
\STATE Each participant $j \in m$ partitions its own dataset $D^p_j$ into support set $S_j$ and query set $Q_j$
% \State Participants download global model $\phi^{t-1}$
\FOR {Each participant $j$ in parallel}
\STATE Download $\Phi^{t-1}$ from the server 
\FOR{local epoch $e = 1, 2, ..., E$}
% \STATE $\ell^{e}_{S_j}(\Phi^{t}) \gets \frac{1}{|S_j|} \sum_{(x,y)\in S_j} \ell(\Phi^{t-1}(x), y)$ \hfill\Comment{// Compute local support loss at each epoch}
\STATE $\ell^{e}_{S_j}(\Phi^{t}) \gets \frac{1}{|S_j|} \sum_{(x,y)\in S_j} \ell(\Phi^{t,e}_j(x), y)$ \hfill\COMMENT{ Compute local support loss at each epoch}
\STATE $\Phi^{t,e}_{j}$ $\gets$ $\Phi^{t,e-1}_{j}$ $-$ $\eta \nabla \ell^{e}_{S_j}(\Phi^t)$ \hfill\COMMENT{ Perform SGD local update using support loss}
\ENDFOR
% \STATE $\mathcal{L}_{S_j}(\Phi^t)$ $\gets$ $\frac{1}{E}$ $\sum_{e = 1}^{E} \ell^{e}_{S_j}(\Phi^t)$ \hfill\Comment{// Compute averaged support loss}
% \STATE $\Phi^t_{j}$ $\gets$ $\Phi^{t-1}$ $-$ $\eta \nabla \mathcal{L}_{S_j}(\Phi^t)$ \hfill\Comment{// Perform SGD local update using averaged support loss $\mathcal{L}_{S_j}$}
\ENDFOR
% \State $\overline{\phi^r}$ $\gets$ $\sum_{j=1}^{m} \frac{|S_j|}{|S|} \phi_j$
% \STATE $\overline{\Phi^t}$ $\gets$ $\sum_{j \in m} \frac{|S_j|}{\sum_{i \in m}|S_i|} \Phi^t_j$ \hfill\Comment{// Model aggregation to construct temporary global model}
\STATE $\overline{\Phi^t}$ $\gets$ $\sum_{j \in m} \frac{|S_j|}{\sum_{i \in m}|S_i|} \Phi^{t,E}_j$ \hfill\COMMENT{ Model aggregation to construct temporary global model}
\FOR {Each participant $j$ in parallel}
% \State Initialize each local model $\phi_j$ with $\overline{\phi^r}$
\STATE Download $\overline{\Phi^t}$ from the server
\STATE $\mathcal{L}_{Q_j}(\overline{\Phi^t})$ $\gets$ $\frac{1}{|Q_j|}$ $\sum_{(x,y) \in Q_j}{} \ell(\overline{\Phi^t}(x), y)$ \hfill\COMMENT{ Compute local loss (and gradient) using query set $Q_j$}
\ENDFOR
\STATE Server computes overall meta-loss $\mathcal{L}_{Q}(\overline{\Phi^t})$ and variance across meta-losses $\sigma_Q^2(\overline{\Phi^t})$ according to~\eqref{eq:ourobj}
\STATE $\mathcal{L}_{meta}(\overline{\Phi^t}) = \gamma \mathcal{L}_{Q}(\overline{\Phi^t}) + (1 - \gamma) \sigma_Q^2(\overline{\Phi^t})$  \hfill\COMMENT{ Customized query meta-loss}
\STATE $\Phi^t$ $\gets$ $\overline{\Phi^t}$ $-$ $\zeta \nabla \mathcal{L}_{meta}(\overline{\Phi^t})$ \hfill\COMMENT{ Meta-learning model update using customized loss}
\ENDFOR
\STATE \textbf{Output:} A pre-trained model for downstream FL tasks: $\Phi^T$
\end{algorithmic}
\end{algorithm}
% \vspace{-5mm}
\setlength{\textfloatsep}{7pt}

%\vspace{-\baselineskip} 

% \vspace{-1.5mm}

We first consider the scenario where pre-training data is distributed across $M$ distributed clients, and no data is stored on the server. 
%The key challenge we address is that the data and FL scenarios in downstream tasks are inherently unseen during the pre-training phase.
%Our goal is and employ meta-learning during the pre-training phase to formulate a pre-trained model that simulates the objectives we aim to reflect in downstream FL tasks.
The detailed procedure of {\system} for this case is given in Algorithm~\ref{alg:scenario1}.
%We first randomly involve a set of clients  $m$ to participate in each round, where each client $j$ holds its own dataset $D^p_j$.
To start, in each round $t = 1,...,T$ of pre-training, a set of clients $m \subset M$ is randomly selected to participate in the current round. Further, each participating client $j \in m$ splits its local pre-training dataset $D^p_j$ into a support set $S_{j}$ and query set $Q_{j}$, which are disjoint.
These steps mimic downstream task variations within and across each pre-training round (i.e., by changing the participating clients across rounds, and holding out the query sets in each round), allowing our meta-learning to improve generalization in unseen downstream scenarios.

\vspace{-0.5mm}

\textbf{Temporary pre-training model construction.} In each round $t$, participating clients $j \in m$ download $\Phi^{t-1}$ from the server. 
Subsequently, clients perform local training using their support sets $S_j$, yielding a local support loss $\ell^{e}_{S_j}(\Phi^{t})$ per epoch $e$, defined in line 8 of Algorithm~\ref{alg:scenario1}, where $\ell(\cdot)$ is the per-datum loss function (e.g., cross-entropy for classification). 
% After all participants finish $E$ epochs, we obtain the averaged support loss $\mathcal{L}_{S_j}(\Phi^t)$ (line 10) and update their local models using it to obtain $\Phi^t_j$ (line 11). 
After all participants finish $E$ epochs, we obtain the updated local model $\Phi^{t,E}_j$. 
Clients then send their updated models to the server for aggregation, resulting in $\overline{\Phi^t}$ (defined in line 12).
% with relative support set sizes $\mu_j =\frac{|S^j|}{\sum_{i \in m}|S_i|}$.
%The general FL scheme considers this point as the end of one round and broadcasts $\overline{\Phi^t}$ to all participants as the initialization for the next round. 
%However, our goal extends beyond this, as we aim to empower the model with the capability to handle unseen scenarios and mitigate performance variance through the use of query set. 
This model can be viewed as the temporary pre-training model that will be further refined using query sets, with the objective of obtaining robust global models at the conclusion of downstream task training.

\textbf{Measuring average performance and variance.} 
% Next, the query sets are used to evaluate the model's performance on each client and to conduct the meta-update process. 
Next, the query sets are used to evaluate the performance of the temporary pre-training model on each client, mimicking the scenario where the pre-trained model encounters unseen data, and to conduct meta-updates to promote downstream generalization.
%\yun{Next, the query sets are used to evaluate the model's performance on each client, mimicking the scenario where the initialization encounters unseen data, and to conduct the meta-update process to address challenges such as performance variance in downstream FL.}
{\system} aims to strike a balance between the following objectives during pre-training:
\vspace{-1mm}
\begin{equation}\label{eq:ourobj}
\small
\begin{split}
& \mathcal{L}_{Q}(\overline{\Phi^t}) = \sum_{j \in m} \mathcal{L}_{Q_j}(\overline{\Phi^t})  \text{ and}\\
& \sigma_Q^2(\overline{\Phi^t}) = \frac{1}{|m|} \sum_{j \in m} \Big(\mathcal{L}_{Q_j}(\overline{\Phi^t}) - \frac{1}{|m|}\mathcal{L}_{Q}(\overline{\Phi^t})\Big)^2,
\end{split}% = \min_{\phi}\sum_{j \in m} \frac{1}{|Q_j|}\sum_{(x,y)\in Q_j} \ell(\overline{\phi^t}(x), y)
\end{equation}
\iffalse
\vspace{-2mm}
\begin{equation}\label{eq:ourobj2}
\min_{\Phi} \sigma_Q^2(\overline{\Phi^t}) = \min_{\Phi} \frac{1}{|m|} \sum_{j \in m} \Big(\mathcal{L}_{Q_j}(\overline{\Phi^t}) - \frac{1}{|m|}\mathcal{L}_{Q}(\overline{\Phi^t})\Big)^2,
\end{equation}
\fi
where $\mathcal{L}_{Q_j}$ represents the loss evaluated using query sets $Q_j$ of participants, $\mathcal{L}_{Q}$ denotes the overall query loss (characterized by aggregating $\mathcal{L}_{Q_j}$ across all participants), and $\sigma_Q^2$ represents the performance variance evaluated using clients' query losses.
To balance the performance-variance tradeoff, we construct a customized query meta-loss function $\mathcal{L}_{meta}(\overline{\Phi^t})$ to minimize both the overall query loss $\mathcal{L}_{Q}(\overline{\Phi^t})$ when encountering unseen data and the variance $\sigma_Q^2(\overline{\Phi^t})$ of query losses across participants. Formally, we aim to solve:
\vspace{-0.8mm}
\begin{equation}\label{eq:balancer}
\small
\min_{\Phi} \mathcal{L}_{meta}(\overline{\Phi^t}) = \min_{\Phi} \left[ \gamma \mathcal{L}_{Q}(\overline{\Phi^t}) + (1 - \gamma) \sigma_Q^2(\overline{\Phi^t})\right],
\end{equation}
where $\gamma \in [0,1]$ represents a balancer between the average performance and variance.
Setting $\gamma = 0$ encourages a more uniform accuracy distribution, aligning with $\sigma_Q^2$, but may sacrifice average performance.
A larger $\gamma$ emphasizes the average performance with less consideration for uniformity, optimizing the pre-trained model more towards $\mathcal{L}_{Q}$.

\textbf{Model-agnostic meta update.}
Considering the objective function in~\eqref{eq:balancer}, each participant $j$ downloads the temporary global model $\overline{\Phi^t}$ and employs its query set $Q_j$ to compute its local query loss $\mathcal{L}_{Q_j}(\overline{\Phi^t})$, as in line 15 in Algorithm~\ref{alg:scenario1}. 
The gradients are also computed locally and sent back to the server, as both are necessary to conduct the meta-update.
On the server-side, the overall query meta-loss $\mathcal{L}_{Q}(\overline{\Phi^t})$ and the performance variance $\sigma_Q^2(\overline{\Phi^t})$ are computed, according to~\eqref{eq:ourobj}.
Then, {\system} updates the temporary pre-training model $\overline{\Phi^t}$ through a gradient step with the customized query meta-loss $\mathcal{L}_{meta}$ and the aggregated received gradients, to align it with~\eqref{eq:balancer}.
To derive the meta-loss $\nabla_{\Phi^{t-1}} \mathcal{L}_{meta}(\overline{\Phi^t})$, we express it through the chain rule as $\nabla_{\overline{\Phi^t}} \mathcal{L}_{meta}(\overline{\Phi^t}) \times \frac{\partial \overline{\Phi^t}}{\partial \Phi^{t-1}}$.
% The meta-update process in Line 20, which uses the derivation of the meta-loss, can be expanded through the chain rule as: $\nabla_{\overline{\Phi^t}} \mathcal{L}_{meta}(\overline{\Phi^t}) \times \frac{\partial \overline{\Phi^t}}{\partial \Phi^{t-1}}$.
% Writing $\overline{\Phi^t} = \sum_{j \in m} \frac{|S_j|}{\sum_{i \in m}|S_i|} \Phi^t_j = \sum_{j \in m} \frac{|S_j|}{\sum_{i \in m}|S_i|} (\Phi^{t-1} - \eta \nabla \mathcal{L}_{S_j}(\Phi^t))$, it follows that
Writing $\overline{\Phi^t} = \sum_{j \in m} \frac{|S_j|}{\sum_{i \in m}|S_i|} \Phi^{t,E}_j = \sum_{j \in m} \frac{|S_j|}{\sum_{i \in m}|S_i|} (\Phi^{t, E-1} - \eta \nabla \ell^{E}_{S_j}(\Phi^{t}))$, it follows that
\vspace{-0.8mm}
\begin{equation} \label{eq:meta-grad}
\small
\begin{split}
\nabla_{\Phi^{t-1}} &\mathcal{L}_{meta} (\overline{\Phi^t}) =  \nabla_{\overline{\Phi^t}} \mathcal{L}_{meta}(\overline{\Phi^t}) \times \\
&\Big(1 - \eta \sum_{j \in m} \frac{|S _j|}{\sum_{i \in m}|S _i|}\frac{\partial}{\partial \Phi^{t-1}} \nabla \ell^{E}_{S_j}(\Phi^{t}) \Big).
\end{split}
\end{equation}
%\begin{equation}
%\nabla_{\overline{\Phi^t}} \mathcal{L}_{meta}(\overline{\Phi^t}) \times \frac{\partial \overline{\Phi^t}}{\partial \Phi^{t-1}} = \nabla _{\overline{\Phi^t}} \mathcal{L} _{meta}(\overline{\Phi^t})\times \sum_{j \in m} \frac{|S _j|}{\sum_{i \in m}|S _i|}\frac{\partial }{\partial \Phi^{t-1}} (\Phi^{t-1} -\eta \nabla \mathcal{L} _{S _j}(\Phi^t)).
%\end{equation}
% By ignoring the second derivative term and adopting the first-order approximation, the derivation of meta-loss can be approximate as $\nabla_{\overline{\Phi^t}} \mathcal{L}_{meta}(\overline{\Phi^t})$.
If we ignore the second derivative term, the meta-loss gradient can be approximated as $\nabla_{\overline{\Phi^t}} \mathcal{L}_{meta}(\overline{\Phi^t})$. This is similar to making a first-order approximation to a meta-update, a common practice in the implementation of MAML variants to reduce complexity~\cite{Finn2017ModelAgnosticMF}.

The server then sends the meta-updated global model $\Phi^t$ to a new set of participants to begin the next round of pre-training. 
After $T$ rounds, the final global model $\Phi^T$ serves as the pre-trained model for initializing FL in the downstream tasks, i.e., in Figure~\ref{fig:overview}, clients in any downstream task conduct FL starting from the pre-trained model $w^0=\Phi^T$.
%\yun{Our method strategically employs two disjoint sets of participant data: support data to construct the temporary global model and query data for meta-updates. This dual approach effectively mimics unseen downstream FL tasks, preparing our model to adapt to new challenges in diverse FL environments.}

% \textbf{Remark 1 (Key characteristics of {\system} meta-update).} Using the \textit{query datasets}, {\system} applies a meta-update to the temporary pre-training model, which is a \textit{global model} obtained through FL on the \textit{support datasets}. This is a key distinction from existing meta-learning based FL methods discussed in Section 2\ref{sec:related}, which conduct meta updates on \textit{client models} for personalization. Our method focuses on tailoring the pre-training model to adapt to any downstream FL tasks, whereas existing personalization methods are not concerned with robustness to unseen and heterogeneous data statistics emerging downstream. 
% As we will see in Section 4\ref{sec:experiment}, this leads to significant improvements of {\system} compared with employing these prior methods for pre-training.
\textbf{Remark 1 (Key characteristics of {\system} meta-update).} 
% Using the \textit{query datasets}, {\system} applies a meta-update to the temporary pre-training model, which is a \textit{global model} obtained through FL on the \textit{support datasets}. This differs significantly from existing meta-learning based FL methods, discussed in Section 2\ref{sec:related}, which update \textit{client models} for personalization. Our method aims to tailor the pre-training model to adapt to any downstream FL tasks, whereas existing personalization methods are not concerned with robustness to unseen and heterogeneous data statistics downstream. 
% As we will see in Section 4\ref{sec:experiment}, this leads to significant improvements of {\system} compared with employing these prior methods for pre-training.
Using the \textit{query datasets}, {\system} applies a meta-update to the temporary pre-training model--a \textit{global model} developed through FL on the \textit{support datasets}. This differs significantly from existing meta-learning based FL methods, which update \textit{client models} for personalization, as discussed in Section~\ref{sec:related}. Our method aims to tailor the pre-training model to adapt to any downstream FL tasks, addressing robustness against unseen and heterogeneous data, unlike existing personalization methods. As we will see in Section~\ref{sec:experiment}, this leads to notable improvements of {\system} over employing these prior methods for pre-training.

\vspace{-2mm}

\subsection{{\system} in Scenario II (Hybrid Client-Server Pre-Training)}
\label{sec:method_sec2}
% \vspace{-2mm}
We next explore a pre-training scenario where the server holds a small dataset $D^s$ drawn from the broader population distribution, alongside client-held data. 
Unlike {\tt scenario} I, where client data was split into support and query sets, in {\tt scenario} II, all client samples are used as support data, while the server's data serves as the query set.
% , allowing us to customize the model according to our objectives. 
%This perspective can also be envisioned as the server receiving updates from clients and customizing the model according to their specific needs.
%\vspace{-0.3mm}
 
%\paragraph{Meta update on server:}
The procedure of {\system} for {\tt scenario} II is detailed in Algorithm~\ref{alg:scenario2} in Appendix~\ref{sec:app_detailscenarioII}.
% Here, we highlight the key differences from Algorithm~\ref{alg:scenario1}. First, the temporary pre-training model $\overline{\Phi^t}$ is aggregated from local models trained on the each participant's entire local dataset $D^p_j$.
% Second, we facilitate the meta-update of the temporary pre-training model $\overline{\Phi^t}$ using the server's data. In doing so, to help mimic the distributed nature of downstream FL tasks, we randomly partition the dataset $D^s$ into $|m|$ equally-sized partitions, from which we can obtain the average loss and variance objectives similar to~\eqref{eq:ourobj}.
% The temporary global model $\overline{\Phi^t}$ is then updated based on %server 
% meta-loss $\mathcal{L}_{meta}(\overline{\Phi^t})$, which is defined similarly to (\ref{eq:balancer}) but calculated through meta-updates on the server's partitioned data.
% \vspace{-4.7mm}
Here, we highlight the key differences from Algorithm~\ref{alg:scenario1}. First, the temporary global model $\overline{\Phi^t}$ is aggregated from local models, each trained on a participant's full local dataset $D^p_j$.
Second, the meta-update of the temporary global model $\overline{\Phi^t}$ utilizes the server's data. To help mimic the distributed nature of downstream FL tasks, we randomly split the dataset $D^s$ into $|m|$ equally to derive the average loss and variance objectives similar to~\eqref{eq:ourobj}.
The temporary global model $\overline{\Phi^t}$ is then updated based on meta-loss $\mathcal{L}_{meta}(\overline{\Phi^t})$, calculated similarly to (\ref{eq:balancer}) but through meta-updates using the server's partitioned data.

Note that unequal and/or non-uniformly random partitionings of the server-side dataset for meta-updating could be considered as alternatives. However, due to the server's lack of prior knowledge about future downstream FL tasks during pre-training, including their dataset sizes and distributions, meta-updating the model with randomly allocated query sets is the most viable solution.
We show in Section~\ref{sec:experiment} that this partitioning provides significant performance improvements over other pre-training strategies.
% Finally, in line 17 of Algorithm~\ref{alg:scenario2}, we employ the customized server meta-loss $\mathcal{L}_{meta}$ to update the temporary global model $\overline{\phi^t}$, aligning it with our controlled objectives. 
% After completing $T$ federated rounds, we regard the final global model $\phi^T$ as the pre-trained model $\Phi$ in scenario II, which serves as the initialization $w^0=\Phi$ for diverse FL tasks as described in Figure~\ref{fig:overview}.

% \vspace{-0.5mm}

\textbf{Remark 2 (Applications to centralized datasets).} 
Although we present {\system} for two distributed scenarios, it is applicable even when all pre-training data is stored at the server (e.g., public datasets). The server can intentionally split the dataset to mimic {\tt scenarios} I or II and directly apply {\system}.
We will show in Section~\ref{sec:experiment} that {\system} surpasses standard centralized pre-training even in this setup, offering initializations better prepared for downstream data heterogeneity in FL setups.

% \textbf{Remark 3.} Providing theoretical analysis on the effect of pre-training strategies continues to be an open problem, especially in linking model initialization to downstream task performance. This challenge has existed both in centralized-to-centralized \cite{Chang2020PretrainingTF,Dong2023SimMTMAS,Zhang2022SelfSupervisedCP,Hu2019StrategiesFP,Raffel2019ExploringTL,yuan2024pretraining} and centralized-to-federated \cite{nguyen2023where,chen2023on} transfers from pre-training to downstream, and persists in the distributed-to-federated case we consider here. We thus leave theoretical analysis of {\system} and other strategies to future work, and instead substantiate the effectiveness of our method through extensive experiments. Nevertheless, the effectiveness of {\system} can be attributed to the advantage provided by meta-learning, which we employ to obtain robustness across heterogeneous sets of downstream tasks.

\textbf{Remark 3.}The theoretical link between pre-training strategies and downstream task performance remains an open problem.
 This challenge has existed both in centralized-to-centralized \cite{Chang2020PretrainingTF,Dong2023SimMTMAS,Zhang2022SelfSupervisedCP,Hu2019StrategiesFP,yuan2024pretraining} and centralized-to-federated \cite{nguyen2023where,chen2023on} transfers from pre-training to downstream, and persists in the distributed-to-federated case we consider here. We thus leave theoretical analysis of {\system} to future work, and instead validate its effectiveness through extensive experiments. The success of {\system} can largely be credited to advantage from meta-learning, which enhances robustness across diverse sets of downstream tasks.

 \vspace{-3mm}

 \section{Experiments }\label{sec:experiment}
  % \vspace{-2.5mm}
\begin{table*}[t]
\vspace{-2mm}
    \centering
    \begin{subtable}{\textwidth}
        \centering
        \scalebox{0.7}{
        \begin{tabular}{c||cccc||cccc}
        \hline
        {\textbf{Pre-training}} & \multicolumn{4}{c||}{\textbf{Downstream:} Non-IID FedAvg (CIFAR-100)} & \multicolumn{4}{c}{\textbf{Downstream:} Non-IID FedAvg (Tiny-ImageNet)}\\        \hline\hline
         {\textbf{Method}} & \textbf{Acc} $\uparrow$ & \textbf{Variance} $\downarrow$ & \textbf{Lowest 10\%} $\uparrow$& \textbf{Lowest 20\%} $\uparrow$& \textbf{Acc} $\uparrow$ & \textbf{Variance} $\downarrow$ & \textbf{Lowest 10\%} $\uparrow$& \textbf{Lowest 20\%} $\uparrow$\\
        \hline
        {FedAvg} & 78.96  $\pm$ 2.98 &	64.80  $\pm$ 3.01 &		62.70  $\pm$ 3.35 &	67.00  $\pm$ 2.95  & 82.94  $\pm$ 2.59 &	37.21  $\pm$ 2.81 &		68.99  $\pm$ 2.43 &	72.29  $\pm$ 2.61 \\
        {FedMeta} & 82.45  $\pm$ 3.07 &	48.72 $\pm$ 2.84 &	68.97 $\pm$ 3.04 &	72.41  $\pm$ 3.06  & 81.03  $\pm$  2.86 &	37.58  $\pm$ 3.00 &		69.44  $\pm$ 2.61 &	71.55 $\pm$ 2.93 \\
        {q-FFL} & 80.01  $\pm$ 2.67& 88.92 $\pm$ 3.31 &		64.39  $\pm$ 2.95 &	67.48  $\pm$ 2.67  & 84.11 $\pm$ 2.49  &	43.96 $\pm$  2.71 &		73.87 $\pm$  2.79 &	76.05  $\pm$ 2.61 \\
        \systemnott & \textbf{83.29} $\pm$ 2.61 &	\textbf{34.69}  $\pm$ 3.17 &		\textbf{71.58}  $\pm$ 3.00&	\textbf{73.20}  $\pm$ 2.98&	 \textbf{85.23}  $\pm$ 2.43 &	\textbf{35.40} $\pm$ 2.75  &	\textbf{76.77} $\pm$ 2.58  &	\textbf{78.46}  $\pm$ 2.47  \\
        \hline
        \end{tabular}       }
        \caption{Results in {\tt scenario} I using CIFAR-100 and Tiny-ImageNet datasets. \label{tbl:sc1}}
        % \vspace{-1mm}         
    \end{subtable}
% \quad%  
\\
    \begin{subtable}{\textwidth}
        \centering
        \scalebox{0.7}{
        \begin{tabular}{c||cccc||cccc}
        \hline
        {\textbf{Pre-training}} & \multicolumn{4}{c||}{\textbf{Downstream:} Non-IID FedAvg (CIFAR-100)} & \multicolumn{4}{c}{\textbf{Downstream:} Non-IID FedAvg (Tiny-ImageNet)}\\       
        \hline\hline
         {\textbf{Method}} & \textbf{Acc} $\uparrow$ & \textbf{Variance} $\downarrow$ & \textbf{Lowest 10\%} $\uparrow$& \textbf{Lowest 20\%} $\uparrow$& \textbf{Acc} $\uparrow$ & \textbf{Variance} $\downarrow$ & \textbf{Lowest 10\%} $\uparrow$& \textbf{Lowest 20\%} $\uparrow$\\
        \hline
        {FedAvg} & 82.82  $\pm$ 3.17&	49.00  $\pm$  3.41&		69.71  $\pm$ 3.25 &	72.54  $\pm$ 3.30 &  82.87  $\pm$ 3.19  &	48.16  $\pm$ 2.94 &		68.94  $\pm$ 3.38 &	72.91  $\pm$ 3.49 \\
        {FedMeta} & 82.69 $\pm$ 3.05 & 	48.44  $\pm$ 2.99 &		68.84  $\pm$ 3.14 &	71.82  $\pm$ 3.27 & 84.19   $\pm$ 2.93 &	49.70  $\pm$ 2.74  &		70.41  $\pm$ 3.16 &	72.63  $\pm$ 3.00 \\
        {q-FFL} & 82.14 $\pm$ 2.76 &73.10  $\pm$ 3.08 &		68.22  $\pm$ 3.00 &	70.64  $\pm$ 2.85 &  83.51   $\pm$ 3.05 &	44.22   $\pm$ 3.22 &	69.91   $\pm$ 2.94 &	73.71  $\pm$ 3.14 \\
        \systemnott-SGD  & 83.63 $\pm$ 3.00  &	41.73 $\pm$ 2.85 &		69.76  $\pm$ 2.94 & 73.46  $\pm$ 3.09   & 84.30   $\pm$ 2.77 &	36.24  $\pm$ 3.04  &	72.83   $\pm$ 2.99 &	75.64  $\pm$ 3.18 \\
        \systemnott  & \textbf{86.63}  $\pm$ 2.93 &	\textbf{31.58}  $\pm$ 2.64 &	\textbf{73.05}  $\pm$ 2.51 &	\textbf{75.82}  $\pm$ 2.88  &  \textbf{84.72}   $\pm$ 2.51 &	\textbf{24.80}  $\pm$ 3.00 &		\textbf{75.84}   $\pm$ 2.87 &	\textbf{77.31}   $\pm$ 3.13 \\
        \hline
        \end{tabular}
        }
        \caption{Results in {\tt scenario} II using CIFAR-100 and Tiny-ImageNet datasets. 
        \label{tbl:sc2}}
        \vspace{-3mm}   
    \end{subtable}
        \vspace{-2mm}
    \caption{\small Performance on 10 non-IID downstream FL tasks, initialized with various non-IID FL pre-training methods. Lowest $X$\% shows the average accuracy of clients with the lowest $X$\% accuracy.
    {\system} provides the best initialization across  scenario, metric, and dataset.}\label{tbl11111}
    \vspace{-2mm}
\end{table*}

\subsection{Experimental Setup}
   % \vspace{-1.5mm}
   \label{sec:expsetup}
\textbf{Datasets and model.}
% For evaluation, we consider CIFAR-100~\citep{Krizhevsky2009LearningML}, Tiny-ImageNet~\citep{Le2015TinyIV}, FEMNIST~\citep{Caldas2018LEAFAB}, and PACS~\cite{Li2017DeeperBA}, adhering to the data splits provided in~\citep{Ravi2016OptimizationAA, Park2021FewRoundLF}, and adopt ResNet-18~\citep{He2015DeepRL}.
% To model practical scenarios where downstream task labels are unknown during pre-training, for CIFAR-100, the dataset is divided into 80 classes for pre-training and 20 classes for downstream FL tasks, while for Tiny-ImageNet, the dataset is separated into 160 classes for pre-training and 40 classes for downstream FL tasks. 
% Additionally, we explore mixed scenarios where there are overlapping classes between pre-training and downstream tasks.
% Following~\cite{Yang2023OnTC,Zhang2020HybridFL}, we randomly select 95\% of the samples from the pre-training dataset for clients, while the remaining 5\% form the server dataset.
% The PACS dataset consists of 4 different data domains, for which we employ a one-domain-leave-out setup~\cite{Li2022UncertaintyMF,Zhou2021DomainGW} between pre-training and downstream FL tasks. More details on the datasets can be found in Appendix C.1\ref{ap:dataset}. For FEMNIST, we report all results in Appendix D.4\ref{ap:femnist}.
 For evaluation, we use CIFAR-100~\citep{Krizhevsky2009LearningML}, Tiny-ImageNet~\citep{Le2015TinyIV}, FEMNIST~\citep{Caldas2018LEAFAB}, and PACS~\cite{Li2017DeeperBA}, following data splits provided in~\citep{Park2021FewRoundLF}, and adopt ResNet-18~\citep{He2015DeepRL}.
To model scenarios where downstream task labels are unknown during pre-training, we divide CIFAR-100 into 80 classes for pre-training and 20 for downstream tasks, and Tiny-ImageNet into 160 and 40 classes, respectively.
We also explore mixed scenarios involving overlapping classes between pre-training and downstream tasks.
Following~\cite{Yang2023OnTC,Zhang2020HybridFL}, we allocate 95\% of the samples from the pre-training dataset for clients, and the remaining 5\% form the server dataset.
For PACS, we adopt a one-domain-leave-out setup~\cite{Li2022UncertaintyMF,Zhou2021DomainGW} using different data domains for pre-training and downstream tasks. 
Detailed dataset information is available in Appendix~\ref{ap:dataset}.

% \vspace{-0.3mm}

\textbf{Pre-training phase.}
We distribute the pre-training dataset to $|M| = 100$ clients following non-IID data partitions according to a Dirichlet distribution~\citep{Morafah2022RethinkingDH,Li2021ModelContrastiveFL}, and select $|m| = 20$ participants out of the $|M|$ clients for each FL round. Results with different $|m|$ and IID setups are reported throughout Appendix~\ref{ap:add_exp}.
% Each participant employs 80\% of its local data as support samples and the remaining 20\% as query samples. 
 We adopt a standard approach commonly used in meta-learning-based research \cite{Jamal2020RethinkingCM, Shu2019MetaWeightNetLA, Park2021FewRoundLF} for support/query splitting, where we randomly partition each client's data into 80\% support and 20\% query sets.
% The SGD optimizer with a learning rate of $\eta = 10^{-3}$ and $\zeta = 10^{-3}$ is adopted for both local and meta updates. \yun{app}
% In \textbf{scenario II}, we additionally perform model updates using the server dataset $D^S$ (\yun{refer to the baseline section}).
See Appendix~\ref{ap:dataset}-\ref{ap:parameter} for more details. %detailed hyperparameters and compute settings. 

% \vspace{-0.3mm}

% \cgb{4.1 seems a bit long. Can the following paragraph be shortened and split into two?}

\textbf{Downstream FL task and evaluation metrics.} 
To generate each downstream FL task, we randomly select 5 of the 20 classes from the CIFAR-100 dataset and 40 classes from the Tiny-ImageNet dataset, and distribute the corresponding data samples to a set of $|G|=10$ clients following non-IID Dirichlet data distributions (see Appendix~\ref{ap:add_exp} for IID results). 
Each participant in the downstream phase utilizes 80\% of its local data as training samples, while the remaining 20\% is reserved for testing samples.
We keep the training procedure consistent for each downstream task (see Appendix~\ref{ap:parameter} for detailed settings). 
For each task, we evaluate the final global model using test samples from each client $g \in G$, reporting the accuracy and variance of the accuracy distribution across the clients. 
We consider a total of $X = 10$ downstream tasks, and the evaluation metrics are reported as averages (with standard deviations) across the tasks.

\textbf{Data distribution.}
% In the IID setup, data samples from each class are distributed equally to $|M| = 100$ clients for pre-training and $|G| = 10$ clients for downstream FL task.
% In the non-IID setup, samples within each class are partitioned among $|M|$ and $|G|$ clients using a Dirichlet($\alpha$) distribution for pre-training and downstream task, respectively, with $\alpha = 0.5$ selected as is in the literature~\citep{Morafah2022RethinkingDH,Li2021ModelContrastiveFL}. 
Data samples are distributed to $|M| = 100$ clients for pre-training and $|G| = 10$ clients for downstream FL tasks using the corresponding dataset based on a Dirichlet($\alpha$) distribution with $\alpha = 0.5$, as done in the  literature~\citep{Morafah2022RethinkingDH,Li2021ModelContrastiveFL}.

\textbf{Baselines for pre-training.}
% We compare {\system} with various established FL algorithms, including standard FedAvg~\citep{McMahan2017CommunicationEfficientLO}, FedMeta~\citep{Chen2018FederatedMW}, which addresses the unseen scenario through meta-learning, and q-FFL(q $>$ 0)~\citep{Li2020Fair}, which aims at balancing performance across clients.
We compare {\system} with several established FL algorithms, including (i) standard FedAvg~\citep{McMahan2017CommunicationEfficientLO}, (ii) FedMeta~\citep{Chen2018FederatedMW}, which employs meta-learning for unseen scenarios, and (iii) q-FFL(q $>$ 0)~\citep{Li2020Fair}, designed to balance performance across clients.
% For fair comparison, all schemes are adopted during the pre-training to construct initial models under the same setting.
%For a fair comparison, {\system} and all baselines were trained from scratch under the same settings to obtain their pre-training models. Subsequently, in each downstream task, FL is conducted using the downstream dataset, starting from the respective pre-trained models. %Both the baseline methods and {\system} served as initializations for these downstream FL tasks. 
% For a fair comparison,   we maintain consistent settings including the number of global rounds, local iterations, and the chosen $|m|$ participants across all schemes. 
When applying these baselines in {\tt scenario} II, in each pre-training round $t$, after the global model $\Phi^t$ has been constructed, we further train with 5 additional iterations on the server dataset.
This extended training %, optimized using the SGD optimizer with a learning rate of $10^{-3}$, 
follows the approach in~\cite{Yang2023OnTC, Bian2023AcceleratingHF}, where the server's data is used to further refine the global model.
%Subsequently, the server broadcasts the server-trained global model to each participant for the next round. 
Similarly, we introduce a baseline called {\tt CoPreFL-SGD}, which first constructs a global model according to {\system} and then further performs SGD iterations using server data on the global model. 
Finally, we also consider initializations based on (iv) conventional centralized pre-training~\cite{nguyen2023where}, and popular FL algorithms like (v) SCAFFOLD~\cite{Karimireddy2019SCAFFOLDSC}, (vi) FedDyn~\cite{Acar2021FederatedLB}, and (vii) PerFedAvg~\cite{Fallah2020PersonalizedFL} (see Table~\ref{tbl:otherinit}). %\textcolor{red}{**OK until here!!**}

 \begin{figure*}[t]
 \vspace{-2mm}
    \centering
\setlength{\abovecaptionskip}{1mm}  
\includegraphics[width=\linewidth]{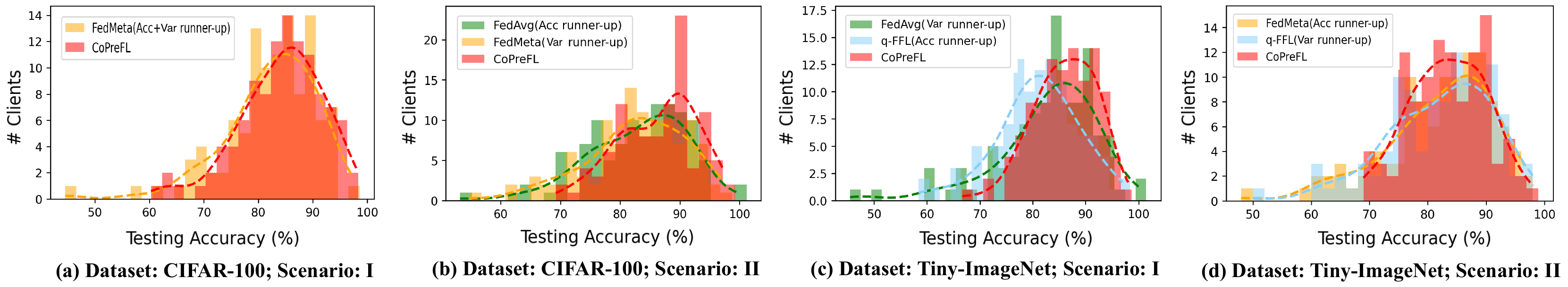}
\caption{\small Testing accuracy distributions in various non-IID FL tasks. {\system} achieves the best average accuracy (i.e., right-leaning distribution) and smaller performance variance (i.e., narrower distribution) while also improving worst-performing clients' accuracies.}
 \vspace{-1mm}
\label{fig:noniidpretrain_noniidfinetune_cifar}
 \end{figure*}

\vspace{-3mm}

\subsection{Experimental Results}
\label{sec:exp_result}
% \vspace{-3mm}

\textbf{Main results for scenarios I \& II.}
Tables \ref{tbl:sc1} and \ref{tbl:sc2} present the test accuracies for {\system} across {\tt scenarios} I and II on the CIFAR-100 and Tiny-ImageNet datasets. In {\tt scenario} I, {\system} shows robust initializations for downstream FL tasks, obtaining a  higher average accuracy and reduced performance variance across clients, and significantly improving accuracies for the worst-performing clients (Lowest 10-20\% metrics).
% This highlights the benefits of balancing the competing objective functions in (\ref{eq:ourobj})  during meta-updates. See Appendix D.1\ref{ap:sc1} for supplemental analyses and results that include varying the number of participants in pre-training, and different data distributions in downstream FL tasks. 
% In {\tt scenario} II, similarly,  {\system} consistently outperforms the baselines, in this case  by effectively utilizing server data in addition to balancing our objectives (\ref{eq:ourobj}). The benefit of a small server-side dataset, when available, can be seen through the performance improvements from Table \ref{tbl:sc1} to \ref{tbl:sc2}. Further, the improvement over \texttt{CoPreFL-SGD} suggests that conducting SGD iterations with server data in a centralized manner after meta-updating the global model might divert the pre-trained model away from our designed objectives. This emphasizes the importance of performing meta-learning on the server data following the partitioning outlined in Algorithm~\ref{alg:scenario2}. For further results on various configurations, including the effects of varying server dataset sizes, see Appendix D.2\ref{ap:sc2}.
% \color{black}
This highlights the benefits of balancing the competing objectives in (\ref{eq:ourobj})  during meta-updates. See Appendix~\ref{ap:sc1} for supplemental analyses and results, including varying the number of participants in pre-training, and different data distributions in downstream FL tasks. 
In {\tt scenario} II, {\system} also consistently outperforms baselines in this case by effectively utilizing server data in addition to balancing our objectives (\ref{eq:ourobj}).
The benefit of a small server-side dataset, when available, can be seen through the performance gains from Table \ref{tbl:sc1} to \ref{tbl:sc2}. Further, the improvement over \texttt{CoPreFL-SGD} suggests that conducting SGD with server data in a centralized manner after meta-updating the global model may divert the pre-trained model away from our designed objectives. This emphasizes the importance of performing meta-learning on the server data following the partitioning outlined in Algorithm~\ref{alg:scenario2}. For more results on various configurations, including impacts of varying server dataset sizes, see Appendix~\ref{ap:sc2}.

% perform as CoPreFL in scenario II.
% Comparing {\system} and \texttt{CoPreFL-SGD}

% kind of dilute the effect of our designed objective (3) and (4).
% comparing our-SGD and our-meta: this is the correct way to learn hybridly if we aim to solve obj funcs X X..

\textbf{Performance distribution comparison.} 
% Figure \ref{fig:noniidpretrain_noniidfinetune_cifar}  visualizes more detailed testing accuracy distributions of the final global model on each client in the downstream tasks.  
% We provide visualizations for our method, as well as the methods with the second-best average accuracy and the second-lowest variance from Table~\ref{tbl11111}.
% CoPreFL's distributions are narrower in each case, consistent with its reduction in performance variance across clients. Distribution shifts to the right (i.e., higher average accuracy) can also be observed.
% Importantly, we see that {\system} effectively shifts nearly all  low-performing clients on the left tail of the baselines towards the right, indicating enhanced accuracy for these clients.
% Additional results across various scenarios are provided in Appendix D.3\ref{ap:histogram}.
Figure \ref{fig:noniidpretrain_noniidfinetune_cifar} show the testing accuracy distributions of the final global model on each client in downstream tasks.  
We visualize results for our method and those with the second-best average accuracy and the second-lowest variance from Table~\ref{tbl11111}. 
{\system} shows consistently narrower distributions, reflecting reduced performance variance.
Distribution shifts to the right can also be found, indicating higher average accuracy.
Importantly, {\system} effectively shifts nearly all low-performing clients from the left tail of the baselines towards the right, enhancing accuracy for these clients.
More results for various scenarios are available in Appendix~\ref{ap:histogram}.

% \begin{wraptable}{r}{0.31\textwidth}

\begin{table}
\vspace{-4.5mm}

\small
\begin{center}
    
\scalebox{0.8}{

\begin{tabular}{c||cc}
\hline

% {\textbf{{\system}} (scenario: I, $m = 20$)} & \textbf{Acc} $\uparrow$ & \textbf{Variance} $\downarrow$ \\
\hline
{$\gamma$} & \textbf{Acc} $\uparrow$ & \textbf{Variance} $\downarrow$ \\
\hline\hline

0.0 & 83.11  $\pm$ 2.17 &	24.70  $\pm$ 1.95\\
0.25 & 84.04  $\pm$ 2.00 &	35.88  $\pm$ 2.59\\
0.5 & 85.23 $\pm$ 2.43 &	35.40 $\pm$ 2.75 \\
0.75 & 85.19  $\pm$ 2.38 &	39.31  $\pm$ 2.64\\
1.0 & 86.33	 $\pm$ 1.92 & 39.81  $\pm$ 2.30 \\
\hline

\end{tabular}
} 
 \vspace{-1mm}
\caption{\small Effect of  balancer $\gamma$ in {\tt scenario} I for Tiny-ImageNet.}%\yun{textbf 86.33 and 24.70?}}
%\caption{Performance of 10 non-IID downstream FL tasks on the Tiny-ImageNet dataset. These tasks are initialized using non-IID {\system} in scenario I, using different balancers.}
% \vspace{-0.1in}
\label{tb:ablation}
\end{center}
% \vspace{-5mm}
\end{table} 
% \end{wraptable} 
\textbf{Effect of balancer $\gamma$ in {\system}.}
% Table \ref{tb:ablation} gives the performance obtained from {\system} using different balancers $\gamma$ on Tiny-ImageNet.
% A larger $\gamma$ implies that the pre-trained model prioritizes the devices' average performance, whereas a smaller $\gamma$ prioritizes performance balance. 
% In Table~\ref{tb:ablation}, as the balancer $\gamma$ increases, we observe an increase in the average accuracy of downstream FL tasks but a higher variance, resulting in performance imbalance.
% The trend shows that {\system} allows control over the relative importance between accuracy and balanced performance  within its pre-training environment.
Table \ref{tb:ablation} presents the performance of {\system} on Tiny-ImageNet using different balancers $\gamma$.
A larger $\gamma$ implies that the pre-trained model prioritizes the devices' average performance, whereas a smaller $\gamma$ emphasizes performance balance. 
We see that increasing $\gamma$ lead to higher average accuracy in downstream FL tasks but also greater variance, indicating performance imbalance.
This trend shows that {\system} allows control over the relative importance between accuracy and balanced performance during pre-training.

\begin{table*}[!htb]
    \vspace{-2mm}

    \begin{subtable}[t]{0.4\linewidth}
      \centering
\scalebox{0.7}{
\begin{tabular}{c||cc}
\hline
\multicolumn{1}{c||}{\textbf{Pre-training} (Scenario I)} & \multicolumn{2}{c}{\textbf{Downstream:} Non-IID FedAvg}\\
\hline\hline
 {\textbf{Method}} & \textbf{Acc} $\uparrow$ & \textbf{Variance} $\downarrow$ \\
\hline

{Random Initialization} & 75.32 $\pm$ 1.68 & 41.39 $\pm$ 3.35 \\
{Centralized~\cite{nguyen2023where}} & 81.30  $\pm$ 2.92 & 69.44 $\pm$ 2.33\\
\hline
{SCAFFOLD~\cite{Karimireddy2019SCAFFOLDSC}} &  79.15  $\pm$ 3.08 & 57.84 $\pm$ 1.95 \\
{FedDyn~\cite{Acar2021FederatedLB}} & 81.23 $\pm$ 2.96 & 53.17 $\pm$ 2.85\\
{PerFedAvg~\cite{Fallah2020PersonalizedFL}} & 81.58  $\pm$ 1.83 & 49.73 $\pm$ 2.65\\
\systemnott & \textbf{83.29} $\pm$ 2.61 &	\textbf{34.69} $\pm$ 3.17 \\
\hline
\end{tabular}
}
\caption{Comparison with other initializations.\label{tbl:otherinit}} 
\vspace{-2mm}
    \end{subtable}%
    \begin{subtable}[t]{0.7\linewidth}
      \centering
\scalebox{0.7}{
\begin{tabular}{c||cc||cc}
\hline
% \multirow{2}{c||}{\textbf{Pre-training}(Scenario I)} & \multicolumn{4}{c}{Downstream}\\
\multirow{2}{*}{\begin{tabular}[c]{@{}c@{}} {\textbf{Pre-training}} \\ {(Scenario I)}\end{tabular}} & \multicolumn{4}{c}{\textbf{Downstream}: Non-IID FL}\\
\cline{2-5}
 & \multicolumn{2}{c||}{FedProx ($\mu$ = 1)}& \multicolumn{2}{c}{q-FFL (q = 2)}\\
\hline\hline
 {\textbf{Method}} & \textbf{Acc} $\uparrow$ & \textbf{Variance} $\downarrow$ & \textbf{Acc} $\uparrow$ & \textbf{Variance} $\downarrow$ \\
\hline
 {Centralized} & 82.39 $\pm$ 3.17 & 51.46  $\pm$ 2.59 & 79.26  $\pm$ 2.33 & 47.10 $\pm$ 3.05 \\
{FedAvg} & 79.53  $\pm$ 2.69 & 46.15  $\pm$ 3.04 & 79.53 $\pm$ 2.38 & 44.59  $\pm$ 2.95\\
{FedMeta} & 81.77  $\pm$ 3.29 & 63.12  $\pm$ 3.62 &  79.30 $\pm$ 3.02 & 39.63  $\pm$ 3.17 \\
{q-FFL} & 83.19  $\pm$ 3.03 & 52.12  $\pm$ 2.97  & 81.38 $\pm$ 2.67 & 37.27 $\pm$ 2.85 \\
\systemnott & \textbf{84.31}  $\pm$ 3.01 & \textbf{30.55}  $\pm$ 2.61 & \textbf{82.71}  $\pm$ 2.45 & \textbf{25.39}  $\pm$ 2.87\\
\hline
\end{tabular}
}
\caption{Integration with other downstream FL algorithms.\label{tbl:otherdown}}
\vspace{-2mm}
    \end{subtable} 
\caption{\small Results with (a) other initializations and (b) other downstream FL algorithms on CIFAR-100.}
\label{tb:subtable_other_initanddown}
% \vspace{-4mm}

\end{table*}

\textbf{Comparison with other initialization methods.}
In addition to the baselines in Table \ref{tbl:sc2} which consider performance balance or meta-learning, Table~\ref{tbl:otherinit} also evaluates  popular algorithms like SCAFFOLD~\cite{Karimireddy2019SCAFFOLDSC}, FedDyn~\cite{Acar2021FederatedLB}, and PerFedAvg~\cite{Fallah2020PersonalizedFL} providing pre-training for {\tt scenario} I.
%we also include SCAFFOLD~\cite{Karimireddy2019SCAFFOLDSC}, which addresses partial client sampling, FedDyn~\cite{Acar2021FederatedLB}, designed to tackle non-IID issues, and PerFedAvg~\cite{Fallah2020PersonalizedFL}, aiming to provide an adaptable personalized model, for a detailed comparison. 
We see that {\system} also outperforms these baselines, validating our meta-learning approach based on (\ref{eq:ourobj}).
These popular FL algorithms struggle with unseen task heterogeneity and performance balance. We also explore initializing downstream FedAvg with random weights or a centrally pre-trained model, a concept introduced in \citep{nguyen2023where}.
While the centralized method boosts downstream FL accuracy over random initialization, it introduces significant performance variance across clients, as it fails to mimic downstream FL characteristics.
More details along with additional results  can be found in Appendix~\ref{ap:otherinitial}.

\textbf{Compatibility with other downstream FL algorithms.}
% Apart from conducting downstream FL using FedAvg, we explored the applicability of our pre-training method to other FL algorithms, especially fairness-aware ones\yun{}. 
We next explore the ability of our pre-training method to enhance the performance of downstream FL algorithms other than FedAvg. For this, we consider FedProx~\cite{Sahu2018FederatedOI} and q-FFL~\cite{Li2020Fair}, more advanced FL algorithms that addresses heterogeneity and performance balance, in each federated downstream task. 
Table~\ref{tbl:otherdown} shows the results. Overall, we see that {\system} consistently achieves superiority in accuracy and variance compared to other pre-training baselines, when combined with different downstream FL algorithms. 
Details on the implementation and further discussions are provided in Appendix~\ref{ap:different_down}.

% \begin{wraptable}{r}{0.55\textwidth}

\begin{table}
 \vspace{-2mm}
\small
\begin{center}
\scalebox{1}{
 \scalebox{0.8}{\begin{tabular}{c||ccccc}
\hline
\multicolumn{1}{c||}{\textbf{Pre-training}} & \multicolumn{4}{c}{\textbf{Downstream:} Non-IID FedAvg}\\
\hline\hline
 {\textbf{Method}} & \textbf{Acc} $\uparrow$ & \textbf{Var} $\downarrow$ & \textbf{Lowest 10\%} & \textbf{Lowest 20\%} \\
\hline

% \multirow{4}{*}{IID} & {FedAvg} & 84.20 &	57.15 &		68.43 &	71.83 &	74.38\\
% & {FedMeta} & 83.80 &	42.64 &		72.30 & 73.79 &	75.47\\
% & {q-FFL} & 82.60 &	45.56 &	70.46 &	73.41 & 75.14\\
% & \systemnott (\gamma = 0.25) & \textbf{84.36} &	\textbf{38.56} &		\textbf{73.66} &	\textbf{75.63} &	\textbf{77.40}\\
% \hline

{Centralized} & 82.63 $\pm$ 2.63  & 63.57  $\pm$ 3.08 & 67.35  $\pm$ 2.57 & 69.22  $\pm$ 3.01 \\
 {FedAvg} & 80.19  $\pm$ 1.19 & 51.35  $\pm$ 2.44 & 68.72  $\pm$ 1.63 & 70.15  $\pm$ 1.45 \\
 {FedMeta} & 83.14  $\pm$ 2.07 & 39.85  $\pm$ 1.38 &	67.29 $\pm$  2.22 &	71.35   $\pm$ 2.53 \\
 {q-FFL} & 81.34  $\pm$ 1.91 & 47.98  $\pm$ 2.00 & 69.22  $\pm$ 1.85 & 70.35 $\pm$ 2.39  \\
 \systemnott %(\gamma = 0.75) 
 & \textbf{84.79}  $\pm$ 1.25 & \textbf{30.51}  $\pm$ 1.72  & \textbf{70.83}  $\pm$ 1.59 & \textbf{72.66}  $\pm$ 1.61 \\
\hline
\end{tabular}}
}
% \caption{Average performance across 10 non-IID downstream FL tasks, initialized with various FL pre-trained methods using 20 out of 100 distributed participants, on the CIFAR-100 dataset.\yun{revmove m}}
% \vspace{-1.5mm}
 \vspace{-1mm}
\caption{\small Results with both seen/unseen classes during downstream FL, using the CIFAR-100 dataset in {\tt scenario} I.%Average performance across 10 non-IID downstream FL tasks, initialized with centralized model and various non-IID FL pre-trained models, encompassing both seen and unseen classes. \yun{can use ACC and Var for spaces.}}
}
% \vspace{-1.5pc}
\label{tb:rebuttal_seen_unseen}
\end{center}
 \vspace{-2mm}
 \end{table}
 % \end{wraptable}

\textbf{Both unseen/seen classes in downstream FL tasks.}
In addition to the setting without overlapping classes between pre-training and downstream tasks, we explore a mixed scenario where clients in the downstream FL also hold ``seen classes.'' 
To implement this, we use CIFAR-100 and randomly sampled 10 classes from the pre-training dataset and 10 classes from the downstream dataset, resulting in  10 seen and 10 unseen downstream classes. 
From this, we constructed 10 downstream tasks by randomly selecting 5 classes in each case, to conduct non-IID downstream FedAvg. 
The pre-trained models were solely trained on the original 80 classes of CIFAR-100. 
Table~\ref{tb:rebuttal_seen_unseen} shows the results. We see that the accuracies are generally higher compared to those in Table~\ref{tbl:sc1}, as the downstream tasks involve classes seen during pre-training. Once again, the improvements in each metric confirm  the advantage of {\system}.

%Also, the observed trends compared with the baselines further align with the outcomes for only unseen classes, confirming the advantage of {\system}.

% \begin{wraptable}{r}{0.49\textwidth}
% \vspace{-4mm}
\begin{table}
\begin{center}
    
\scalebox{0.85}{

\begin{tabular}{c||cc}
\hline
\multicolumn{1}{c||}{\textbf{Pre-training}} & \multicolumn{2}{c}{\textbf{Downstream:} Non-IID FedAvg}\\
\hline\hline
{\textbf{Method}} & \textbf{Acc} $\uparrow$ & \textbf{Variance} $\downarrow$ \\
\hline

 {Centralized} & 86.75   $\pm$  2.89 & 67.34   $\pm$ 2.17  \\
\systemnott &\textbf{87.96}   $\pm$ 1.95 & \textbf{30.79}   $\pm$ 2.79  \\
% {IMAGENET1K\_V1} & 89.64 &	65.62 \\
\hline

\end{tabular}
}
 \vspace{-1mm}
\caption{\small Results with centrally stored dataset. ImageNet is used for pre-training, while CIFAR-100 is used for downstream FL.}%Average performance of non-IID FedAvg tasks, initialized by different pre-trained models. Note that ImageNet is used for pre-training, while CIFAR-100 is used for downstream FL.}
\label{tb:rebuttal_imagenet}
\end{center}
 \end{table}
 % \vspace{-5mm}
  % \end{wraptable}

\textbf{Application to centrally stored public dataset.}
% We further consider the applicability of {\system} when  the pre-training dataset is centrally stored, as discussed in \textbf{Remark 2}.
% To achieve this, we performed pre-training using the ImageNet\_1K dataset and use FedAvg with CIFAR-100  as the downstream task. We  intentionally split the dataset according to {\tt scenario} I to mimic the distributed nature of downstream FL.
% The results are shown in Table~\ref{tb:rebuttal_imagenet}, where we compare against standard centralized pre-training. The superiority in average accuracy and balanced performance achieved by {\system}   confirms that CoPreFL is advantageous even when a large public dataset is used for pre-training.  Detailed implementations and additional results can be found in Appendix D.7\ref{ap:largescale}. 
We also explore the applicability of {\system} with centrally stored pre-training data, as detailed in \textbf{Remark 2}. We conducted pre-training using the ImageNet\_1K dataset and use FedAvg with CIFAR-100 as the downstream task. We intentionally split the dataset according to {\tt scenario} I to mimic the distributed nature of downstream FL. Results in Table~\ref{tb:rebuttal_imagenet} show that {\system} outperforms standard centralized pre-training in both average accuracy and balanced performance, showing {\system}'s advantage even when a large public dataset is used for pre-training. Further details and additional results are available in Appendix~\ref{ap:largescale}.

\begin{table}
% \vspace{-5mm}
\begin{center}
    
\scalebox{0.8}{

\begin{tabular}{c||cc}
\hline
\multicolumn{1}{c||}{\textbf{Pre-training}} & \multicolumn{2}{c}{\textbf{Downstream:} Non-IID FedAvg}\\
\hline\hline
{\textbf{Method}} & \textbf{Acc} $\uparrow$ & \textbf{Variance} $\downarrow$ \\
\hline
FedAvg & 62.23 $\pm$ 2.65 & 51.29  $\pm$ 3.11 \\
FedMeta  & 64.35  $\pm$ 3.07 & 44.38 $\pm$ 2.98 \\
q-FFL	& 60.79	 $\pm$ 3.15 & 27.96  $\pm$ 3.03\\
\systemnott & \textbf{66.83}	 $\pm$ 2.85 & \textbf{24.31} $\pm$ 2.83\\
\hline

\end{tabular}
}
 \vspace{-2mm}
\caption{\small Results in the domain shift scenario using PACS dataset.  %PACS dataset is used to model data discrepancy between pre-training and downstream stages. 
}
\label{tb:PACS_main}
\end{center}
 % \vspace{-5mm}
  \end{table}

\textbf{Comparison under domain shifts.} 
% Our experiments so far have focused on the robustness of the pre-trained model to unseen labels.  We now explore {\system}'s capability to handle unseen dataset domains during downstream tasks, using the PACS dataset. In Table~\ref{tb:PACS_main}, we use 3 domains (Art, Cartoon, and Photo) for pre-training in {\tt scenario} I and conduct downstream FedAvg using the remaining Sketch domain, distributing samples across clients as in Table \ref{tbl11111}.  
%  These results show that CoPreFL's improvements extend to cases when facing arbitrary downstream FL tasks with domain distributions  that are distinct from those in the pre-training phase. This highlights {\system}'s ability to provide robust initializations when confronted with several different types of downstream data heterogeneity.  We provide more detailed settings and results in Appendix D\ref{ap:sc1}.
Our experiments so far have focused on the robustness of the pre-trained model to unseen labels. We now explore {\system}'s ability to handle unseen data domains during downstream tasks using the PACS dataset. In Table~\ref{tb:PACS_main}, we use 3 domains (Art, Cartoon, and Photo) for pre-training in {\tt scenario} I and conduct downstream FedAvg using the remaining Sketch domain, distributing samples across clients as in Table \ref{tbl11111}.  
Results show that {\system} effectively handles domain distributions in downstream FL tasks that differ from the pre-training phase. This highlights {\system}'s ability to provide robust initializations amid various types of downstream data heterogeneity. Additional settings and results are available in Appendix~\ref{ap:sc1}.

% \textbf{Other experimental results.} Additional details and results on varying degrees of non-IIDness and participant ratios during pre-training for both scenarios, the effect of server's data size, domain shift experiments, and the use of centrally stored public datasets for pre-training can be found in Appendix~\ref{ap:add_exp}.

\vspace{-2.5mm}
\section{Conclusion}
We presented {\system}, a collaborative pre-training method that provides a robust model initialization for an arbitrary set of  downstream FL tasks. %aimed at providing any unseen FL task with a robust initialization.
%Our pre-training approach takes into account practical scenarios where  data may be distributed across the clients and the server during pre-training.
{\system} leverages meta-learning to equip the pre-trained model with the ability to handle different forms of data heterogeneity that manifest in downstream FL, while balancing between average performance and variance across clients.
We developed {\system} for different distributed pre-training scenarios, and showed its benefit even for centrally stored public data.
%during the pre-training phase. 
Extensive experiments demonstrated the advantages of {\system} compared with several baselines methods, for a multitude of settings capturing statistical heterogeneity in downstream FL.
%including unseen/seen class labels, different downstream FL algorithms, varying pre-training data locations, and domain distribution shifts.

% \section*{Impact Statements}
% This paper presents a pre-training method with potential applications in various AI domains, including natural language processing and computer vision. It is essential to recognize and address potential ethical and privacy concerns associated with the pre-training dataset. For instance, considerations should be made for privacy issues related to images containing human faces and ethical concerns regarding toxic texts. By acknowledging and mitigating any such issues that arise, we can further promote responsible and ethical advancement of AI/ML technologies.

\bibliography{ref}

\clearpage
\appendix
\section{Key Applications}\label{ap:application}
Consider a healthcare application where each client, such as a hospital or an individual patient, aims to build a comprehensive global model capable of classifying a wide range of diseases. However, individual clients may possess limited types of diseases in their local datasets – for instance, one client may have data on diseases A and B but lacks information on diseases C and D. In this context, federated learning becomes essential. Clients need to collaborate to construct a global model that not only reflects the diseases available locally but also incorporates information about diseases not present in their individual datasets, ensuring a more robust and universally applicable healthcare model. 
Similarly, in the domain of autonomous vehicles, each self-driving car may strive to develop a global model for scenario detection in various weather conditions. However, individual cars might encounter limited weather scenarios locally – one car might navigate through a desert environment, while another faces challenges in a snowy storm. Through federated learning, these cars can collectively construct a global model that accounts for a broad spectrum of weather conditions, ensuring robust scenario detection capabilities for all vehicles involved.

As noted in Remark 2, the server can intentionally partition the centralized dataset and implement our scheme, utilizing multiple computing units available at the server, to obtain a pre-trained model. The advantage of this approach, compared to simple centralized training, lies in mitigating side effects such as performance biases and the substantial variance associated with centralized training. This phenomenon stems from the lack of generalizability in the model's design. When a model undergoes pre-training in a centralized manner based on SGD, it becomes rigidly bound to the knowledge in the pre-training dataset. This fixation presents a challenge in adapting the model to the diverse clients that may possess new or unseen data in downstream tasks. Such variations can arise from factors like the time-varying environment or new clients joining the system, as exemplified in the aforementioned applications: classifying different scenarios based on the self-driving car’s environment, identifying diverse diseases based on patient interests, or enabling face/speech recognition for new phone users.

In our experimental comparison, we consider a FL baseline, FedMeta~\cite{Chen2018FederatedMW}, which also incorporates meta-learning.
We would like to emphasize that the distinction between our {\system} and FedMeta lies in the application, which requires us to conduct meta-update in a totally different way. 
FedMeta is designed to offer personalization to individual clients, e.g., when a specific client is interested in predicting only diseases A and B, or when a specific self-driving car is interested in the model tailored to a specific weather.
In contrast, our emphasis is on creating an initial model that can construct a good ``global model'' during downstream instead of ``personalized models'', targeting the aforementioned applications. 
This is the reason why we need to update the temporary global model instead of the local models, which is the key technical difference with FedMeta.
Another technical difference is the consideration of performance balance in our method. 
These two key techniques enables {\system} to construct a robust initial model that can quickly adapt to ``any group of clients'' (instead of individual clients) to construct a global model during downstream tasks.

\section{Detailed Procedure for {\system} in Scenario II}
\label{sec:app_detailscenarioII}
This section provides a detailed introduction to our {\system} in \texttt{scenario} II, as discussed in Section~\ref{sec:method_sec2}.
Similar to the goals of {\system} in \texttt{scenario} II, we still aim to balance between the objective functions in (\ref{eq:ourobj}), but in this scenario, the data used to perform meta-updates and control our objectives is different.
During each federated round $t$ in the pre-training phase, participants download the global model $\Phi^{t-1}$ from the previous round (line 6 in Algorithm~\ref{alg:scenario2}). 
Subsequently, they perform few local training iterations utilizing their respective local datasets $D^p_j$ (line 8 in Algorithm~\ref{alg:scenario2}). 
This process leads to a training loss $\ell^{e}_{D^p_j}(\Phi^{t})$ for each local epoch $e$, defined as $\frac{1}{|D^p_j|} \sum_{(x,y)\in D^p_j} \ell(\Phi^{t, e}_j(x), y)$, where $x$ represents the input (e.g., images), $y$ denotes the true label, and $\ell(\cdot)$ denotes the loss function (e.g.,cross-entropy loss).
After all participants finish $E$ epochs, we obtain the local model $\Phi^{t,E}_j$. 
Upon the completion of local training by all participants, participants' local models are transmitted to the server (line 12 in Algorithm~\ref{alg:scenario2}), and the server aggregates these models into a temporary global model $\overline{\Phi^t} = \sum_{j \in m} \mu_j \Phi^{t,E}_j$, which is weighted by relative dataset sizes $\mu_j =\frac{|D^p_j|}{\sum_{i \in m}|D^p_i|}$.

We then perform meta-updates on the temporary global model $\overline{\Phi^t}$ using server's dataset $D^s$.
To start, we first randomly divide the server's dataset $D^s$ into $|m|$ equal partitions. 
Instead of equal partitioning, one can also divide the server-side dataset into partitions with unequal sizes and use them for meta-update. 
However, during pre-training, the server does not know the dataset sizes of clients in future downstream tasks.
In this case, one intuitive way is to treat all clients equally/fairly, by meta-updating the model with the same query set sizes. We show that this equal partitioning provides significant performance improvements as can be seen in our experiments.
Subsequently, the server evaluates the temporary global model $\overline{\Phi^t}$ using each subset $D^s_j$ (line 14 in Algorithm~\ref{alg:scenario2}), resulting in the corresponding gradient and loss $\mathcal{L}_{D^s_j}(\overline{\Phi^t})$ = $\frac{1}{|D^s_j|}$ $\sum_{(x,y) \in D^s_j}{} \ell(\overline{\Phi^t}(x), y)$. 
The collective server's loss, denoted as $\mathcal{L}_{D^s}(\overline{\Phi^t})$ in line 16 of Algorithm~\ref{alg:scenario2}, is determined by aggregating all the collected loss values obtained from $D^s_j$, and we also calculate the variance $\sigma_{D^s}^2 = \frac{1}{m} \sum_{i \in m} (\mathcal{L}_{D^s_i}(\overline{\Phi^t}) - \frac{1}{m}\mathcal{L}_{D^s}(\overline{\Phi^t}))$ across server's losses to examine the performance distribution.
We then tailor a customized server meta-loss $\mathcal{L}_{meta}(\overline{\Phi^t}) = \gamma \mathcal{L}_{D^s}(\overline{\Phi^t}) + (1 - \gamma) \sigma_{D^s}^2(\overline{\Phi^t})$ to achieve a balance between optimizing for performance and performance balance. 
Finally, in line 19 of Algorithm~\ref{alg:scenario2}, we employ the customized server meta-loss $\mathcal{L}_{meta}(\overline{\Phi^t})$ and the aggregated gradient gathered from the server's subsets to update the temporary global model $\overline{\Phi^t}$, aligning it with our controlled objective. 
The server then sends this meta-updated global model $\Phi^t$ to the participants in the next round for initialization.
After completing $T$ federated rounds, we regard the final global model $\Phi^T$ as the pre-trained model in \texttt{scenario} II, which serves as the initialization for the downstream FL tasks.

\begin{algorithm*}[t]
\caption{Our Pre-training Method {\system} (Pre-training Phase in {\tt Scenario} II)}
\label{alg:scenario2}
\begin{algorithmic}[1]

\STATE \textbf{Input:} $M$ clients in the pre-training phase, with each client $i$ holding their own dataset $D^p_i$; the server also holds a dataset $D^s$.

% \State Split $D^C$ to $M$ clients based on IID or non-IID distribution, and select $m \in M$ participants.
% \State Each participant $j \in m$ has their own dataset $D_j$ without splitting into support and query sets
\FOR {Each communication round $t = 1, 2, ..., T$}
\STATE Randomly select a set of client $m \subset M$ to participate in learning
\STATE Randomly split server's dataset $D^s$ into $|m|$ subsets
\FOR {Each participant $j$ in parallel}
\STATE Downloads $\Phi^{t-1}$ from the server

\FOR{local epoch $e = 1, 2, ... E$}
\STATE $\ell^{e}_{D^p_j}(\Phi^{t}) \gets \frac{1}{|D^p_j|} \sum_{(x,y)\in D^p_j} \ell(\Phi^{t, e}_j(x), y)$ \hfill\COMMENT{  Compute local support loss $\ell^{e}_{D^p_j}$}
\STATE $\Phi^{t,e}_{j}$ $\gets$ $\Phi^{t,e-1}_j$ $-$ $\eta \nabla \ell^{e}_{D^p_j}(\Phi^t)$ \hfill\COMMENT{ Perform SGD local update using support loss}

\ENDFOR
% \STATE $\mathcal{L}_{D^p_j}(\Phi^t)$ $\gets$ $\frac{1}{E} $\sum_{e = 1}^{E}\ell^{e}_{D^p_j}(\Phi^t)$ \hfill\Comment{// Get averaged support loss}

% \STATE $\mathcal{L}_{D^p_j}(\Phi^t)$ $\gets$ $\frac{1}{|D^p_j|}$ $\sum_{(x,y) \in D^p_j}{} \ell(\Phi^{t-1}(x), y)$ \hfill\Comment{// Get local loss using local dataset $D^p_j$}

% \STATE $\Phi^t_{j}$ $\gets$ $\Phi^{t-1}$ $-$ $\eta \nabla \mathcal{L}_{D^p_j}(\Phi^t)$ \hfill\Comment{// Perform SGD local update using averaged support loss $\mathcal{L}_{D^p_j}$}

\ENDFOR
\STATE $\overline{\Phi^t}$ $\gets$ $\sum_{j \in m} \frac{|D^p_j|}{\sum_{i \in m}|D^p_i|} \Phi^{t,E}_j$  \hfill\COMMENT{ Model aggregation to construct temporary global model}

% \State Server executes updates using split $D^S$

\FOR {Each split server's dataset $D^s_j$ in parallel, Server}
% \State Initialize each local model $\phi_j$ with $\overline{\phi^r}$

% \State $\mathcal{L}_{Q_j}(\overline{\phi^t})$ $\gets$ $\frac{1}{|Q_j|}$ $\sum_{(x,y) \in Q_j}{} \ell(\overline{\phi^t}(x), y)$ \Comment{Local meta-loss using query set $Q_j$}

\STATE $\mathcal{L}_{D^s_j}(\overline{\Phi^t})$ $\gets$ $\frac{1}{|D^s_j|}$ $\sum_{(x,y) \in D^s_j}{} \ell(\overline{\Phi^t}(x), y)$ \hfill\COMMENT{ Server's loss corresponding to each partition}
\ENDFOR
\STATE Overall meta-loss on server: $\mathcal{L}_{D^s}(\overline{\Phi^t}) = \sum_{j \in m} \mathcal{L}_{D^s_j}(\overline{\Phi^t})$
\STATE Variance across server meta-losses: $\sigma_{D^s}^2(\overline{\Phi^t}) = \frac{1}{|m|} \sum_{j \in m} (\mathcal{L}_{D^s_j}(\overline{\Phi^t}) - \frac{1}{|m|}\mathcal{L}_{D^s}(\overline{\Phi^t}))^2$

\STATE Customized server meta-loss: $\mathcal{L}_{meta}(\overline{\Phi^t}) = \gamma \mathcal{L}_{D^s}(\overline{\Phi^t}) + (1 - \gamma) \sigma_{D^s}^2(\overline{\Phi^t})$

\STATE $\Phi^t$ $\gets$ $\overline{\Phi^t}$ $-$ $\zeta \nabla \mathcal{L}_{meta}(\overline{\Phi^t})$ \hfill\COMMENT{ Model meta-updates using customized loss}
\ENDFOR

\STATE \textbf{Output:} A pre-trained model for downstream FL tasks: $\Phi^T$
\end{algorithmic}
\end{algorithm*}
% \vspace{-\baselineskip} 

\section{Detailed Settings for Datasets and Hyperparameters}
\subsection{Dataset Details}
\label{ap:dataset}

In a practical scenario where labels for downstream tasks are unknown during pre-training, we split the dataset based on classes.
For CIFAR-100 (100 classes with 600 images per class), 80 classes are used for pre-training and 20 classes for downstream FL tasks, resulting in 48,000 images for pre-training and 12,000 images for downstream tasks.
Similarly, for Tiny-ImageNet (200 classes with 600 images per class), 160 classes are employed for pre-training and 40 classes for downstream FL tasks, providing 96,000 images for pre-training and 24,000 images for downstream tasks.
We randomly select 95\% of the pre-training dataset samples for clients, while the remaining 5\% of samples for the server.
Specifically, for CIFAR-100, this results in 45,600 images for clients and 2,400 images for the server.
For Tiny-ImageNet, we allocate 91,200 images for clients and 4,800 images for the server.
We will distribute 45,600 and 91,200 images to $|M| = 100$ clients based on IID or non-IID data distribution for CIFAR-100 and Tiny-ImageNet, respectively. 
Each client further divides its local data into 80\% support samples and 20\% query samples.

For the downstream phase samples (12,000 images for CIFAR-100 and 24,000 images for Tiny-ImageNet), we randomly select 5 classes from the available downstream classes (20 classes for CIFAR-100 and 40 classes for Tiny-ImageNet) to form a single FL task.
This results in 3,000 images (5 classes with 600 images per class) for each FL task in both CIFAR-100 and Tiny-ImageNet datasets.
These 3,000 images are distributed to $|G| = 10$ clients based on either IID or non-IID data distribution.
Within each client, 80\% of local data is used for FL training, while the remaining 20\% is reserved to evaluate the final global model.

In the IID setup, data samples from each class are distributed equally to $|M| = 100$ clients for pre-training and $|G| = 10$ clients for downstream FL task.
Taking the CIFAR-100 dataset in the IID pre-training phase as an example, there are $|M| = 100$ clients, each holding 456 images.
We will randomly select $|m|$ clients to participate in pre-training in each round.
In the IID downstream phase of CIFAR-100, there are $|G| = 10$ clients, each holding 300 images.
In the non-IID setup, samples within each class are partitioned among $|M|$ and $|G|$ clients using a Dirichlet($\alpha$) distribution for pre-training and downstream task, respectively, with $\alpha = 0.5$ selected as is in the literature~\citep{Morafah2022RethinkingDH,Li2021ModelContrastiveFL}.

In addition to the label distributional shifts discussed in our manuscript, we also explore our method's capability to handle domain  shifts. To assess this, we conducted an additional experiment using the PACS dataset~\cite{Li2017DeeperBA}. 
% Specifically, we used the art painting, cartoon, and photo data domains for pre-training and performed downstream non-IID FedAvg training using the sketch data domain.
As PACS dataset consists of four domains (Art, Cartoon, Photo, Sketch), we separate them into three domains for pre-training and one domain for downstream FL, following the one-domain-leave-out setup~\cite{Li2022UncertaintyMF,Zhou2021DomainGW}.

\subsection{Hyperparameters and Compute Settings}
\label{ap:parameter}
We use ResNet-18 as the model structure for image classification, following the setting of~\cite{nguyen2023where,chen2023on}.
For our method, the SGD optimizer with a learning rate of $\eta = 10^{-3}$ and $\zeta = 10^{-3}$ is adopted for both local and meta updates. 
Both local and meta learning rates are searched within the range of [1e-2, 5e-3, 1e-3, 5e-4].
We searched for learning rates within the range of [1e-2, 5e-3, 1e-3, 5e-4] for local training of all FL pre-training baselines and selected 1e-3 as the optimal learning rate for them. 
In \texttt{scenario} II, each FL baseline will continue to conduct a few SGD iterations using the server's data after constructing their global model. We searched for learning rates in the range of [1e-2, 1e-3] for this additional training and selected 1e-3 as the optimal learning rate for the server.
% In scenario II, we search for learning rates in the range of [1e-2 and 1e-3] when further training the global model of each baseline using server's data and select 1e-3 as the optimal learning rate for the server.
Regarding hyperparameters in the q-FFL baseline, we conducted experiments with q-values of 1, 3, and 5 and reported the corresponding best statistics.
We select a learning rate $\eta$ from the range [1e-2, 5e-3, 1e-3, 5e-4] for local updates in our {\system} and determined that 1e-3 provides the best results.
Additionally, for meta-updates in both scenarios, we search for the learning rate $\zeta$ within the range [1e-2, 1e-3] and find that 1e-3 is the optimal value.
In the case of the centralized baseline mentioned in Section~\ref{sec:exp_result}, we searched for the optimal learning rate within the range [1e-2, 5e-3, 1e-3, 5e-4, 1e-4], ultimately selecting 1e-3.
We utilized the SGD optimizer for all updates across all methods, and the batch size is set to be 32 for all experiments.

For the pre-training phase, we set the number of rounds to $T = 50$, and each round of local training takes $E = 5$ epochs for each client. For downstream FL tasks, we employ the widely used FedAvg algorithm to isolate the effects of different FL approaches and focus specifically on the impact of different initializations.
We consider $R = 50$ FL rounds using the training set, involving 5 iterations per round for local training using the SGD optimizer with a learning rate of $10^{-3}$. 
For the settings of other downstream FL algorithms, see Appendix~\ref{ap:different_down}.
In our simulations of {\system}, we assessed various balancer values $\gamma$ from the range [0.0, 0.25, 0.5, 0.75, 1.0] in all scenarios during pre-training.
For evaluation of downstream FL, we report the best-performing (highest average accuracy) value in our paper.
For fair selection/comparison, we also report the results of other baselines with their own best accuracy when searching for hyperparameters.
% We conduct simulations of our {\system} using each balancer $\gamma$ within the range of [0.0, 0.25, 0.5, 0.75, 1.0] in all scenarios and report the best-performing value in our paper.
We run all experiments on a 3-GPU cluster of Tesla V100 GPUs, with each GPU having 32GB of memory.

% \subsection{Details for Downstream Task}
% \label{ap:downstream}
% To generate each downstream FL task, we randomly select 5 classes out of the 20 classes from the CIFAR-100 dataset and 40 classes from the Tiny-ImageNet dataset, and distribute the corresponding data samples to $|G|=10$ clients following either IID or non-IID data distributions. 
% It is important to note that these classes (20 and 40 classes) are distinct from those used in the pre-training phase.  
% Each participant in the downstream phase utilizes 80\% of its local data as training samples, while the remaining 20\% is reserved for testing samples. 
% To see the impact of different pre-training methods, we fix the training procedure at each downstream task to the commonly adopted FedAvg algorithm. 
% We consider $R = 50$ FL rounds using the training set, involving 5 iterations per round for local training using the SGD optimizer with a learning rate of $10^{-3}$. 
% We evaluate the final global model's performance using testing samples from each participant and report both the accuracy and the variance of the accuracy distribution across $|G|$ clients for each FL task.
% We consider a total of $X = 10$ downstream FL tasks, and the evaluation metrics are reported as the average across these $X$ downstream FL tasks.

\section{Additional Experiments and Analyses}
\label{ap:add_exp}
\subsection{Downstream FL Results with Scenario I Pre-training}
\label{ap:sc1}
\textbf{Additional results with varying degrees of IIDness and numbers of participants.}
This section provides supplementary results for pre-training \texttt{scenario} I, as discussed in Section~\ref{sec:exp_result}.
We train pre-trained models using both IID and non-IID distributions, varying the number of participants in each federated round during the pre-training phase. 
To be more specific, we specify the number of participants $|m|$ as 15, 20, 25, and 30 out of 100 clients to participate in FL during the pre-training phase.
Subsequently, we evaluate these pre-trained models by initializing them for IID and non-IID downstream FL tasks.
Tables~\ref{tb:cfiar_15_iidfinetune},  \ref{tb:cfiar_20_iidfinetune}, \ref{tb:cfiar_25_iidfinetune}, and \ref{tb:cfiar_30_iidfinetune} display the average performance across 10 \textbf{IID} FL downstream tasks and Tables~\ref{tb:cfiar_15_noniidfinetune},  \ref{tb:cfiar_20_noniidfinetune}, \ref{tb:cfiar_25_noniidfinetune}, and \ref{tb:cfiar_30_noniidfinetune} show the average performance across 10 \textbf{non-IID} FL downstream tasks. 
In both cases, the downstream FL  were initialized by pre-trained models trained on 15, 20, 25, and 30 participants out of 100 clients, respectively, on the CIFAR-100 dataset.
For the Tiny-ImageNet dataset, Tables~\ref{tb:tiny_15_iidfinetune},  \ref{tb:tiny_20_iidfinetune}, \ref{tb:tiny_25_iidfinetune}, and \ref{tb:tiny_30_iidfinetune} show the average performance across 10 \textbf{IID} FL downstream tasks and Tables~\ref{tb:tiny_15_noniidfinetune}, \ref{tb:tiny_20_noniidfinetune}, \ref{tb:tiny_25_noniidfinetune}, and \ref{tb:tiny_30_noniidfinetune} display the average performance across 10 \textbf{non-IID} FL downstream tasks.
In both cases, the downstream FL  were also initialized by pre-trained models trained on 15, 20, 25, and 30 participants out of 100 clients, respectively.
% \yun{across these experiment results considering different downstream FL tasks and different dataset, .....}
Across these experimental results, considering different data distribution setup during the pre-training phase and different datasets, our {\system} consistently demonstrates superiority over the baseline when used as an initialization for various downstream FL tasks.
By creating an environment that mimics downstream FL tasks and specifically addressing the challenges encountered in these tasks, our designed pre-training objectives in (\ref{eq:ourobj}) establish an ideal pre-trained model for FL.
As initialization for various unseen FL tasks, our {\system} provide downstream FL tasks with both better average performance and balanced predictions across clients.

\textbf{Comparison of IID and non-IID downstream tasks.} In some cases, when comparing IID and non-IID downstream tasks, we see that data heterogeneity has only a slight impact on the proposed approach. To be more specific, the performance of non-IID downstream FL experiences only a slight performance drop compared to IID downstream FL. In an IID setting, the model must classify all classes present in the system, as each client will have most classes in its local dataset. However, in a non-IID downstream scenario, the model may only need to classify a few classes since each client typically has a limited subset of classes. This simplification of local tasks in the non-IID setting reduces the complexity of evaluation for each client, minimizing the impact of data heterogeneity on the overall federated learning process. This phenomenon aligns with findings from other FL studies, where global model performance is assessed using local test sets from individual clients~\cite{Li2020Fair,Diao2023ExploitingLS}.

\textbf{Varying the number of downstream tasks.} To model various unseen downstream scenarios, we conduct 5-way classification during downstream FL (i.e., sampling 5 classes from 20 in CIFAR-100 downstream dataset to conduct one task.) The goal is to design a ``generalized initial model'' that can adapt to arbitrary downstream tasks that potentially contain unseen classes and to evaluate the versatility of the pre-trained model in providing a robust starting point for various scenarios. We also consider a ``single downstream scenario'' with 20-way classification using all downstream classes (that have not appeared during pre-training). 
To be more specific, we distribute the entire downstream dataset of CIFAR-100 among $|G| = 10$ clients based on non-IID  distribution and performed FedAvg.
The results are provided in Table~\ref{tb:20way}, indicating that {\system} still performs better than other initialization baselines.

\textbf{Domain shifts.}
 Each column in Table~\ref{tb:PACS} represents the domain for downstream FL in a one-domain-leave-out setting. For example, in the first column, we used the art painting, cartoon, and photo data domains for pre-training and performed downstream non-IID FedAvg training using the sketch data domain.
We followed the same experimental settings as described in Table~\ref{tbl:sc1}, considering \texttt{scenario} I, with pre-training data distributed to 100 clients based on a non-IID ($\alpha = 0.5$) distribution, and selected 20 participants for each pre-training round.
Table~\ref{tb:PACS} presents the results for each pre-trained method using the PACS dataset. 
We observed that our pre-trained method {\system} outperforms other pre-training methods in scenarios involving domain shifts, both in terms of performance accuracy and performance balance.

\subsection{Downstream FL Results with Scenario II Pre-training}
\label{ap:sc2}
\textbf{Additional results with varying degrees of IIDness and numbers of participants.} This section provides supplementary results for \texttt{scenario} II, where the server holds a small portion of the dataset during the pre-training phase.
We also consider varying numbers of participants $|m|$, specifically 15, 20, 25, and 30 out of 100 clients, during the pre-training phase for these models. 
Tables~\ref{tb:cfiar_15_iidfinetune_sc2},  \ref{tb:cfiar_20_iidfinetune_sc2}, \ref{tb:cfiar_25_iidfinetune_sc2}, and \ref{tb:cfiar_30_iidfinetune_sc2} display the average performance across 10 \textbf{IID} FL downstream tasks and Tables~\ref{tb:cfiar_15_noniidfinetune_sc2},  \ref{tb:cfiar_20_noniidfinetune_sc2}, \ref{tb:cfiar_25_noniidfinetune_sc2}, and \ref{tb:cfiar_30_noniidfinetune_sc2} show the average performance across 10 \textbf{non-IID} FL downstream tasks. 
In both cases, the downstream FL  were initialized by pre-trained models trained on 15, 20, 25, and 30 participants out of 100 clients, respectively, on the CIFAR-100 dataset.
For the Tiny-ImageNet dataset, Tables~\ref{tb:tiny_15_iidfinetune_sc2},  \ref{tb:tiny_20_iidfinetune_sc2}, \ref{tb:tiny_25_iidfinetune_sc2}, and \ref{tb:tiny_30_iidfinetune_sc2} show the average performance across 10 \textbf{IID} FL downstream tasks and Tables~\ref{tb:tiny_15_noniidfinetune_sc2}, \ref{tb:tiny_20_noniidfinetune_sc2}, \ref{tb:tiny_25_noniidfinetune_sc2}, and \ref{tb:tiny_30_noniidfinetune_sc2} display the average performance across 10 \textbf{non-IID} FL downstream tasks.
In both cases, the downstream FL  were also initialized by pre-trained models trained on 15, 20, 25, and 30 participants out of 100 clients, respectively.
It is important to note that in this scenario, FedAvg, FedMeta, and q-FFL undergo further training using server data through the SGD optimizer after each method completes its local iterations and obtains its respective global model in each round~\citep{Yang2023OnTC, Bian2023AcceleratingHF}. 
Similarly, \texttt{CoPreFL-SGD} is trained using server data with the SGD optimizer on $\Phi^t$ in line 17 of Algorithm \ref{alg:scenario1} in each round. This process involves conducting meta-updates and balancing performance and variance using clients' data first, followed by updating the aggregate model again using the server's dataset.
Finally, {\system} follows Algorithm \ref{alg:scenario2}, utilizing server data for meta-updates.
By incorporating meta-updates using server data to align with our objectives in (\ref{eq:ourobj}), our pre-training method consistently outperforms other baselines, leading to improved average accuracy and reduced variance.
Comparing {\system} with {\system}-\texttt{SGD} strongly suggests that, rather than conducting a few SGD iterations using server data, which may dilute our objectives, we recommend building pre-training objectives upon server data using meta-updates.

\textbf{Alternative implementation for baselines in scenario II.} In addition to the hybrid training approach introduced in~\cite{Yang2023OnTC, Bian2023AcceleratingHF}, which utilizes clients' data ($\bigcup_{i \in M} D^p_i$) to train local models and then refines the aggregated global model using the server's data ($D^S$), we explore an alternative implementation for other FL baselines in \texttt{scenario} II.
In this case, we distribute the entire training dataset ($\bigcup_{i \in M} D^p_i + D^S$) to $|M| = 100$ clients and select $|m| = 20$ participants in each round for federated learning without a further refining step since there is no server's data in this case. 
Therefore, each client holds more samples compared to their previous scheme. 
Table~\ref{tb:baseline_dataserverandclient} shows the average performance of downstream non-IID FedAvg using non-IID federated methods as initialization.
 It is important to note that we maintain a fixed data splitting setup for our method, meaning we use $\bigcup_{i \in M} D^p_i$ for local training and meta-update the temporary model using $D^S$.
 These comparisons also show the superiority of our method.

\textbf{Varying the amount of the server's dataset used for our method.} We conduct an experiment to see how the quantity of the server's dataset, $|D^S|$, impacts the performance. For the same experimental settings shown in Table~\ref{tbl:sc2} of the manuscript, while keeping the configurations for clients unchanged, we vary the amount of samples used in the server-side data, reducing it from 5\% (default) to 1\% or 2\%. Table~\ref{tb:amount_of_server} shows the result of our method using different amount of server's data for meta-updating temporary global model. We can see that though $D^S$ is small, it effectively contributes to obtaining a well-pretrained model when server data is available. able to benefit in terms of performance accuracy from the amount of pre-training data available at the server. Nevertheless, comparing with the baselines shown in Table~\ref{tbl:sc2}, our CoPreFL achieves a better accuracy with a smaller variance, even with just 1\% server-side data, further confirming its effectiveness.

\subsection{Testing Accuracy Distribution of Downstream FL tasks}
\label{ap:histogram}

\begin{figure*}[t]
    \centering
    \setlength{\abovecaptionskip}{1mm}   \includegraphics[width=\linewidth]{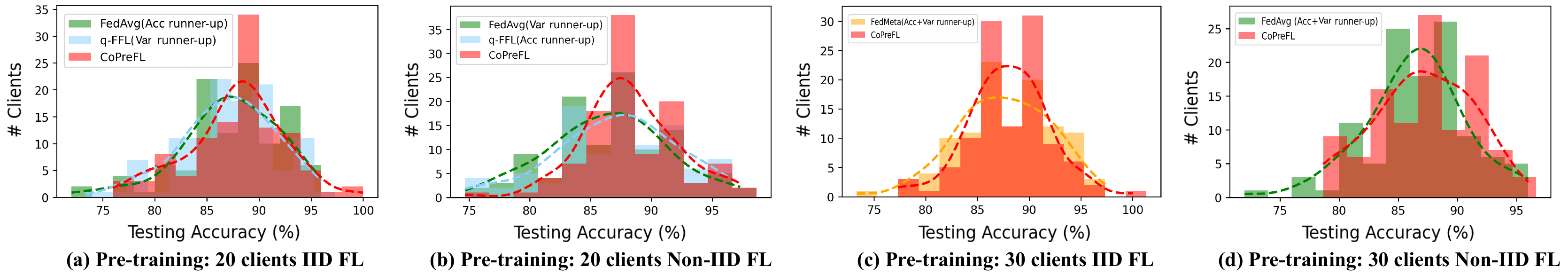}
    \caption{The distributions of testing accuracy in \textbf{IID FL} downstream tasks under various pre-training setups in \textbf{scenario I} on the \textbf{CIFAR-100} dataset.}
    \label{fig:AP_iidfinetune_cifar}
\end{figure*}

\begin{figure*}[t]
    \centering
    \setlength{\abovecaptionskip}{1mm}   \includegraphics[width=\linewidth]{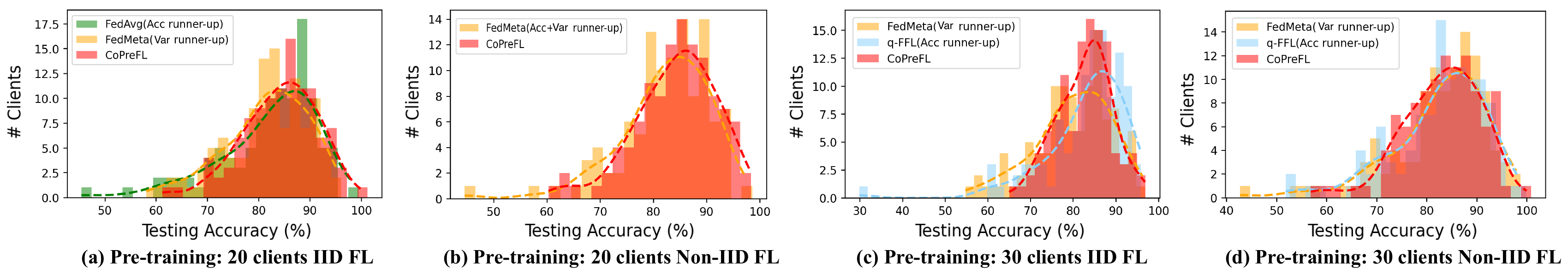}
    % \caption{The distributions of testing accuracy in non-IID FL downstream tasks under various pre-training setups on the CIFAR-100 dataset show that our pre-trained method provides a fairer initialization, resulting in more centered testing accuracy distributions.}
    \caption{The distributions of testing accuracy in \textbf{non-IID FL} downstream tasks under various pre-training setups in \textbf{scenario I} on the \textbf{CIFAR-100} dataset.}
    \label{fig:AP_noniidfinetune_cifar}
\end{figure*}

\begin{figure*}[t]
    \centering
    \setlength{\abovecaptionskip}{1mm}   \includegraphics[width=\linewidth]{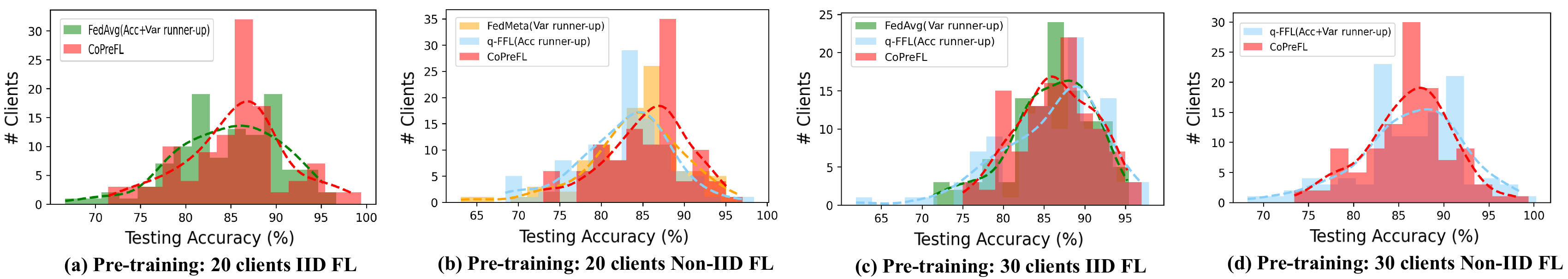}
        \caption{The distributions of testing accuracy in \textbf{IID FL} downstream tasks under various pre-training setups in \textbf{scenario I} on the \textbf{Tiny-ImageNet} dataset.} 
        \label{fig:AP_iidfinetune_tiny}
\end{figure*}

\begin{figure*}[t]
    \centering
    \setlength{\abovecaptionskip}{1mm}   \includegraphics[width=\linewidth]{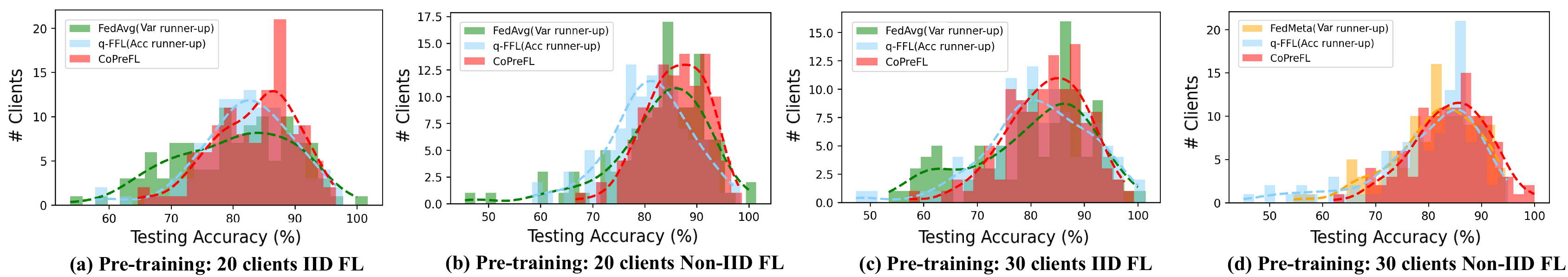}
\caption{The distributions of testing accuracy in \textbf{non-IID FL} downstream tasks under various pre-training setups in \textbf{scenario I} on the \textbf{Tiny-ImageNet} dataset} 
\label{fig:AP_noniidfinetune_tiny}
\end{figure*}

\begin{figure*}[t]
    \centering
    \setlength{\abovecaptionskip}{1mm}   \includegraphics[width=\linewidth]{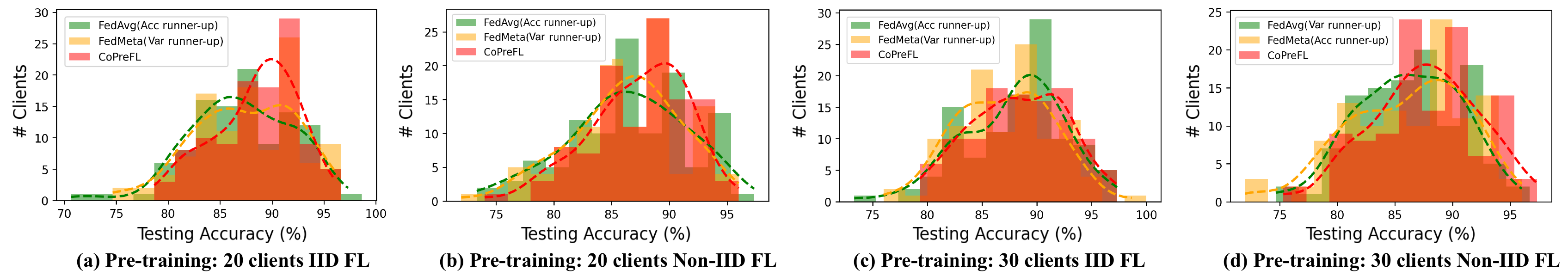}
\caption{The distributions of testing accuracy in \textbf{IID FL} downstream tasks under various pre-training setups in \textbf{scenario II} on the \textbf{CIFAR-100} dataset.} 
\label{fig:AP_iidfinetune_cifar_sc2}
\end{figure*}

\begin{figure*}[t]
    \centering
    \setlength{\abovecaptionskip}{1mm}   \includegraphics[width=\linewidth]{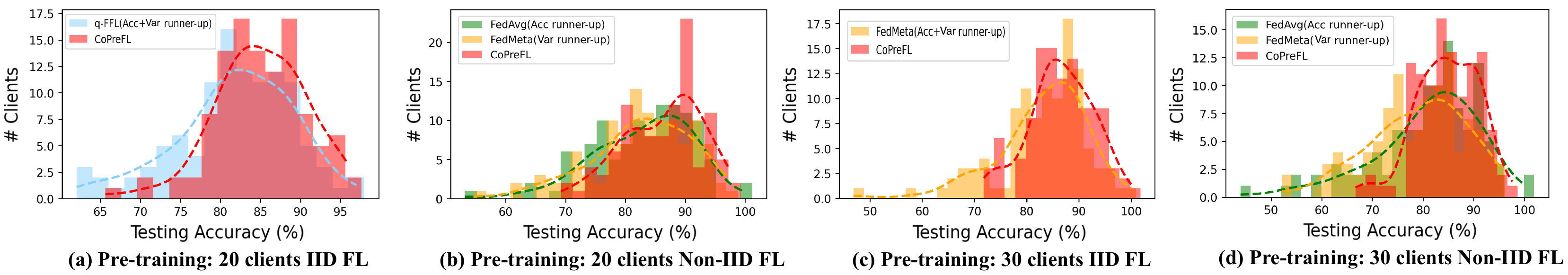}
\caption{The distributions of testing accuracy in \textbf{non-IID FL} downstream tasks under various pre-training setups in \textbf{scenario II} on the \textbf{CIFAR-100} dataset} 
\label{fig:AP_noniidfinetune_cifar_sc2}
\end{figure*}

\begin{figure*}[t]
    \centering
    \setlength{\abovecaptionskip}{1mm}   \includegraphics[width=\linewidth]{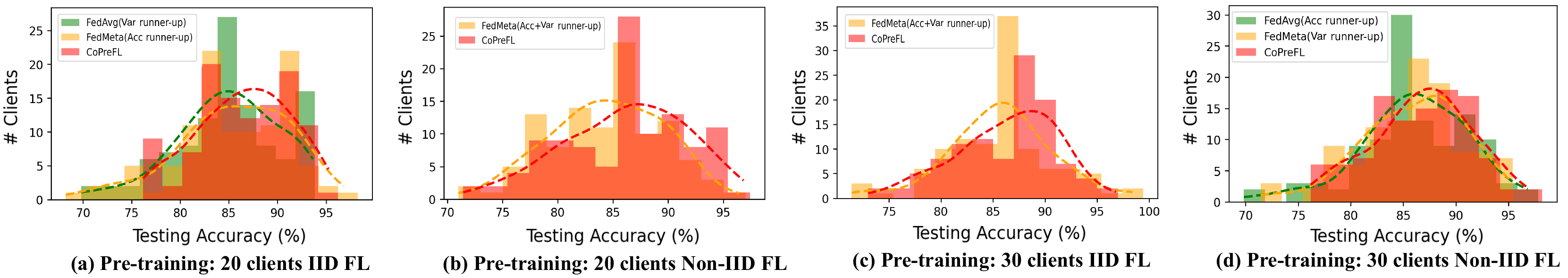}
\caption{The distributions of testing accuracy in \textbf{IID FL} downstream tasks under various pre-training setups in \textbf{scenario II} on the \textbf{Tiny-ImageNet} dataset.} 
\label{fig:AP_iidfinetune_tiny_sc2}
\end{figure*}

\begin{figure*}[t]
    \centering
    \setlength{\abovecaptionskip}{1mm}   \includegraphics[width=\linewidth]{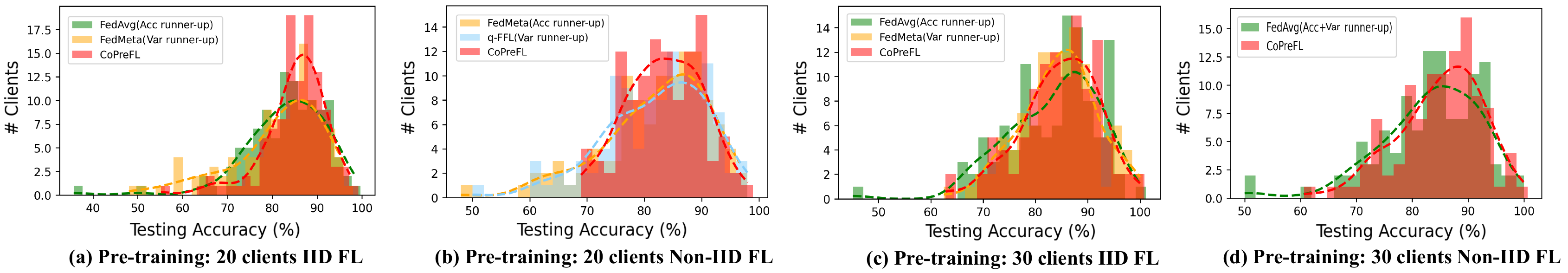}
\caption{The distributions of testing accuracy in \textbf{non-IID FL} downstream tasks under various pre-training setups in \textbf{scenario II} on the \textbf{Tiny-ImageNet} dataset.} 
\label{fig:AP_noniidfinetune_tiny_sc2}
\end{figure*}

This section presents supplementary distribution results to evaluate the performance balance of the pre-trained models discussed in Section~\ref{sec:exp_result}.
For pre-trained models trained in \texttt{scenario} I, Figures~\ref{fig:AP_iidfinetune_cifar} and \ref{fig:AP_noniidfinetune_cifar} show the testing accuracy distribution of IID and non-IID FL tasks on CIFAR-100 dataset, and Figures \ref{fig:AP_iidfinetune_tiny} and \ref{fig:AP_noniidfinetune_tiny} display the respective distribution on Tiny-ImageNet dataset.
Figures~\ref{fig:AP_iidfinetune_cifar_sc2} and \ref{fig:AP_noniidfinetune_cifar_sc2} present the testing accuracy distribution of IID and non-IID FL tasks initialized by pre-trained models trained in \texttt{scenario} II on CIFAR-100 dataset, and Figures \ref{fig:AP_iidfinetune_tiny_sc2} and \ref{fig:AP_noniidfinetune_tiny_sc2} show the respective distribution on Tiny-ImageNet dataset.
Across our experimental results, which encompass different data distribution setups and scenarios during the pre-training phase and various datasets, our {\system} consistently enhances the performance balance of testing accuracy distributions for diverse downstream FL tasks.
In general, distributions of FL tasks initialized by our {\system} tend to shift towards the right, indicating improved prediction performance. 
Moreover, when analyzing clients positioned at the left end of the distribution in each pre-training method, our approach effectively elevates underperforming clients towards the right end, resulting in enhanced predictive accuracy for these clients.

\subsection{Details and Additional Results for FEMNIST Dataset}
\label{ap:femnist}
We also consider the FEMNIST dataset, widely used in FL research, following the data partition provided in~\citep{Park2021FewRoundLF}.
We divide the 62 classes into 52 alphabet classes for the pre-training phase, reserving the remaining 10 digit classes for downstream FL tasks. Instead of using a ResNet-18 model, we employ a model consisting of two 3$\times$3 convolutional layers followed by two linear layers. 
We fixed the total number of clients as $|M| = 100$ for pre-training and $|G| = 10$ for downstream FL tasks.
During the pre-training phase, we set the number of participants $|m| = 20$ and the federated round $T = 50$ for each federated pre-trained method. 
We use the SGD optimizer with a learning rate of $10^{-3}$ and batch size 32 for baselines and our method.

For downstream tasks, we randomly select 5 classes from a pool of 10 classes to conduct each FL task using FedAvg. We perform a total of $X = 10$ FL tasks and report the average evaluations across these tasks. Each task executes FedAvg for $R = 10$ rounds using the SGD optimizer with a learning rate of $10^{-3}$.
Tables \ref{tb:femnist_20_iidfinetune} and \ref{tb:femnist_20_noniidfinetune} display the averaged performance of 10 IID and 10 non-IID FL downstream tasks, initialized by various pre-training methods trained in \texttt{scenario} I, on the FEMNIST dataset.
For \texttt{scenario} II, Tables~\ref{tb:femnist_20_iidfinetune_sc2} and \ref{tb:femnist_20_noniidfinetune_sc2} show the performance of IID and non-IID downstream FL tasks.
The results also demonstrate that our proposed {\system} serves as a robust initialization for various FL setups, benefiting both averaged accuracy and performance balance.

\subsection{Implementation Details and Additional Results for Different Initialization Methods}
\label{ap:otherinitial}

This section presents supplementary details and results with different initialization methods discussed in Section~\ref{sec:exp_result}, including random initialization, centralized model initialization, and other FL algorithms used for initializing downstream FL.
For \texttt{scenario} I, the centralized model is trained on a dataset collected from all $|M| = 100$ clients during the pre-training phase. 
In \texttt{scenario} II, the centralized model is trained on a dataset obtained from both $|M| = 100$ clients and the server. 
This centralized training is conducted using the SGD optimizer with a learning rate of $10^{-3}$ chosen from the range [1e-2, 5e-3, 1e-3, 5e-4], with a batch size of 64 and 50 epochs.

For FL baselines, we additionally consider SCAFFOLD~\cite{Karimireddy2019SCAFFOLDSC}, which addresses partial client sampling, FedDyn~\cite{Acar2021FederatedLB}, designed to tackle non-IID issues, and PerFedAvg~\cite{Fallah2020PersonalizedFL}, aiming to provide an adaptable personalized model, for a detailed comparison. 
We train all FL algorithms for 50 rounds with $|m| = 20$ participants selected from $|M| = 100$ clients under a non-IID setting (Dirichlet $\alpha = 0.5$) , and the final global model is used as initialization for downstream FedAvg.
In the case of non-IID related FL, FedDyn, we set the parameter $\alpha$ to 0.01. 
For personalized FL, PerFedAvg, we employ a two-step gradient descent for local client training introduced in their paper. 
We use the SGD optimizer with a learning rate of $10^{-3}$, a batch size of 32, and 50 federated rounds for these FL-based baselines.

% For scenario I, the centralized pre-trained model is trained on a dataset collected from all $M = 100$ clients during the pre-training phase.
% In scenario II, the centralized method is trained on a dataset obtained from both $M = 100$ clients and the server. 
% The centralized training is conducted using the SGD optimizer with a learning rate of $10^{-3}$ chosen from the range [1e-2, 5e-3, 1e-3, 5e-4], with a batch size of 64 and 50 epochs.
Tables \ref{tb:cfiar_othercomp_iiddown} and \ref{tb:tiny_othercomp_iiddown} display the average performance of 10 FL downstream tasks initialized by different pre-training methods trained in two scenarios on the CIFAR-100 dataset and Tiny-ImageNet dataset, respectively.
Comparing centralized and random initialization, we observe that the centralized method generally improves the average accuracy of downstream FL but at the cost of higher variance in most cases. 
However, our {\system} consistently enhances both average accuracy and performance balance in various downstream FL tasks, demonstrating that with proper FL designs as pre-trained model, FL can be improved through initialization.
Comparing with other FL designs as initialization, the results demonstrate the superiority of our method due to the considerations for unseen adaptation and performance balance during pre-training phase.

\subsection{Implementation Details and Additional Results for Different Downstream FL Tasks}\label{ap:different_down}
In addition to the general downstream FL tasks built by FedAvg, we consider FedProx~\cite{Sahu2018FederatedOI} and q-FFL~\cite{Li2020Fair}, more advanced FL algorithms that addresses heterogeneity and performance balance compared to FedAvg, to examine the robustness and generalizability of our pre-trained method.
The experiments are conducted using the CIFAR-100 dataset under non-IID pre-training \texttt{scenario} I.
Two additional FL algorithms, non-IID FedProx and non-IID q-FFL, are considered for downstream phase.
We randomly sample 5 classes from the 20 available in our CIFAR-100 downstream dataset for each downstream task.
The sampled data is then distributed to 10 clients, and the training consists of 50 rounds with 5 local iterations per round, utilizing an SGD optimizer with a learning rate of $10^{-3}$.
We set the parameters $\mu = 1$ for the proximal term coefficient in FedProx and $q = 2$ for the loss-reweighting coefficient in q-FFL, following the optimal values reported by the authors for the CIFAR-100 dataset.

In Tables~\ref{tb:rebuttal_FedProx} and ~\ref{tb:rebuttal_qffl}, the results demonstrate that our pre-trained method maintains superiority in different downstream FL algorithms compared to other pre-training methods. 
It is important to note that the choice of FedAvg as our downstream task is made to minimize the varying impact introduced by other FL algorithms. 
Comparing the pre-training $+$ downstream pairs, the improvement of CoPreFL $+$ FedAvg (in Table~\ref{tbl:sc1}) over Centralized $+$ FedProx/q-FFL (in Table~\ref{tb:rebuttal_FedProx} and ~\ref{tb:rebuttal_qffl}) shows that a better initialization, which considers the distributed scenario and balances performance in the pre-training phase, could potentially benefit the inferior downstream FL algorithm.

\subsection{Implementation Details when A Public Large-Scale Dataset is Available}\label{ap:largescale}

In addition to utilizing CIFAR-100 and Tiny-ImageNet datasets, where we partition the datasets for pre-training and downstream tasks, we also explore a scenario where public large datasets are available for pre-training phase.
We conducted experiments using pre-trained models with the ImageNet dataset~\citep{Deng2009ImageNetAL}, a widely used large public dataset for image classification.
% We sampled 200 images for each of the 1000 classes in ImageNet\_1K.
% Both a centralized model and our proposed {\system} were pre-trained using ImageNet\_1K, and we initialized downstream FedAvg tasks using these methods. 
We sampled 200 images for each of the 1,000 classes in ImageNet\_1K as a pre-training dataset. 
We conducted pre-training using both the centralized method and our proposed {\system} with ImageNet\_1K. 
Subsequently, we conducted 10 non-IID FedAvg tasks using the CIFAR-100 dataset and initialized the models with these pre-trained models.
For the centralized model in pre-training phase, we trained the model with the SGD optimizer and a learning rate of 1e-3, training the model for 50 epochs. 
For our proposed method during pre-training, we distributed all the sampled data across $|M| = 100$ clients based on non-IID distribution (Dirichlet $\alpha = 0.5$), sampling $|m| = 20$ clients in each round, and conducted {\system} for 50 rounds. 
Since the goal of this experiment is to demonstrate that even with a centrally-stored public large dataset, we can intentionally distribute the dataset and apply our method, we only consider conducting our method under scenario I.
For the downstream FL phase, we apply 10 IID and non-IID FedAvg tasks to the 20-class and 40-class downstream datasets we used in CIFAR-100 and Tiny-ImageNet, respectively.
Each task executes FedAvg for $R = 10$ rounds using the SGD optimizer with a learning rate of $10^{-3}$.
Note that all classes observed during downstream tasks are the seen classes that have already appeared during pre-training, given that CIFAR-100 and Tiny-ImageNet are the subsets of ImageNet.

Table~\ref{tb:imagenet_cifar_tiny} shows the performance of FL downstream tasks on CIFAR-100 and Tiny-ImageNet, where the downstream tasks are initialized by different methods trained on ImageNet\_1K dataset.
As we can expect, models pre-trained on ImageNet\_1K provide downstream FL tasks with higher accuracy compared to those pre-trained on CIFAR-100 or Tiny-ImageNet (in Table~\ref{tbl:sc1}) since there is no unseen classes when pre-trained on ImageNet. 
More importantly, as mentioned, we can still apply our method by intentionally splitting the dataset and mimicking the distributed nature of downstream FL to achieve further performance improvements: The centrally pre-trained model on ImageNet achieves lower accuracy and higher variance compared to our {\system}.
This advantage of {\system} is achieved by initializing the model to get higher accuracy and balanced performance  in federated settings based on meta-learning.
The overall results further confirm the advantage and applicability of our approach.

\section{Limitations and Impact Statements}
\label{app:limit}
This paper presents a pre-training method with potential applications in various AI domains, including natural language processing and computer vision. It is essential to recognize and address potential ethical and privacy concerns associated with the pre-training dataset. For instance, considerations should be made for privacy issues related to images containing human faces and ethical concerns regarding toxic texts. By acknowledging and mitigating any such issues that arise, we can further promote responsible and ethical advancement of AI/ML technologies.

\clearpage
\begin{table*}
\small
\begin{center}
\scalebox{0.9}{
\begin{tabular}{cc||cccccc}
\hline
\multicolumn{2}{c||}{\textbf{Pre-training} (Scenario I, $|m| = 15$)} & \multicolumn{5}{c}{\textbf{Downstream:} IID FedAvg}\\
\hline\hline
 \textbf{Distribution} & {\textbf{Method}} & \textbf{Acc} $\uparrow$ & \textbf{Variance} $\downarrow$ & \textbf{Lowest 10\%} $\uparrow$& \textbf{Lowest 20\%} $\uparrow$&\textbf{Lowest 30\%} $\uparrow$\\
\hline

\multirow{4}{*}{IID} & {FedAvg} & 87.01 $\pm$ 1.07 &	15.44 $\pm$ 1.43 &		78.91 $\pm$ 1.95 &	80.55 $\pm$ 2.38 &	81.41 $\pm$ 1.94\\
& {FedMeta} & 87.09 $\pm$ 1.42 &	14.67 $\pm$ 1.95 &	81.45 $\pm$ 2.07 &	82.42 $\pm$ 1.98 &	83.15 $\pm$ 2.05\\
& {q-FFL} & 87.25 $\pm$ 1.64 &	13.25 $\pm$ 1.38&		80.85 $\pm$ 2.35 &	81.52 $\pm$ 1.66 &	82.26 $\pm$ 2.31\\
& \systemnott ($\gamma$ = 0.5) & \textbf{87.84} $\pm$ 1.05 &	\textbf{11.49} $\pm$ 1.44 &		\textbf{82.61} $\pm$ 1.84&	\textbf{83.52} $\pm$ 1.73 &	\textbf{84.44} $\pm$ 2.03\\
\hline

\multirow{4}{*}{Non-IID} & {FedAvg} & 85.85 $\pm$ 1.35 &	15.37 $\pm$ 1.71 &		78.91 $\pm$ 2.03 &	80.55 $\pm$ 1.88  &	81.41 $\pm$ 2.15\\
& {FedMeta} & 86.84 $\pm$ 1.65 &	12.25 $\pm$ 1.44 &		81.45 $\pm$ 2.07 & 82.42 $\pm$ 1.68 &	83.15 $\pm$ 1.95 \\
& {q-FFL} & 86.37 $\pm$ 2.03 & 13.54 $\pm$ 1.51 &		80.85 $\pm$ 1.17 &	81.52 $\pm$ 2.04 & 82.26 $\pm$ 2.16 \\
& \systemnott ($\gamma$ = 0.25) & \textbf{86.90} $\pm$ 1.41 &	\textbf{8.70} $\pm$ 1.55 &	\textbf{81.52} $\pm$ 1.30 &	\textbf{82.58} $\pm$ 1.79 &	\textbf{83.21} $\pm$ 2.07\\
\hline

\end{tabular}
}
\caption{Average performance across 10 \textbf{IID} downstream FL tasks, initialized with various FL pre-trained methods using \textbf{15} out of 100 participants in \textbf{scenario I}, on the \textbf{CIFAR-100} dataset.}
% \vspace{-1.5pc}
\label{tb:cfiar_15_iidfinetune}
\end{center}
\end{table*}
\begin{table*}
\small
\begin{center}
\scalebox{0.9}{
\begin{tabular}{cc||cccccc}
\hline
\multicolumn{2}{c||}{\textbf{Pre-training} (Scenario I, $|m| = 20$)} & \multicolumn{5}{c}{\textbf{Downstream:} IID FedAvg}\\
\hline\hline
 \textbf{Distribution} & {\textbf{Method}} & \textbf{Acc} $\uparrow$ & \textbf{Variance} $\downarrow$ & \textbf{Lowest 10\%} $\uparrow$& \textbf{Lowest 20\%} $\uparrow$&\textbf{Lowest 30\%} $\uparrow$\\
\hline

\multirow{4}{*}{IID} & {FedAvg} & 87.34 $\pm$  0.95 &	12.46 $\pm$ 1.23 & 	80.48 $\pm$ 2.33  &	81.64 $\pm$ 1.97 &	82.51 $\pm$ 1.65 \\
& {FedMeta} & 86.70 $\pm$ 1.33 &	14.52 $\pm$ 1.29 &		81.33 $\pm$ 1.87 & 82.06 $\pm$ 2.09 &	82.75 $\pm$ 2.15 \\
& {q-FFL} & 86.95 $\pm$ 0.68 &	11.97 $\pm$ 1.48 &		80.48 $\pm$ 1.95 &	81.58 $\pm$ 2.33 &	82.51 $\pm$ 2.07 \\
& \systemnott ($\gamma$ = 0.5) & \textbf{87.54} $\pm$ 0.79 &	\textbf{10.18} $\pm$ 1.25 &		\textbf{81.94} $\pm$ 1.61 & 	\textbf{82.97} $\pm$ 2.05 &	\textbf{83.84} $\pm$ 1.94\\
\hline

\multirow{4}{*}{Non-IID} & {FedAvg} & 86.04 $\pm$ 1.31 &	14.36 $\pm$ 2.07  &		80.85 $\pm$ 1.98 &	81.39 $\pm$ 3.07 &	82.02 $\pm$ 2.55\\
& {FedMeta} & 86.15 $\pm$ 1.11 &	16.16 $\pm$ 2.09 &		79.52 $\pm$ 1.98 & 81.27 $\pm$ 2.71 &	82.18 
 $\pm$ 1.67 \\
& {q-FFL} &  86.30 $\pm$ 1.35&	17.14 $\pm$ 1.93 &		80.24 $\pm$ 1.66 & 81.58 $\pm$ 2.35 & 82.46 $\pm$ 1.79 \\
& \systemnott ($\gamma$ = 0.25) & \textbf{86.32} $\pm$ 1.04 &	\textbf{14.14} $\pm$ 1.58 &	\textbf{81.45} $\pm$ 1.77 &	\textbf{82.20} $\pm$ 2.05 &	\textbf{82.75} $\pm$ 1.88 \\
\hline

\end{tabular}
}
\caption{Average performance across 10 \textbf{IID} downstream FL tasks, initialized with various FL pre-trained methods using \textbf{20} out of 100 participants in \textbf{scenario I}, on the \textbf{CIFAR-100} dataset.}
% \vspace{-1.5pc}
\label{tb:cfiar_20_iidfinetune}
\end{center}
\end{table*}
\begin{table*}
\small
\begin{center}
\scalebox{0.9}{
\begin{tabular}{cc||cccccc}
\hline
\multicolumn{2}{c||}{\textbf{Pre-training} (Scenario I, $|m| = 25$)} & \multicolumn{5}{c}{\textbf{Downstream:} IID FedAvg}\\
\hline\hline
 \textbf{Distribution} & {\textbf{Method}} & \textbf{Acc} $\uparrow$ & \textbf{Variance} $\downarrow$ & \textbf{Lowest 10\%} $\uparrow$& \textbf{Lowest 20\%} $\uparrow$&\textbf{Lowest 30\%} $\uparrow$\\
\hline

\multirow{4}{*}{IID} & {FedAvg} & 87.68 $\pm$ 1.04 & 11.36 $\pm$ 1.88 & 81.58 $\pm$ 2.30 &	82.79 $\pm$ 1.92  &	83.68 $\pm$ 1.63\\
& {FedMeta} & 87.10 $\pm$ 1.28 & 13.62 $\pm$ 2.39 & 81.45 $\pm$  1.94 &	82.61 $\pm$  2.05 &	83.23 $\pm$ 1.99\\
& {q-FFL} & 87.07 $\pm$ 2.01 & 17.89 $\pm$ 1.95 & 80.48 $\pm$ 2.05 &	81.82 $\pm$ 1.83 &	82.59 $\pm$ 2.30\\
& \systemnott ($\gamma$ = 0.0) & \textbf{88.13} $\pm$ 1.16 & \textbf{9.30} $\pm$ 1.95 & \textbf{82.85} $\pm$ 2.00 &	\textbf{83.94} $\pm$ 1.98 &	\textbf{84.75} $\pm$ 2.00\\
\hline

\multirow{4}{*}{Non-IID} & {FedAvg} & 86.78  $\pm$ 1.67 &	11.90  $\pm$ 2.03 & 81.09 $\pm$ 1.89 &	81.82 $\pm$  2.37 &	82.55 $\pm$ 2.51 \\
& {FedMeta} & 85.41  $\pm$ 1.88 & 15.05  $\pm$ 2.35 & 79.15 $\pm$ 2.04 &	79.94 $\pm$ 1.96 &	80.81 $\pm$ 2.21\\
& {q-FFL} & 85.92  $\pm$ 2.05 &	12.11 $\pm$ 1.97 & 79.03  $\pm$ 1.95 &	80.55 $\pm$ 2.71 &	81.49 $\pm$ 2.33\\
& \systemnott ($\gamma$ = 0.0) & \textbf{86.84} $\pm$ 1.53 & \textbf{11.16} $\pm$ 2.03 & \textbf{82.06} $\pm$ 1.91  &	\textbf{82.85} $\pm$ 2.33  &	\textbf{83.43} $\pm$ 2.07 \\
\hline

\end{tabular}
}
\caption{Average performance across 10 \textbf{IID} downstream FL tasks, initialized with various FL pre-trained methods using \textbf{25} out of 100 participants in \textbf{scenario I}, on the \textbf{CIFAR-100} dataset.}
% \vspace{-1.5pc}
\label{tb:cfiar_25_iidfinetune}
\end{center}
\end{table*}
\begin{table*}
\small
\begin{center}
\scalebox{0.9}{
\begin{tabular}{cc||cccccc}
\hline
\multicolumn{2}{c||}{\textbf{Pre-training} (Scenario I, $|m| = 30$)} & \multicolumn{5}{c}{\textbf{Downstream:} IID FedAvg}\\
\hline\hline
 \textbf{Distribution} & {\textbf{Method}} & \textbf{Acc} $\uparrow$ & \textbf{Variance} $\downarrow$ & \textbf{Lowest 10\%} $\uparrow$& \textbf{Lowest 20\%} $\uparrow$&\textbf{Lowest 30\%} $\uparrow$\\
\hline

\multirow{4}{*}{IID} & {FedAvg} & 86.52  $\pm$ 1.55 &	13.54  $\pm$ 2.07 & 80.85 $\pm$ 1.35 &	81.76 $\pm$ 1.69  & 82.55 $\pm$ 1.93\\
& {FedMeta} & 87.65  $\pm$ 2.01 &	13.47  $\pm$ 2.25 & 81.58  $\pm$ 1.94 &	82.91  $\pm$ 2.71 & 83.64 $\pm$ 1.33 \\
& {q-FFL} & 86.40 $\pm$ 1.39 &	15.68 $\pm$ 1.87 & 79.27  $\pm$ 1.85 &	80.12  $\pm$ 1.97 & 81.45 $\pm$ 1.64 \\
& \systemnott ($\gamma$ = 0.0 ) & \textbf{87.90} $\pm$ 1.39 & \textbf{11.16}  $\pm$ 1.96 & \textbf{82.67}  $\pm$ 1.81 &	\textbf{83.58}  $\pm$ 1.73& \textbf{84.36} $\pm$ 1.87\\
\hline

\multirow{4}{*}{Non-IID} & {FedAvg} & 86.78  $\pm$ 1.72 &	12.32  $\pm$ 1.94 & 80.85  $\pm$ 2.03 &	81.76 $\pm$ 1.88 &	82.55 $\pm$ 1.65\\
& {FedMeta} & 85.87 $\pm$ 1.49 &	17.22 $\pm$ 2.22 & 81.58 $\pm$ 2.31 &	82.09 $\pm$ 2.25 & 82.64 $\pm$ 2.07 \\
& {q-FFL} & 85.77 $\pm$ 1.61 &	13.40 $\pm$ 1.88 & 79.27 $\pm$ 1.74 &	80.12  $\pm$ 2.35 & 81.45  $\pm$ 2.41 \\
& \systemnott ($\gamma$ = 0.5)  & \textbf{87.04} $\pm$ 1.52 &	\textbf{9.18}  $\pm$ 1.64 & \textbf{81.70}  $\pm$ 1.93&	\textbf{82.12} $\pm$ 1.99 &	\textbf{82.91} $\pm$ 2.07 \\
\hline

\end{tabular}
}
\caption{Average performance across 10 \textbf{IID} downstream FL tasks, initialized with various FL pre-trained methods using \textbf{30} out of 100 participants in \textbf{scenario I}, on the \textbf{CIFAR-100} dataset.}
% \vspace{-1.5pc}
\label{tb:cfiar_30_iidfinetune}
\end{center}
\end{table*}

\begin{table*}
\small
\begin{center}
\scalebox{0.9}{
\begin{tabular}{cc||cccccc}
\hline
\multicolumn{2}{c||}{\textbf{Pre-training} (Scenario I, $|m| = 15$)} & \multicolumn{5}{c}{\textbf{Downstream:} Non-IID FedAvg}\\
\hline\hline
 \textbf{Distribution} & {\textbf{Method}} & \textbf{Acc} $\uparrow$ & \textbf{Variance} $\downarrow$ & \textbf{Lowest 10\%} $\uparrow$& \textbf{Lowest 20\%} $\uparrow$&\textbf{Lowest 30\%} $\uparrow$\\
\hline

\multirow{4}{*}{IID} & {FedAvg} & 81.46 $\pm$ 2.39 &	62.09 $\pm$ 2.95 & 	68.87 $\pm$ 3.01 &	71.12 $\pm$  2.79 &	72.76 $\pm$ 2.95\\
& {FedMeta} & 81.20 $\pm$ 3.07 &	63.84 $\pm$ 3.35  &		69.39  $\pm$ 2.38 &	71.52  $\pm$ 2.74&	73.14 $\pm$ 3.06 \\
& {q-FFL} & 83.45  $\pm$ 2.88 &	39.94 $\pm$ 3.05 &	69.95 $\pm$ 2.96 &	73.66 $\pm$ 2.57 &	75.43 $\pm$ 3.15\\
& \systemnott ($\gamma$ = 0.75) & \textbf{84.79} $\pm$ 2.61 &	\textbf{37.09}  $\pm$ 2.83&		\textbf{72.71}  $\pm$ 2.54 &	\textbf{74.80}  $\pm$ 2.61&	\textbf{76.75} $\pm$ 2.66 \\
\hline

\multirow{4}{*}{Non-IID} & {FedAvg} & 83.76 $\pm$ 2.54 &	51.84 $\pm$ 3.08 &		69.50  $\pm$ 2.69 &	72.80  $\pm$ 3.33 &	74.30 $\pm$ 3.01\\
& {FedMeta} & 82.65 $\pm$ 3.37 &	39.19  $\pm$ 3.52 &		69.39  $\pm$ 3.07 & 72.76 $\pm$ 2.94  &	74.87 $\pm$ 2.85\\
& {q-FFL} & 82.00 $\pm$ 2.65 &	53.00 $\pm$ 3.19  &		70.78 $\pm$ 2.77 &	73.03 $\pm$ 2.93 &	74.22 $\pm$ 2.85\\
& \systemnott ($\gamma$ = 0.25) & \textbf{84.55} $\pm$ 2.39 &	\textbf{38.07} $\pm$ 3.45 &		\textbf{71.47} $\pm$ 2.63 &	\textbf{73.40} $\pm$ 3.05 &	\textbf{75.20} $\pm$ 2.91\\
\hline

\end{tabular}
}
\caption{Average performance across 10 \textbf{non-IID} downstream FL tasks, initialized with various FL pre-trained methods using \textbf{15} out of 100 participants in \textbf{scenario I}, on the \textbf{CIFAR-100} dataset.}
% \vspace{-1.5pc}
\label{tb:cfiar_15_noniidfinetune}
\end{center}
\end{table*}
\begin{table*}
\small
\begin{center}
\scalebox{0.9}{
\begin{tabular}{cc||cccccc}
\hline
\multicolumn{2}{c||}{\textbf{Pre-training} (Scenario I, $|m| = 20$)} & \multicolumn{5}{c}{\textbf{Downstream:} Non-IID FedAvg}\\
\hline\hline
 \textbf{Distribution} & {\textbf{Method}} & \textbf{Acc} $\uparrow$ & \textbf{Variance} $\downarrow$ & \textbf{Lowest 10\%} $\uparrow$& \textbf{Lowest 20\%} $\uparrow$&\textbf{Lowest 30\%} $\uparrow$\\
\hline

\multirow{4}{*}{IID} & {FedAvg} & 84.20  $\pm$ 2.61 &	57.15  $\pm$ 2.38 &		68.43  $\pm$ 2.95 &	71.83 $\pm$ 3.22 &	74.38 $\pm$ 3.07\\
& {FedMeta} & 83.80  $\pm$ 2.95 &	42.64  $\pm$ 2.66 &		72.30 $\pm$ 3.07 & 73.79  $\pm$ 2.95&	75.47 $\pm$ 3.14\\
& {q-FFL} & 82.60  $\pm$ 3.05&	45.56 $\pm$ 2.26 &	70.46 $\pm$ 3.15 &	73.41 $\pm$ 3.07 & 75.14 $\pm$ 3.33\\
& \systemnott ($\gamma$ = 0.25) & \textbf{84.36}  $\pm$ 2.35 &	\textbf{38.56}  $\pm$ 2.33 &		\textbf{73.66}  $\pm$ 2.89 &	\textbf{75.63}  $\pm$ 3.07  &	\textbf{77.40} $\pm$ 3.11\\
\hline

\multirow{4}{*}{Non-IID} & {FedAvg} & 78.96  $\pm$ 2.98 &	64.80  $\pm$ 3.01 &		62.70  $\pm$ 3.35 &	67.00  $\pm$ 2.95 &	69.80 $\pm$ 3.09 \\
& {FedMeta} & 82.45  $\pm$ 3.07 &	48.72 $\pm$ 2.84 &	68.97 $\pm$ 3.04 &	72.41  $\pm$ 3.06 &	74.35 $\pm$ 2.98 \\
& {q-FFL} & 80.01  $\pm$ 2.67& 88.92 $\pm$ 3.31 &		64.39  $\pm$ 2.95 &	67.48  $\pm$ 2.67 & 70.30 $\pm$ 3.08 \\
& \systemnott ($\gamma$ = 0.75) & \textbf{83.29} $\pm$ 2.61 &	\textbf{34.69}  $\pm$ 3.17 &		\textbf{71.58}  $\pm$ 3.00&	\textbf{73.20}  $\pm$ 2.98&	\textbf{74.59} $\pm$ 2.85\\
\hline

\end{tabular}
}
\caption{Average performance across 10 \textbf{non-IID} downstream FL tasks, initialized with various FL pre-trained methods using \textbf{20} out of 100 participants in \textbf{scenario I}, on the \textbf{CIFAR-100} dataset.}
% \vspace{-1.5pc}
\label{tb:cfiar_20_noniidfinetune}
\end{center}
\end{table*}
\begin{table*}
\small
\begin{center}
\scalebox{0.9}{
\begin{tabular}{cc||cccccc}
\hline
\multicolumn{2}{c||}{\textbf{Pre-training} (Scenario I, $|m| = 25$)} & \multicolumn{5}{c}{\textbf{Downstream:} Non-IID FedAvg}\\
\hline\hline
 \textbf{Distribution} & {\textbf{Method}} & \textbf{Acc} $\uparrow$ & \textbf{Variance} $\downarrow$ & \textbf{Lowest 10\%} $\uparrow$& \textbf{Lowest 20\%} $\uparrow$&\textbf{Lowest 30\%} $\uparrow$\\
\hline

\multirow{4}{*}{IID} & {FedAvg} &84.02 $\pm$ 3.17  &	51.98 $\pm$ 2.98 & 71.26 $\pm$ 3.05 &	73.21 $\pm$ 2.88 &	75.57 $\pm$ 2.54\\
& {FedMeta} & 82.44 $\pm$ 3.05 &	55.06 $\pm$ 3.51 & 68.53  $\pm$ 3.29&	71.73 $\pm$ 3.74 &	73.95 $\pm$ 3.85\\
& {q-FFL} & 82.63 $\pm$ 2.94 & 47.20 $\pm$ 3.08 & 70.52  $\pm$ 3.37 &	72.42  $\pm$ 3.49 &	74.01 $\pm$ 3.75 \\
& \systemnott ($\gamma$ = 0.75) & \textbf{85.60}  $\pm$ 2.87 &	\textbf{37.45}  $\pm$ 3.04 & \textbf{74.42} $\pm$ 3.11 & \textbf{76.53}  $\pm$ 3.29 &	\textbf{78.43} $\pm$ 3.57\\
\hline

\multirow{4}{*}{Non-IID} & {FedAvg} & 82.01  $\pm$ 3.32 &	39.82 $\pm$ 3.09 & 70.75  $\pm$ 3.46 &	73.02 $\pm$ 3.81 &	74.63 $\pm$ 3.68 \\
& {FedMeta} & 84.02  $\pm$ 2.77&	39.56  $\pm$ 2.94 & 71.86  $\pm$ 2.61 &	75.17 $\pm$ 2.98  &	76.80 $\pm$ 2.74\\
& {q-FFL} & 82.18 $\pm$ 3.04 &	46.79 $\pm$ 2.85  & 70.53  $\pm$ 2.79 &	72.43 $\pm$ 3.34 &	73.61  $\pm$ 3.49 \\
& \systemnott ($\gamma$ = 0.25) & \textbf{85.72} $\pm$ 3.00 &	\textbf{29.38} $\pm$ 2.67 & \textbf{75.81} $\pm$ 2.83 &	\textbf{77.24} $\pm$ 2.71 &	\textbf{78.54} $\pm$ 3.05\\
\hline

\end{tabular}
}
\caption{Average performance across 10 \textbf{non-IID} downstream FL tasks, initialized with various FL pre-trained methods using \textbf{25} out of 100 participants in \textbf{scenario I}, on the \textbf{CIFAR-100} dataset.}
% \vspace{-1.5pc}
\label{tb:cfiar_25_noniidfinetune}
\end{center}
\end{table*}
\begin{table*}
\small
\begin{center}
\scalebox{0.9}{
\begin{tabular}{cc||cccccc}
\hline
\multicolumn{2}{c||}{\textbf{Pre-training} (Scenario I, $|m| = 30$)} & \multicolumn{5}{c}{\textbf{Downstream:} Non-IID FedAvg}\\
\hline\hline
 \textbf{Distribution} & {\textbf{Method}} & \textbf{Acc} $\uparrow$ & \textbf{Variance} $\downarrow$ & \textbf{Lowest 10\%} $\uparrow$& \textbf{Lowest 20\%} $\uparrow$&\textbf{Lowest 30\%} $\uparrow$\\
\hline

\multirow{4}{*}{IID} & {FedAvg} & 79.60  $\pm$ 3.00 &	83.36 $\pm$ 3.41 & 62.68 $\pm$ 3.82 &	65.87  $\pm$ 3.59&	68.57 $\pm$ 3.33 \\
& {FedMeta} & 79.90  $\pm$ 2.98 &	48.02  $\pm$ 3.61 & 67.01  $\pm$ 2.65 &	69.69 $\pm$ 3.01 &	71.46 $\pm$ 2.85\\
& {q-FFL} & 83.02  $\pm$ 2.65 &	52.27 $\pm$ 3.04 & 70.64 $\pm$ 2.88 &	72.71 $\pm$ 2.73 &	74.50 $\pm$ 2.71\\
& \systemnott ($\gamma$ = 0.75) & \textbf{83.48}  $\pm$ 2.39 & \textbf{45.16} $\pm$ 3.45 & \textbf{70.80}  $\pm$ 3.01 &	\textbf{72.72} $\pm$ 2.87 & \textbf{74.59} $\pm$ 2.72\\
\hline

\multirow{4}{*}{Non-IID} & {FedAvg} & 81.79  $\pm$ 3.41 &	50.84 $\pm$ 3.85  & 69.70  $\pm$ 2.93 &	72.08  $\pm$ 2.88 &	74.20 $\pm$ 3.07 \\
& {FedMeta} & 82.61  $\pm$ 3.54 &	43.43  $\pm$ 3.95 & 71.84  $\pm$ 3.39 &	73.30  $\pm$ 3.86 &	74.68 $\pm$ 3.51 \\
& {q-FFL} & 82.68 $\pm$ 2.94 &	54.17 $\pm$ 3.33 & 68.68 $\pm$ 3.06  &	72.08  $\pm$ 3.17&	74.06 $\pm$ 2.84 \\
& \systemnott ($\gamma$ = 0.75)  & \textbf{83.48} $\pm$ 2.87  & \textbf{40.20}  $\pm$ 3.07 & \textbf{72.83}  $\pm$ 3.16&	\textbf{74.29}  $\pm$ 2.94& \textbf{75.80} $\pm$ 2.98\\
\hline

\end{tabular}
}
\caption{Average performance across 10 \textbf{non-IID} downstream FL tasks, initialized with various FL pre-trained methods using \textbf{30} out of 100 participants in \textbf{scenario I}, on the \textbf{CIFAR-100} dataset.}
% \vspace{-1.5pc}
\label{tb:cfiar_30_noniidfinetune}
\end{center}
\end{table*}

\begin{table*}
\small
\begin{center}
\scalebox{0.9}{
\begin{tabular}{cc||cccccc}
\hline
\multicolumn{2}{c||}{\textbf{Pre-training} (Scenario I, $|m| = 15$)} & \multicolumn{5}{c}{\textbf{Downstream:} IID FedAvg}\\
\hline\hline
 \textbf{Distribution} & {\textbf{Method}} & \textbf{Acc} $\uparrow$ & \textbf{Variance} $\downarrow$ & \textbf{Lowest 10\%} $\uparrow$& \textbf{Lowest 20\%} $\uparrow$&\textbf{Lowest 30\%} $\uparrow$\\
\hline

\multirow{4}{*}{IID} & {FedAvg} & 84.27 $\pm$ 1.37 &	16.48 $\pm$ 1.61  &		77.20 $\pm$ 1.19 &	78.43 $\pm$ 1.53 &	79.27  $\pm$ 1.55 \\
& {FedMeta} & 84.15 $\pm$ 1.28 &	19.27 $\pm$ 2.07 &		78.50 $\pm$ 1.39 &	79.65 $\pm$ 1.21 &	80.66 $\pm$ 1.09\\
& {q-FFL} & 84.24 $\pm$ 0.95 &	17.64 $\pm$ 1.23 &		77.20  $\pm$ 1.09 &	78.86 $\pm$ 1.10 &	79.99 $\pm$ 1.38\\
& \systemnott ($\gamma$ = 0.0) & \textbf{85.05} $\pm$ 1.07 &	\textbf{15.21} $\pm$ 1.06 &		\textbf{79.08} $\pm$ 1.08 &	\textbf{80.38} $\pm$ 1.13 & \textbf{81.19} $\pm$ 1.19 \\
\hline

\multirow{4}{*}{Non-IID} & {FedAvg} & 85.19 $\pm$ 1.52 &15.13 $\pm$ 1.83 &	77.78 $\pm$ 1.66 &	79.29 $\pm$ 1.59 &	80.38 $\pm$ 1.58 \\
& {FedMeta} & 85.35 $\pm$ 1.39 &	15.60 $\pm$ 1.94 &		78.50 $\pm$ 1.43 &	80.01  $\pm$ 1.37&	81.00 $\pm$ 1.57\\
& {q-FFL} & 85.91 $\pm$ 1.44 &	15.76 $\pm$ 2.03  &		78.22 $\pm$ 1.65 &	80.38 $\pm$ 1.37 &	81.12 $\pm$ 1.59\\
& \systemnott ($\gamma$ = 0.0) & \textbf{86.39} $\pm$ 1.27 &	\textbf{10.63}  $\pm$ 1.89 &	\textbf{79.08} $\pm$ 1.33 &	\textbf{80.45} $\pm$ 1.45 &	\textbf{81.24} $\pm$ 1.49\\
\hline

\end{tabular}
}
\caption{Average performance across 10 \textbf{IID} downstream FL tasks, initialized with various FL pre-trained methods using \textbf{15} out of 100 participants in \textbf{scenario I}, on the \textbf{Tiny-ImageNet} dataset.}
% \vspace{-1.5pc}
\label{tb:tiny_15_iidfinetune}
\end{center}
\end{table*}
\begin{table*}
\small
\begin{center}
\scalebox{0.9}{
\begin{tabular}{cc||cccccc}
\hline
\multicolumn{2}{c||}{\textbf{Pre-training} (Scenario I, $|m| = 20$)} & \multicolumn{5}{c}{\textbf{Downstream:} IID FedAvg}\\
\hline\hline
 \textbf{Distribution} & {\textbf{Method}} & \textbf{Acc} $\uparrow$ & \textbf{Variance} $\downarrow$ & \textbf{Lowest 10\%} $\uparrow$& \textbf{Lowest 20\%} $\uparrow$&\textbf{Lowest 30\%} $\uparrow$\\
\hline

\multirow{4}{*}{IID} & {FedAvg} & 85.74 $\pm$ 1.69 &	17.39 $\pm$ 2.05 &		77.20 $\pm$ 1.57 &	78.79 $\pm$ 1.77 &	79.80 $\pm$ 1.83 \\
& {FedMeta} & 85.56 $\pm$  1.93 &	17.64 $\pm$ 1.33 &	77.92 $\pm$ 2.19  &	79.51 $\pm$ 1.84 &	80.57 $\pm$ 1.74\\
& {q-FFL} & 84.64  $\pm$ 2.07 &	21.07 $\pm$ 1.35 &	78.79 $\pm$ 1.66  &	79.74 $\pm$ 1.94 &	80.91 $\pm$ 2.19\\
& \systemnott ($\gamma$ = 0.5) & \textbf{86.03}  $\pm$ 1.32 &	\textbf{13.99}  $\pm$ 1.14 &		\textbf{79.08}  $\pm$ 1.65 &	\textbf{80.09}  $\pm$ 1.33 & \textbf{81.24} $\pm$ 1.46\\
\hline

\multirow{4}{*}{Non-IID} & {FedAvg} & 85.43  $\pm$ 1.38 &	17.31  $\pm$ 2.55 &		77.49  $\pm$ 1.54 & 79.65 $\pm$ 1.69  & 80.76 $\pm$ 1.83\\
& {FedMeta} & 84.16 $\pm$ 2.00 &	16.89 $\pm$ 2.85 &	77.20 $\pm$  2.38&	79.73 $\pm$ 2.19 &	81.05 $\pm$ 2.22\\
& {q-FFL} & 85.83 $\pm$ 1.10 & 19.18 $\pm$ 1.95 &	78.07 $\pm$ 1.77 &	79.37 $\pm$ 1.54 &	80.62 $\pm$ 1.39\\
& \systemnott ($\gamma$ = 0.5) & \textbf{86.00}  $\pm$ 1.13 &	\textbf{16.16} $\pm$ 2.00 &		\textbf{79.37}  $\pm$ 1.38 &	\textbf{80.30} $\pm$ 1.60 & \textbf{81.19} $\pm$ 1.54\\
\hline

\end{tabular}
}
\caption{Average performance across 10 \textbf{IID} downstream FL tasks, initialized with various FL pre-trained methods using \textbf{20} out of 100 participants in \textbf{scenario I}, on the \textbf{Tiny-ImageNet} dataset.}
% \vspace{-1.5pc}
\label{tb:tiny_20_iidfinetune}
\end{center}
\end{table*}
\begin{table*}
\small
\begin{center}
\scalebox{0.9}{
\begin{tabular}{cc||cccccc}
\hline
\multicolumn{2}{c||}{\textbf{Pre-training} (Scenario I, $|m| = 25$)} & \multicolumn{5}{c}{\textbf{Downstream:} IID FedAvg}\\
\hline\hline
 \textbf{Distribution} & {\textbf{Method}} & \textbf{Acc} $\uparrow$ & \textbf{Variance} $\downarrow$ & \textbf{Lowest 10\%} $\uparrow$& \textbf{Lowest 20\%} $\uparrow$&\textbf{Lowest 30\%} $\uparrow$\\
\hline

\multirow{4}{*}{IID} & {FedAvg} & 85.24 $\pm$ 1.19 &	19.71 $\pm$  2.07 & 78.35 $\pm$ 1.38 &	79.87 $\pm$  1.74&	81.05 $\pm$ 1.23 \\
& {FedMeta} & 85.19 $\pm$ 1.44 &	22.00 $\pm$ 2.35 & 78.21 $\pm$ 1.68 &	79.73 $\pm$ 1.54 &	80.71 $\pm$ 1.70\\
& {q-FFL} & 85.26 $\pm$ 2.04 &	16.89 $\pm$ 2.85 & 78.50 $\pm$ 2.39 &	79.94 $\pm$ 2.17 &	81.00 $\pm$ 2.43\\
& \systemnott ($\gamma$ = 0.75) & \textbf{85.74} $\pm$ 1.13 &	\textbf{12.81} $\pm$ 1.96 & \textbf{79.84} $\pm$ 1.30  &	\textbf{80.68} $\pm$ 1.63 &	\textbf{81.49} $\pm$ 1.69 \\
\hline

\multirow{4}{*}{Non-IID} & {FedAvg} & 85.47 $\pm$ 1.38 &	14.36  $\pm$ 2.41 & 77.63 $\pm$ 1.69  &	79.29  $\pm$ 1.35 &	80.33 $\pm$ 1.58\\
& {FedMeta} & 85.74  $\pm$ 1.63 &	17.64  $\pm$ 1.94 & 77.92 $\pm$ 1.47 &	79.80  $\pm$ 1.83 &	81.10 $\pm$ 1.79\\
& {q-FFL} & 85.82 $\pm$ 2.06 &	17.64  $\pm$ 2.58 & 79.08 $\pm$ 1.99  &	80.52  $\pm$ 1.87&	81.58 $\pm$ 2.11 \\
& \systemnott ($\gamma$ = 0.0) & \textbf{86.25} $\pm$ 1.47 &	\textbf{12.96} $\pm$ 1.33 & \textbf{79.87} $\pm$ 1.54 &	\textbf{80.99} $\pm$ 1.35  & \textbf{81.68} $\pm$ 1.41\\
\hline

\end{tabular}
}
\caption{Average performance across 10 \textbf{IID} downstream FL tasks, initialized with various FL pre-trained methods using \textbf{25} out of 100 participants in \textbf{scenario I}, on the \textbf{Tiny-ImageNet} dataset.}
% \vspace{-1.5pc}
\label{tb:tiny_25_iidfinetune}
\end{center}
\end{table*}
\begin{table*}
\small
\begin{center}
\scalebox{0.9}{
\begin{tabular}{cc||cccccc}
\hline
\multicolumn{2}{c||}{\textbf{Pre-training} (Scenario I, $|m| = 30$)} & \multicolumn{5}{c}{\textbf{Downstream:} IID FedAvg}\\
\hline\hline
 \textbf{Distribution} & {\textbf{Method}} & \textbf{Acc} $\uparrow$ & \textbf{Variance} $\downarrow$ & \textbf{Lowest 10\%} $\uparrow$& \textbf{Lowest 20\%} $\uparrow$&\textbf{Lowest 30\%} $\uparrow$\\
\hline

\multirow{4}{*}{IID} & {FedAvg} & 85.71 $\pm$ 1.92 &	\textbf{14.90} $\pm$ 1.61 & 79.08  $\pm$ 1.94 &	80.66 $\pm$ 2.38 &	81.53 $\pm$ 1.77\\
& {FedMeta} & 85.48 $\pm$ 2.07 & 15.92 $\pm$ 2.22 & 79.51 $\pm$ 1.85 &	80.16 $\pm$ 2.03  &	81.19 $\pm$ 2.25\\
& {q-FFL} & 85.95 $\pm$ 2.17 &	21.25 $\pm$ 1.54 & 77.34 $\pm$ 2.08 &	78.93  $\pm$ 2.34 &	80.13 $\pm$ 2.01 \\
& \systemnott ($\gamma$ = 0.75) & \textbf{86.05} $\pm$ 1.85 &	\textbf{14.90}  $\pm$ 1.56 & \textbf{80.31}  $\pm$ 1.73&	\textbf{81.49} $\pm$ 1.99 &	\textbf{82.27} $\pm$ 1.86\\
\hline

\multirow{4}{*}{Non-IID} & {FedAvg} & 85.64 $\pm$ 1.37 &	21.07 $\pm$ 1.94 & 75.90  $\pm$ 1.53 &	77.99 $\pm$ 1.49  &	79.32 $\pm$ 1.55 \\
& {FedMeta} & 85.90  $\pm$ 1.55 &	17.89 $\pm$ 2.37  & 80.23 $\pm$ 1.69 &	81.10  $\pm$ 1.43 &	81.87 $\pm$ 1.79\\
& {q-FFL} & 86.49 $\pm$ 1.45  &14.75  $\pm$ 2.00 & 78.79  $\pm$ 1.67 &	80.16  $\pm$ 1.32 &	81.24 $\pm$ 1.45 \\
& \systemnott ($\gamma$ = 0.0)  & \textbf{86.51}  $\pm$ 1.30 &	\textbf{14.06} $\pm$ 1.73 & \textbf{80.74}  $\pm$ 1.35 &	\textbf{81.35} $\pm$ 1.47 &	\textbf{82.16} $\pm$ 1.60\\
\hline

\end{tabular}
}
\caption{Average performance across 10 \textbf{IID} downstream FL tasks, initialized with various FL pre-trained methods using \textbf{30} out of 100 participants in \textbf{scenario I}, on the \textbf{Tiny-ImageNet} dataset.}
% \vspace{-1.5pc}
\label{tb:tiny_30_iidfinetune}
\end{center}
\end{table*}

\begin{table*}
\small
\begin{center}
\scalebox{0.9}{
\begin{tabular}{cc||cccccc}
\hline
\multicolumn{2}{c||}{\textbf{Pre-training} (Scenario I, $|m| = 15$)} & \multicolumn{5}{c}{\textbf{Downstream:} Non-IID FedAvg}\\
\hline\hline
 \textbf{Distribution} & {\textbf{Method}} & \textbf{Acc} $\uparrow$ & \textbf{Variance} $\downarrow$ & \textbf{Lowest 10\%} $\uparrow$& \textbf{Lowest 20\%} $\uparrow$&\textbf{Lowest 30\%} $\uparrow$\\
\hline

\multirow{4}{*}{IID} & {FedAvg} & 78.88 $\pm$ 2.66 &	64.16 $\pm$ 2.05 &		67.03 $\pm$ 2.39 &	68.42  $\pm$ 3.01 &	69.95 $\pm$ 2.85 \\
& {FedMeta} & 82.62  $\pm$ 2.74 &	43.16  $\pm$ 2.71 &		70.76  $\pm$ 2.55 & 73.48  $\pm$ 2.38 &	74.48  $\pm$ 2.96 \\
& {q-FFL} & 83.58  $\pm$ 3.00 & 49.70  $\pm$ 2.89 &		67.11  $\pm$ 2.93 &	71.38  $\pm$ 2.81  &	73.78  $\pm$ 3.05\\
& \systemnott ($\gamma$ = 0.5) & \textbf{83.83}  $\pm$ 2.54 &	\textbf{41.22}  $\pm$ 2.17 &		\textbf{73.28} $\pm$ 2.41 &	\textbf{74.39}  $\pm$ 2.30&	\textbf{75.50} $\pm$ 2.88\\
\hline

\multirow{4}{*}{Non-IID} & {FedAvg} & 82.19 $\pm$ 2.71 &	38.32 $\pm$ 1.94 &	72.64  $\pm$ 2.10 & 73.90 $\pm$ 1.98 &	75.38 $\pm$ 2.53 \\
& {FedMeta} & 81.45 $\pm$ 2.39 & 	53.73 $\pm$ 2.61 &		68.42 $\pm$ 2.55 &	71.17 $\pm$ 2.08 &	72.98 $\pm$ 2.34\\
& {q-FFL} & 82.85 $\pm$ 2.83 &	32.26 $\pm$ 3.01 &		73.89 $\pm$ 3.19  & 76.14  $\pm$ 2.95 &	77.28 $\pm$ 2.91 \\
& \systemnott ($\gamma$ = 0.25) & \textbf{83.65} $\pm$ 2.39 &	\textbf{25.81}  $\pm$ 2.33 &		\textbf{75.41} $\pm$ 2.37 &	\textbf{76.45} $\pm$ 2.10 &	\textbf{77.73} $\pm$ 2.03\\
\hline

\end{tabular}
}
\caption{Average performance across 10 \textbf{non-IID} downstream FL tasks, initialized with various FL pre-trained methods using \textbf{15} out of 100 participants in \textbf{scenario I}, on the \textbf{Tiny-ImageNet} dataset.}
% \vspace{-1.5pc}
\label{tb:tiny_15_noniidfinetune}
\end{center}
\end{table*}
\begin{table*}
\small
\begin{center}
\scalebox{0.9}{
\begin{tabular}{cc||cccccc}
\hline
\multicolumn{2}{c||}{\textbf{Pre-training} (Scenario I, $|m| = 20$)} & \multicolumn{5}{c}{\textbf{Downstream:} Non-IID FedAvg}\\
\hline\hline
 \textbf{Distribution} & {\textbf{Method}} & \textbf{Acc} $\uparrow$ & \textbf{Variance} $\downarrow$ & \textbf{Lowest 10\%} $\uparrow$& \textbf{Lowest 20\%} $\uparrow$&\textbf{Lowest 30\%} $\uparrow$\\
\hline

\multirow{4}{*}{IID} & {FedAvg} & 79.45 $\pm$ 2.33  &	35.40 $\pm$ 2.79  &		64.86 $\pm$  2.51 &	68.61 $\pm$ 2.49  &	70.78 $\pm$ 2.55 \\
& {FedMeta} & 81.68 $\pm$ 2.57  &	65.61 $\pm$  2.38 &		65.96 $\pm$ 2.61  &	69.17 $\pm$ 2.55  &	71.59 $\pm$ 2.73 \\
& {q-FFL} & 82.65 $\pm$ 2.65  &	39.69 $\pm$  2.59 &		70.65 $\pm$ 2.47  &	74.14 $\pm$ 2.54  &	76.27 $\pm$ 2.89 \\
& \systemnott ($\gamma$ = 0.0) & \textbf{83.79} $\pm$ 2.30  &	\textbf{34.93} $\pm$ 2.44  &		\textbf{72.59}  $\pm$ 2.49  &	\textbf{75.05}  $\pm$ 2.38 &	\textbf{76.76} $\pm$ 2.41  \\
\hline

\multirow{4}{*}{Non-IID} & {FedAvg} & 82.94  $\pm$ 2.59 &	37.21  $\pm$ 2.81 &		68.99  $\pm$ 2.43 &	72.29  $\pm$ 2.61 &	74.40 $\pm$ 2.69 \\
& {FedMeta} & 81.03  $\pm$  2.86 &	37.58  $\pm$ 3.00 &		69.44  $\pm$ 2.61 &	71.55 $\pm$ 2.93  &	72.93 $\pm$ 2.78 \\
& {q-FFL} & 84.11 $\pm$ 2.49  &	43.96 $\pm$  2.71 &		73.87 $\pm$  2.79 &	76.05  $\pm$ 2.61 &	77.37 $\pm$ 2.45 \\
& \systemnott ($\gamma$ = 0.5) & \textbf{85.23}  $\pm$ 2.43 &	\textbf{35.40} $\pm$ 2.75  &	\textbf{76.77} $\pm$ 2.58  &	\textbf{78.46}  $\pm$ 2.47 &	\textbf{79.86} $\pm$ 2.53 \\
\hline

\end{tabular}
}
\caption{Average performance across 10 \textbf{non-IID} downstream FL tasks, initialized with various FL pre-trained methods using \textbf{20} out of 100 participants in \textbf{scenario I}, on the \textbf{Tiny-ImageNet} dataset.}
% \vspace{-1.5pc}
\label{tb:tiny_20_noniidfinetune}
\end{center}
\end{table*}
\begin{table*}
\small
\begin{center}
\scalebox{0.9}{
\begin{tabular}{cc||cccccc}
\hline
\multicolumn{2}{c||}{\textbf{Pre-training} (Scenario I, $|m| = 25$)} & \multicolumn{5}{c}{\textbf{Downstream:} Non-IID FedAvg}\\
\hline\hline
 \textbf{Distribution} & {\textbf{Method}} & \textbf{Acc} $\uparrow$ & \textbf{Variance} $\downarrow$ & \textbf{Lowest 10\%} $\uparrow$& \textbf{Lowest 20\%} $\uparrow$&\textbf{Lowest 30\%} $\uparrow$\\
\hline

\multirow{4}{*}{IID} & {FedAvg} & 83.71 $\pm$ 2.89  &	50.41 $\pm$  2.94 & 69.91 $\pm$ 2.81  &	73.50  $\pm$ 2.97 &	75.40  $\pm$  3.09 \\
& {FedMeta} & 84.19  $\pm$ 2.61 &	42.90 $\pm$ 2.65  & 73.77 $\pm$  2.53 & 76.22  $\pm$ 2.88 &	77.77 $\pm$ 2.97 \\
& {q-FFL} & 80.11 $\pm$ 3.19  &	55.20  $\pm$ 3.07 & 65.45  $\pm$ 2.95 &	68.54  $\pm$ 3.28 & 70.72 $\pm$ 2.97  \\
& \systemnott ($\gamma$ = 0.0) & \textbf{84.29}  $\pm$ 2.66 &	\textbf{36.60} $\pm$ 2.83  & \textbf{76.02} $\pm$ 2.39  &	\textbf{77.56}  $\pm$ 3.01 &	\textbf{78.95} $\pm$ 2.88 \\
\hline

\multirow{4}{*}{Non-IID} & {FedAvg} & 79.08  $\pm$ 2.54 &	55.80  $\pm$ 3.05 & 66.80  $\pm$ 2.79 & 69.06 $\pm$  2.81 & 71.38 $\pm$ 2.53 \\
& {FedMeta} & 81.58  $\pm$ 2.90 &	38.07  $\pm$ 2.88 & 70.86  $\pm$ 3.17 &	72.83  $\pm$ 2.84 &	74.39 $\pm$ 3.05 \\
& {q-FFL} & 83.16  $\pm$ 3.03 &	45.56  $\pm$ 3.15 & 72.39  $\pm$ 2.98 & 75.29  $\pm$ 3.07 &	77.09 $\pm$ 2.88 \\
& \systemnott ($\gamma$ = 0.25) & \textbf{83.87}  $\pm$ 2.61 &	\textbf{25.60} $\pm$ 2.94  & \textbf{75.16} $\pm$ 2.89  &	\textbf{76.87} $\pm$ 2.74  & \textbf{78.05} $\pm$ 2.83 \\
\hline

\end{tabular}
}
\caption{Average performance across 10 \textbf{non-IID} downstream FL tasks, initialized with various FL pre-trained methods using \textbf{25} out of 100 participants in \textbf{scenario I}, on the \textbf{Tiny-ImageNet} dataset.}
% \vspace{-1.5pc}
\label{tb:tiny_25_noniidfinetune}
\end{center}
\end{table*}
\begin{table*}
\small
\begin{center}
\scalebox{0.9}{
\begin{tabular}{cc||cccccc}
\hline
\multicolumn{2}{c||}{\textbf{Pre-training} (Scenario I, $|m| = 30$)} & \multicolumn{5}{c}{\textbf{Downstream:} Non-IID FedAvg}\\
\hline\hline
 \textbf{Distribution} & {\textbf{Method}} & \textbf{Acc} $\uparrow$ & \textbf{Variance} $\downarrow$ & \textbf{Lowest 10\%} $\uparrow$& \textbf{Lowest 20\%} $\uparrow$&\textbf{Lowest 30\%} $\uparrow$\\
\hline

\multirow{4}{*}{IID} & {FedAvg} & 80.37  $\pm$ 3.07 &	43.56  $\pm$  2.93 & 69.27  $\pm$ 2.95  &	70.91   $\pm$ 2.88 &	72.44  $\pm$ 3.19\\
& {FedMeta} & 80.51  $\pm$ 2.87 &	44.09  $\pm$ 3.11 & 68.05  $\pm$  2.79 &	70.74   $\pm$ 2.95 &	72.18  $\pm$ 2.97\\
& {q-FFL} & 81.89  $\pm$ 3.10 &	45.97   $\pm$ 2.87 & 68.85   $\pm$ 3.31 &	72.07   $\pm$ 3.19 &	73.99  $\pm$ 2.93\\
& \systemnott ($\gamma$ = 0.0) & \textbf{83.17}  $\pm$ 2.65 &	\textbf{31.81}   $\pm$ 2.88 & \textbf{71.16}   $\pm$ 2.85 &	\textbf{73.64}   $\pm$ 2.79 &	\textbf{75.49}  $\pm$ 2.80\\
\hline

\multirow{4}{*}{Non-IID} & {FedAvg} & 82.73   $\pm$ 2.74 &	42.51  $\pm$ 3.11 & 72.90   $\pm$ 2.58 &	74.84  $\pm$ 2.99  &	76.50  $\pm$ 2.73 \\
& {FedMeta} & 82.58  $\pm$ 2.38  &	34.81  $\pm$ 3.00 & 71.67   $\pm$ 2.59 &	74.39  $\pm$ 2.41 &	75.85  $\pm$ 2.51\\
& {q-FFL} & 83.39  $\pm$ 2.66 &	38.07  $\pm$ 2.95 & 72.60  $\pm$ 2.98 &	74.97  $\pm$ 2.73 &	76.57  $\pm$ 2.87 \\
& \systemnott ($\gamma$ = 0.75)  & \textbf{84.25}   $\pm$ 2.35 & \textbf{30.11}  $\pm$ 2.47 & \textbf{76.18}  $\pm$ 2.31 &	\textbf{77.54}  $\pm$ 2.60 &	\textbf{78.73}  $\pm$ 2.47\\
\hline

\end{tabular}
}
\caption{Average performance across 10 \textbf{non-IID} downstream FL tasks, initialized with various FL pre-trained methods using \textbf{30} out of 100 participants in \textbf{scenario I}, on the \textbf{Tiny-ImageNet} dataset.}
% \vspace{-1.5pc}
\label{tb:tiny_30_noniidfinetune}
\end{center}
\end{table*}

\begin{table*}
\small
\begin{center}
\scalebox{0.9}{
\begin{tabular}{c||cccccc}
\hline
{\textbf{Pre-training} (Scenario I, $|m| = 20$)} & \multicolumn{5}{c}{\textbf{Downstream:} Non-IID FedAvg}\\
\hline\hline
 {\textbf{Method}} & \textbf{Acc} $\uparrow$ & \textbf{Variance} $\downarrow$ & \textbf{Lowest 10\%} $\uparrow$& \textbf{Lowest 20\%} $\uparrow$&\textbf{Lowest 30\%} $\uparrow$\\
\hline

 {FedAvg} & 81.35	 & 59.69	 & 70.35  &	70.91	  & 71.63\\
 {FedMeta} & 82.69 & 46.38 & 72.74 & 73.51 & 74.39\\
 {q-FFL} & 83.27 & 	51.48 & 	73.01	 & 73.82	 & 75.02\\
 \systemnott ($\gamma$ = 0.5) & \textbf{84.71} &	 \textbf{39.26} &	 \textbf{73.28} &	 \textbf{74.99} &	 \textbf{75.35}\\
\hline

\end{tabular}
}
\caption{Performance of 20-way classification downstream FedAvg, initialized with various non-IID FL pre-trained methods in \textbf{scenario I}, on the \textbf{CIFAR-100} dataset.}
% \vspace{-1.5pc}
\label{tb:20way}
\end{center}
\end{table*}

\begin{table*}
\begin{center}

\scalebox{0.85}{

\begin{tabular}{c||cc||cc||cc||cc}
\hline
\multirow{2}{*}{\begin{tabular}[c]{@{}c@{}} {\textbf{Pre-training}} \\ {(PACS)}\end{tabular}} & \multicolumn{8}{c}{\textbf{Downstream}: Non-IID FedAvg} \\
\cline{2-9}
 & \multicolumn{2}{c||}{Sketch}& \multicolumn{2}{c||}{Art}& \multicolumn{2}{c||}{Cartoon}& \multicolumn{2}{c}{Photo}\\
\hline\hline
 {\textbf{Method}} & \textbf{Acc} $\uparrow$ & \textbf{Variance} $\downarrow$ & \textbf{Acc} $\uparrow$ & \textbf{Variance} $\downarrow$ & \textbf{Acc} $\uparrow$ & \textbf{Variance} $\downarrow$ & \textbf{Acc} $\uparrow$ & \textbf{Variance} $\downarrow$ \\
\hline

FedAvg & 62.23 $\pm$ 2.65  & 51.29 $\pm$ 3.11  &70.99 $\pm$  3.05 &	49.32  $\pm$  3.33 & 64.39 $\pm$  2.96 &	69.31 $\pm$ 2.84 & 77.24 $\pm$ 2.75  & 61.35  $\pm$ 3.06  \\
FedMeta  & 64.35 $\pm$ 3.07  & 44.38  $\pm$ 2.98  &73.26  $\pm$ 2.64 &	44.61  $\pm$  3.00 &	62.17 $\pm$ 3.16  &	47.95  $\pm$ 3.08 & 76.59 $\pm$ 2.37  & 70.38  $\pm$ 3.11  \\
q-FFL	& 60.79	 $\pm$ 3.15 & 27.96  $\pm$ 3.03 &73.79 $\pm$ 2.99 	& 30.11 $\pm$ 2.95  & 	66.94 $\pm$  2.63	& 53.22 $\pm$ 2.99 &  80.63  $\pm$ 3.00 & 54.44  $\pm$ 2.98 \\
\systemnott & \textbf{66.83} $\pm$  2.85	& \textbf{24.31} $\pm$ 2.83  & \textbf{75.19} $\pm$ 2.71	& \textbf{26.36}	 $\pm$ 3.02 & \textbf{68.33} $\pm$ 2.51 	&\textbf{39.26} $\pm$ 3.13  & \textbf{82.19}  $\pm$ 2.80 & \textbf{46.32}  $\pm$ 3.05 \\
\hline

\end{tabular}
}

\caption{The results of domain shifts scenario using the PACS dataset.}
\label{tb:PACS}
\end{center}
 \end{table*} 
%%%

% \input{Tab/C100_sc2_pre10_downiid}
\begin{table*}
\small
\begin{center}
\scalebox{0.9}{

\begin{tabular}{cc||cccccc}
\hline
\multicolumn{2}{c||}{\textbf{Pre-training} (Scenario II, $|m| = 15$)} & \multicolumn{5}{c}{\textbf{Downstream:} IID FedAvg}\\
\hline\hline
 \textbf{Distribution} & {\textbf{Method}} & \textbf{Acc} $\uparrow$ & \textbf{Variance} $\downarrow$ & \textbf{Lowest 10\%} $\uparrow$& \textbf{Lowest 20\%} $\uparrow$&\textbf{Lowest 30\%} $\uparrow$\\
\hline

\multirow{5}{*}{IID} & {FedAvg} & 87.02   $\pm$ 1.17 &	12.82   $\pm$ 1.35  &		82.42   $\pm$ 1.44 &	83.33   $\pm$ 1.30 &	84.04   $\pm$ 1.26\\
& {FedMeta} & 87.07   $\pm$ 1.28 &	10.76   $\pm$ 1.95  &	81.94   $\pm$ 1.36 &	82.73  $\pm$ 1.20  &	83.43   $\pm$ 1.49\\
& {q-FFL} & 87.27  $\pm$ 1.33 &	13.69   $\pm$  1.61 &	81.21   $\pm$ 1.29 &	82.30   $\pm$ 1.37 &	82.95   $\pm$ 1.45\\
& \systemnott-SGD ($\gamma$ = 0.25) & 87.87  $\pm$ 1.31 &	13.32  $\pm$  1.11 &		82.73  $\pm$ 1.35 &	83.58  $\pm$ 1.16 &	84.06   $\pm$ 1.19\\
& \systemnott ($\gamma$ = 0.25) & \textbf{88.58}   $\pm$ 1.19 &	\textbf{8.70}  $\pm$ 1.15 &	\textbf{83.39}  $\pm$ 1.23 &	\textbf{83.88}   $\pm$ 1.25&	\textbf{84.69}  $\pm$ 1.29\\
\hline

\multirow{5}{*}{Non-IID} & {FedAvg} & 86.22   $\pm$ 1.28 &	15.44   $\pm$ 1.55 &	79.15  $\pm$ 1.41  &	80.36  $\pm$ 1.39 &	81.54  $\pm$ 1.46 \\
& {FedMeta} & 86.09   $\pm$ 1.39 &	11.42   $\pm$ 1.57 &		80.12  $\pm$ 1.44 &	81.03   $\pm$ 1.58 &	81.98  $\pm$ 1.37 \\
& {q-FFL} & 86.56   $\pm$ 1.33&	15.29   $\pm$ 1.94&	78.42  $\pm$ 1.61 &	80.00   $\pm$ 1.52 & 81.37   $\pm$ 1.39 \\
& \systemnott-SGD ($\gamma$ = 0.75) & 86.73   $\pm$ 1.17&	12.46  $\pm$  1.39 &	80.61  $\pm$ 1.08 &	81.76  $\pm$ 1.23  &	82.63  $\pm$ 1.19\\
& \systemnott ($\gamma$ = 0.75) & \textbf{87.42}   $\pm$ 1.23&	\textbf{9.06}   $\pm$ 1.46 &		\textbf{81.21}  $\pm$ 1.11 &	\textbf{82.30}  $\pm$ 1.25 &	\textbf{82.95}   $\pm$ 1.27\\
\hline

\end{tabular}
}
\caption{Average performance across 10 \textbf{IID} downstream FL tasks, initialized with various FL pre-trained methods using \textbf{15} out of 100 participants in \textbf{scenario II}, on the \textbf{CIFAR-100} dataset.}
% \vspace{-1.5pc}
\label{tb:cfiar_15_iidfinetune_sc2}
\end{center}
\end{table*}
\begin{table*}
\small
\begin{center}
\scalebox{0.9}{

\begin{tabular}{cc||cccccc}
\hline
\multicolumn{2}{c||}{\textbf{Pre-training} (Scenario II, $|m| = 20$)} & \multicolumn{5}{c}{\textbf{Downstream:} IID FedAvg}\\
\hline\hline
 \textbf{Distribution} & {\textbf{Method}} & \textbf{Acc} $\uparrow$ & \textbf{Variance} $\downarrow$ & \textbf{Lowest 10\%} $\uparrow$& \textbf{Lowest 20\%} $\uparrow$&\textbf{Lowest 30\%} $\uparrow$\\
\hline

\multirow{5}{*}{IID} & {FedAvg} & 87.28   $\pm$ 1.33 &	15.21   $\pm$ 2.06 &		80.00   $\pm$ 1.49&	81.21   $\pm$ 1.58 &	82.10   $\pm$ 1.37\\
& {FedMeta} &  87.27   $\pm$ 1.46 &	12.46   $\pm$ 1.77 &		81.45   $\pm$ 1.38 &	82.12   $\pm$ 1.63 &	82.95  $\pm$ 1.52\\
& {q-FFL} & 86.84   $\pm$ 1.29 &	12.74   $\pm$ 1.86 &	80.73   $\pm$ 1.57 &	82.12   $\pm$ 1.34 &	82.87  $\pm$ 1.39\\
& \systemnott-SGD ($\gamma$ = 0.5) & 87.67  $\pm$ 1.27 &	12.32   $\pm$1.49 &	81.82   $\pm$ 1.30 &	83.09   $\pm$ 1.38 &	83.92   $\pm$ 1.40\\
& \systemnott ($\gamma$ = 0.5) & \textbf{88.10}	  $\pm$ 1.30 & \textbf{9.30}   $\pm$ 1.35 &		\textbf{83.52}   $\pm$ 1.42 &	\textbf{84.30}   $\pm$ 1.33 &	\textbf{85.05}  $\pm$ 1.43\\
\hline

\multirow{5}{*}{Non-IID} & {FedAvg} & 86.39  $\pm$ 1.28 &	17.31   $\pm$ 1.64 &	79.64  $\pm$ 1.77 &	80.79   $\pm$ 1.65 &	81.78  $\pm$ 1.52 \\
& {FedMeta} & 86.32   $\pm$ 1.60 &	12.46  $\pm$ 2.00 &	80.61   $\pm$ 1.83&	81.45  $\pm$ 1.49 &	82.26  $\pm$ 1.77 \\
& {q-FFL} & 86.17  $\pm$ 1.93 & 16.24   $\pm$ 2.31 &	79.27   $\pm$ 2.07 &	81.09   $\pm$ 1.89 &	82.02   $\pm$ 1.95 \\
& \systemnott-SGD ($\gamma$ = 0.25) & 86.63   $\pm$ 1.54 & 11.76  $\pm$ 1.66 &		81.21  $\pm$ 1.65  &	82.00   $\pm$ 1.73 &	82.46   $\pm$ 1.88\\
& \systemnott ($\gamma$ = 0.25) & \textbf{87.02}   $\pm$ 1.39 &	\textbf{10.50}   $\pm$ 1.47 &		\textbf{81.70}   $\pm$ 1.73 &	\textbf{82.42}   $\pm$ 1.52 &	\textbf{83.23}  $\pm$ 1.66\\
\hline

\end{tabular}
}
\caption{Average performance across 10 \textbf{IID} downstream FL tasks, initialized with various FL pre-trained methods using \textbf{20} out of 100 participants in \textbf{scenario II}, on the \textbf{CIFAR-100} dataset.}
% \vspace{-1.5pc}
\label{tb:cfiar_20_iidfinetune_sc2}
\end{center}
\end{table*}
\begin{table*}
\small
\begin{center}
\scalebox{0.9}{

\begin{tabular}{cc||cccccc}
\hline
\multicolumn{2}{c||}{\textbf{Pre-training} (Scenario II, $|m| = 25$)} & \multicolumn{5}{c}{\textbf{Downstream:} IID FedAvg}\\
\hline\hline
 \textbf{Distribution} & {\textbf{Method}} & \textbf{Acc} $\uparrow$ & \textbf{Variance} $\downarrow$ & \textbf{Lowest 10\%} $\uparrow$& \textbf{Lowest 20\%} $\uparrow$&\textbf{Lowest 30\%} $\uparrow$\\
\hline

\multirow{5}{*}{IID} & {FedAvg} &  87.31   $\pm$ 1.09 &	14.90   $\pm$ 1.93 & 79.39   $\pm$ 1.17&	81.64   $\pm$ 1.25 &	82.75  $\pm$ 1.22\\
& {FedMeta} &  86.81   $\pm$ 0.94 &	10.89  $\pm$ 2.08 & 80.97  $\pm$ 1.35 &	82.30   $\pm$ 0.89 &	83.07  $\pm$ 1.15 \\
& {q-FFL} &  87.36   $\pm$ 1.32 &	17.47   $\pm$ 1.99 & 80.73   $\pm$ 1.45 &	81.76   $\pm$ 1.39 &	82.87  $\pm$ 1.17 \\
& \systemnott-SGD ($\gamma$ = 0.5) &  87.98   $\pm$ 1.00 &	11.22   $\pm$ 1.45 & 82.55   $\pm$ 1.28 &	83.15   $\pm$ 1.10 &	83.92  $\pm$ 1.04 \\
& \systemnott ($\gamma$ = 0.5) & \textbf{88.67}   $\pm$ 1.03 &	\textbf{9.98}   $\pm$ 1.35 & \textbf{83.88}  $\pm$ 1.22 &	\textbf{84.55}   $\pm$ 1.03 &	\textbf{85.29}  $\pm$ 0.95\\
\hline

\multirow{5}{*}{Non-IID} & {FedAvg} & 86.37   $\pm$ 1.23 &	15.44   $\pm$ 1.54 & 78.79  $\pm$ 1.37 &	80.12   $\pm$ 1.12 &	81.25  $\pm$ 1.00\\
& {FedMeta} & 85.49   $\pm$ 1.08 &	16.89  $\pm$ 1.25 & 79.27   $\pm$ 1.14 &	80.36  $\pm$ 1.03 & 81.25   $\pm$ 1.11 \\
& {q-FFL} &  85.67  $\pm$ 1.39 &	17.06   $\pm$ 1.08 & 80.61   $\pm$ 1.53 & 81.45   $\pm$ 1.47 &	81.98  $\pm$ 1.62 \\
& \systemnott-SGD ($\gamma$ = 0.75) & 86.40  $\pm$ 1.05 &	13.10  $\pm$ 1.30 & 80.62  $\pm$ 1.23 &	81.45  $\pm$ 1.10 &	82.34   $\pm$ 1.18 \\
& \systemnott ($\gamma$ = 0.75) & \textbf{87.32}  $\pm$  1.08 &	\textbf{11.22}  $\pm$ 1.00 & \textbf{82.42}  $\pm$ 1.35 & \textbf{83.27}   $\pm$ 1.18&	\textbf{83.84}  $\pm$ 1.23 \\
\hline

\end{tabular}
}
\caption{Average performance across 10 \textbf{IID} downstream FL tasks, initialized with various FL pre-trained methods using \textbf{25} out of 100 participants in \textbf{scenario II}, on the \textbf{CIFAR-100} dataset.}
% \vspace{-1.5pc}
\label{tb:cfiar_25_iidfinetune_sc2}
\end{center}
\end{table*}
\begin{table*}
\small
\begin{center}
\scalebox{0.9}{
\begin{tabular}{cc||cccccc}
\hline
\multicolumn{2}{c||}{\textbf{Pre-training} (Scenario II, $|m| = 30$)} & \multicolumn{5}{c}{\textbf{Downstream:} IID FedAvg}\\
\hline\hline
 \textbf{Distribution} & {\textbf{Method}} & \textbf{Acc} $\uparrow$ & \textbf{Variance} $\downarrow$ & \textbf{Lowest 10\%} $\uparrow$& \textbf{Lowest 20\%} $\uparrow$&\textbf{Lowest 30\%} $\uparrow$\\
\hline

\multirow{5}{*}{IID} & {FedAvg} & 87.51  $\pm$ 0.95  &	13.76  $\pm$ 2.03 & 82.55  $\pm$ 1.14  &	83.58  $\pm$ 1.25 &	84.20   $\pm$ 1.08 \\
& {FedMeta} & 87.25   $\pm$ 1.20 &	12.39  $\pm$ 1.94 & 81.70   $\pm$ 1.33 &	82.55   $\pm$ 1.45 &	83.03  $\pm$ 1.11 \\
& {q-FFL} & 86.78   $\pm$ 1.29 &	13.76  $\pm$ 1.53  & 81.21   $\pm$ 1.47 &	82.06  $\pm$ 1.35 &	82.79   $\pm$ 1.53\\
& \systemnott-SGD ($\gamma$ = 0.75) & 87.75  $\pm$ 1.17 &	13.40  $\pm$ 1.64 & 81.52  $\pm$ 1.25 &	82.94   $\pm$ 1.08 &	83.70   $\pm$ 1.27 \\
& \systemnott ($\gamma$ = 0.75) &  \textbf{88.27}  $\pm$  1.04 &	 \textbf{9.06}  $\pm$ 1.66  &  \textbf{84.06}  $\pm$ 1.25 &	 \textbf{84.55}   $\pm$ 1.17 &	 \textbf{85.05}  $\pm$ 1.08 \\
\hline

\multirow{5}{*}{Non-IID} & {FedAvg} & 86.07  $\pm$ 1.33 &	11.09  $\pm$ 1.94 & 80.61  $\pm$ 1.58 &	81.64   $\pm$ 1.49 &	82.38  $\pm$ 1.76 \\
& {FedMeta} & 86.25  $\pm$ 1.07 &	12.96   $\pm$ 2.15 & 79.03   $\pm$ 1.35&	80.36   $\pm$ 1.24 &	81.66  $\pm$ 1.19 \\
& {q-FFL} & 85.50  $\pm$ 1.25 &	15.29   $\pm$ 2.07 & 77.58  $\pm$ 1.54 &	79.39   $\pm$ 1.39 &	80.57  $\pm$ 1.33 \\
& \systemnott-SGD ($\gamma$ = 0.5)  & 86.47   $\pm$ 1.16 &	 \textbf{10.96}  $\pm$ 1.85 & 80.36  $\pm$ 1.37 &	81.36   $\pm$ 1.40 & 82.00  $\pm$ 1.29\\
& \systemnott ($\gamma$ = 0.5)  &  \textbf{87.54}  $\pm$ 1.14 &	 \textbf{10.96}   $\pm$ 1.80 &  \textbf{81.09}  $\pm$ 1.10 &	 \textbf{81.67}   $\pm$ 1.26 &	 \textbf{82.40}  $\pm$ 1.33\\

\hline
\end{tabular}
}
\caption{Average performance across 10 \textbf{IID} downstream FL tasks, initialized with various FL pre-trained methods using \textbf{30} out of 100 participants in \textbf{scenario II}, on the \textbf{CIFAR-100} dataset.}
% \vspace{-1.5pc}
\label{tb:cfiar_30_iidfinetune_sc2}
\end{center}
\end{table*}

\begin{table*}
\small
\begin{center}
\scalebox{0.9}{
\begin{tabular}{cc||cccccc}
\hline
\multicolumn{2}{c||}{\textbf{Pre-training} (Scenario II, $|m| = 15$)} & \multicolumn{5}{c}{\textbf{Downstream:} Non-IID FedAvg}\\
\hline\hline
 \textbf{Distribution} & {\textbf{Method}} & \textbf{Acc} $\uparrow$ & \textbf{Variance} $\downarrow$ & \textbf{Lowest 10\%} $\uparrow$& \textbf{Lowest 20\%} $\uparrow$&\textbf{Lowest 30\%} $\uparrow$\\
\hline
\multirow{5}{*}{IID} & {FedAvg} & 84.02 $\pm$ 2.73 &	46.79  $\pm$ 3.61 &		71.01  $\pm$ 2.95 &	74.53  $\pm$ 3.08&	76.81  $\pm$ 3.21 \\
& {FedMeta} & 83.47  $\pm$ 3.18 &	34.11  $\pm$ 3.29 &		73.68 $\pm$  3.44 &	75.20  $\pm$ 3.71 &	76.35  $\pm$ 3.25 \\
& {q-FFL} & 85.03  $\pm$ 2.95 &	35.64  $\pm$ 2.79 &		74.04  $\pm$ 3.07&	76.39  $\pm$ 3.11 &	78.12  $\pm$ 2.94 \\
& \systemnott-SGD ($\gamma$ = 0.75) & 85.04  $\pm$ 2.80 &		35.64  $\pm$ 2.45 &		\textbf{74.61}  $\pm$ 2.88& 76.40  $\pm$ 2.98 &	78.34  $\pm$ 3.01 \\
& \systemnott ($\gamma$ = 0.75) & \textbf{85.08}  $\pm$ 2.62 &	\textbf{31.70}  $\pm$ 2.45  &		\textbf{74.61}  $\pm$ 2.76 & \textbf{76.87}  $\pm$ 3.00 &	\textbf{78.63} $\pm$ 3.04\\
\hline
\multirow{5}{*}{Non-IID} & {FedAvg} & 82.91  $\pm$ 3.25 &	41.99  $\pm$ 3.68 &	71.90  $\pm$ 3.51 &	75.23  $\pm$ 3.73 &	76.81 $\pm$ 3.49\\
& {FedMeta} & 78.77  $\pm$ 2.99 & 70.39  $\pm$ 3.37 &		65.13  $\pm$ 3.41 &	67.47 $\pm$ 3.29  &	69.28 $\pm$ 2.95 \\
& {q-FFL} & 80.94 $\pm$ 3.07  &	49.42  $\pm$ 2.84 &		69.57  $\pm$ 3.39& 71.46  $\pm$ 2.95&	72.86 $\pm$ 3.37\\
& \systemnott-SGD ($\gamma$ = 0.25) & 83.42 $\pm$ 3.03 &40.20  $\pm$ 2.97 &		73.09  $\pm$ 3.19 &	74.54  $\pm$ 3.07&	76.29 $\pm$ 2.88\\
& \systemnott ($\gamma$ = 0.25) & \textbf{83.83} $\pm$  3.00 &	\textbf{39.31}  $\pm$  3.04&		\textbf{74.26}  $\pm$ 2.92 &	\textbf{76.42}  $\pm$ 3.10 & \textbf{78.10} $\pm$ 3.04\\
\hline
\end{tabular}
}
\caption{Average performance across 10 \textbf{non-IID} downstream FL tasks, initialized with various FL pre-trained methods using \textbf{15} out of 100 participants in \textbf{scenario II}, on the \textbf{CIFAR-100} dataset.}
% \vspace{-1.5pc}
\label{tb:cfiar_15_noniidfinetune_sc2}
\end{center}
\end{table*}
\begin{table*}
\small
\begin{center}

\scalebox{0.9}{
\begin{tabular}{cc||cccccc}
\hline
\multicolumn{2}{c||}{\textbf{Pre-training} (Scenario II, $|m| = 20$)} & \multicolumn{5}{c}{\textbf{Downstream:} Non-IID FedAvg}\\
\hline\hline
 \textbf{Distribution} & {\textbf{Method}} & \textbf{Acc} $\uparrow$ & \textbf{Variance} $\downarrow$ & \textbf{Lowest 10\%} $\uparrow$& \textbf{Lowest 20\%} $\uparrow$&\textbf{Lowest 30\%} $\uparrow$\\
\hline

\multirow{5}{*}{IID} & {FedAvg} & 81.79 $\pm$ 3.55  &	41.73  $\pm$ 2.89  &		69.84  $\pm$ 3.19 & 73.47  $\pm$ 3.24&	75.11 $\pm$ 3.88 \\
& {FedMeta} & 82.29 $\pm$ 3.19 &	47.75  $\pm$  2.74 &		71.69  $\pm$ 3.07 &	74.17  $\pm$ 2.99&	75.71 $\pm$ 2.81\\
& {q-FFL} &  82.40  $\pm$ 2.99 &	40.32  $\pm$ 3.15 &		73.96  $\pm$  3.08 &	75.30  $\pm$ 2.87 &	76.59 $\pm$ 3.12\\
& \systemnott-SGD ($\gamma$ = 0.75) & 82.90  $\pm$ 2.79 & 38.94  $\pm$ 2.35 &		73.02  $\pm$ 2.89 &	75.60 $\pm$ 3.05 & 77.18 $\pm$ 2.94\\
& \systemnott ($\gamma$ = 0.75) & \textbf{85.68}  $\pm$ 2.51 &	\textbf{27.14}  $\pm$ 2.43 &		\textbf{75.36} $\pm$ 2.66 & \textbf{77.25}  $\pm$ 2.97 &	\textbf{78.49} $\pm$ 2.69\\
\hline

\multirow{5}{*}{Non-IID} & {FedAvg} & 82.82  $\pm$ 3.17&	49.00  $\pm$  3.41&		69.71  $\pm$ 3.25 &	72.54  $\pm$ 3.30 & 74.58 $\pm$ 3.21\\
& {FedMeta} & 82.69 $\pm$ 3.05 & 	48.44  $\pm$ 2.99 &		68.84  $\pm$ 3.14 &	71.82  $\pm$ 3.27&	74.14 $\pm$ 3.09 \\
& {q-FFL} & 82.14 $\pm$ 2.76 &73.10  $\pm$ 3.08 &		68.22  $\pm$ 3.00 &	70.64  $\pm$ 2.85 &	73.77 $\pm$ 3.14\\
& \systemnott-SGD ($\gamma$ = 0.25) & 83.63 $\pm$ 3.00  &	41.73 $\pm$ 2.85 &		69.76  $\pm$ 2.94 & 73.46  $\pm$ 3.09 &	75.64  $\pm$ 3.15 \\
& \systemnott ($\gamma$ = 0.25) & \textbf{86.63}  $\pm$ 2.93 &	\textbf{31.58}  $\pm$ 2.64 &	\textbf{73.05}  $\pm$ 2.51 &	\textbf{75.82}  $\pm$ 2.88 & \textbf{77.58} $\pm$ 2.99 \\

\hline
\end{tabular}
}
\caption{Average performance across 10 \textbf{non-IID} downstream FL tasks, initialized with various FL pre-trained methods using \textbf{20} out of 100 participants in \textbf{scenario II}, on the \textbf{CIFAR-100} dataset.}
% \vspace{-1.5pc}
\label{tb:cfiar_20_noniidfinetune_sc2}
\end{center}
\end{table*}
\begin{table*}
\small
\begin{center}
\scalebox{0.9}{

\begin{tabular}{cc||cccccc}
\hline
\multicolumn{2}{c||}{\textbf{Pre-training} (Scenario II, $|m| = 25$)} & \multicolumn{5}{c}{\textbf{Downstream:} Non-IID FedAvg}\\
\hline\hline
 \textbf{Distribution} & {\textbf{Method}} & \textbf{Acc} $\uparrow$ & \textbf{Variance} $\downarrow$ & \textbf{Lowest 10\%} $\uparrow$& \textbf{Lowest 20\%} $\uparrow$&\textbf{Lowest 30\%} $\uparrow$\\
\hline

\multirow{5}{*}{IID} & {FedAvg} & 80.53  $\pm$ 2.95 &	62.57  $\pm$ 3.11 & 66.51  $\pm$  3.05 &	68.54  $\pm$ 3.19 & 70.78 $\pm$ 2.98 \\
& {FedMeta} & 82.37  $\pm$ 3.00 &	45.97  $\pm$ 2.54 & 70.68  $\pm$ 2.98 &	73.40  $\pm$ 3.01 &	75.21  $\pm$ 3.12 \\
& {q-FFL} &  82.06 $\pm$ 2.77 &	48.44  $\pm$ 3.05 & 71.08 $\pm$ 3.16 &	73.03  $\pm$ 2.85 &	74.71 $\pm$ 3.01 \\
& \systemnott-SGD ($\gamma$ = 0.25) & 82.62  $\pm$ 2.43 &	75.86 $\pm$ 2.95 & 68.12  $\pm$ 2.83 & 70.73  $\pm$ 2.64 &	72.51 $\pm$ 2.51\\
& \systemnott ($\gamma$ = 0.25) & \textbf{85.05} $\pm$ 2.61  &	\textbf{33.99} $\pm$ 2.98 & \textbf{75.12}  $\pm$ 2.79 &	\textbf{76.74}  $\pm$ 2.90 &	\textbf{77.79}  $\pm$ 2.64 \\
\hline

\multirow{5}{*}{Non-IID} & {FedAvg} & 84.06  $\pm$ 3.11 &	40.07  $\pm$ 2.95 & 71.11  $\pm$ 3.44 & 73.36 $\pm$ 3.29 &	75.67 $\pm$ 3.36\\
& {FedMeta} & 81.40 $\pm$ 3.08  &	47.33 $\pm$ 3.45 & 67.41 $\pm$ 3.19 &	70.87  $\pm$ 2.88 &	72.49 $\pm$ 3.17\\
& {q-FFL} & 82.30 $\pm$ 2.84  &55.06 $\pm$ 3.33  & 67.82 $\pm$ 3.15 &	71.53  $\pm$ 3.29 &	73.70 $\pm$ 2.95\\
& \systemnott-SGD ($\gamma$ = 0.5) & 84.25  $\pm$ 3.01 &	53.88  $\pm$ 2.38 & 71.62  $\pm$ 3.11 &	73.48  $\pm$ 3.17 &	75.92 $\pm$ 3.09\\
& \systemnott ($\gamma$ = 0.5) & \textbf{84.92}  $\pm$ 2.88 &	\textbf{39.82}  $\pm$ 2.54 & \textbf{75.04}  $\pm$ 3.09 &	\textbf{77.45}  $\pm$ 2.94 &	\textbf{78.93} $\pm$ 3.04\\

\hline

\end{tabular}
}
\caption{Average performance across 10 \textbf{non-IID} downstream FL tasks, initialized with various FL pre-trained methods using \textbf{25} out of 100 participants in \textbf{scenario II}, on the \textbf{CIFAR-100} dataset.}
% \vspace{-1.5pc}
\label{tb:cfiar_25_noniidfinetune_sc2}
\end{center}
\end{table*}
\input{Tab/C100_sc2_pre30_downnoniid}

\input{Tab/Tiny_sc2_pre15_downiid}
\input{Tab/Tiny_sc2_pre20_downiid}
\input{Tab/Tiny_sc2_pre25_downiid}
\input{Tab/Tiny_sc2_pre30_downiid}

\input{Tab/Tiny_sc2_pre15_downnoniid}
\input{Tab/Tiny_sc2_pre20_downnoniid}
\input{Tab/Tiny_sc2_pre25_downnoniid}
\input{Tab/Tiny_sc2_pre30_downnoniid}

\input{Tab/rebuttal_sever+clientdata}

% \begin{wraptable}{r}{0.5\textwidth}

\begin{table*}
\begin{center}
    
\scalebox{0.8}{

\begin{tabular}{c||cc}
\hline
\multicolumn{1}{c||}{\textbf{Pre-training}} & \multicolumn{2}{c}{\textbf{Downstream:} Non-IID FedAvg}\\
\hline\hline
{\textbf{Method}} & \textbf{Acc} $\uparrow$ & \textbf{Variance} $\downarrow$ \\
\hline

 \systemnott (default: 5\%) & 86.63 & 31.58  \\
 \systemnott (2\%)  & 84.18 & 37.97\\
  \systemnott (1\%)  & 83.69 & 34.61\\

\hline

\end{tabular}}
 % \vspace{-2mm}
\caption{The results of varying the amount of the server's data $|D^S|$ for our method using the CIFAR-100 dataset.}
\label{tb:amount_of_server}
\end{center}
 \end{table*}
  % \end{wraptable}

\input{Tab/FEMNIST_sc1_pre20_downiid}
\input{Tab/FEMNIST_sc1_pre20_downnoniid}
\input{Tab/FEMNIST_sc2_pre20_downiid}
\input{Tab/FEMNIST_sc2_pre20_downnoniid}

\input{Tab/other_pre-train_CIFAR100_downIID}
\input{Tab/other_pre-train_TinyIMG_downIID}

\input{Tab/rebuttal_down_FedProx}

\clearpage

\input{Tab/rebuttal_down_qFFL}

\begin{table*}[t]
    \centering
    \begin{subtable}{\textwidth}
        \centering
        \scalebox{0.9}{
        \begin{tabular}{c||cc||cc}
        \hline
\multirow{2}{*}{\begin{tabular}[c]{@{}c@{}} {\textbf{Pre-training}} \\ {(ImageNet)}\end{tabular}} & \multicolumn{4}{c}{\textbf{Dataset}: CIFAR-100}\\
\cline{2-5}
 & \multicolumn{2}{c||}{\textbf{Downstream}: IID FedAvg}& \multicolumn{2}{c}{\textbf{Downstream}: Non-IID FedAvg}\\
        \hline
 {\textbf{Method}} & \textbf{Acc} $\uparrow$ & \textbf{Variance} $\downarrow$ & \textbf{Acc} $\uparrow$ & \textbf{Variance} $\downarrow$\\
        \hline
 {Centralized} & 87.91  $\pm$  0.99 &	13.96   $\pm$ 2.35  & 86.75   $\pm$  2.89 & 67.34   $\pm$ 2.17 \\
\systemnott & \textbf{88.39}   $\pm$ 1.15  & \textbf{11.37}   $\pm$ 2.03  & \textbf{87.96}   $\pm$ 1.95 & \textbf{30.79}   $\pm$ 2.79 \\
\hline
     \end{tabular}}
        \caption{Results of downstream FL using CIFAR-100 dataset, initialized with a model pre-trained on ImageNet.}
    \end{subtable}
% \quad%  
\\
    \begin{subtable}{\textwidth}
        \centering
        \scalebox{0.9}{
        \begin{tabular}{c||cc||cc}
        \hline
\multirow{2}{*}{\begin{tabular}[c]{@{}c@{}} {\textbf{Pre-training}} \\ {(ImageNet)}\end{tabular}} & \multicolumn{4}{c}{\textbf{Dataset}: Tiny-ImageNet}\\
\cline{2-5}
 & \multicolumn{2}{c||}{\textbf{Downstream}: IID FedAvg}& \multicolumn{2}{c}{\textbf{Downstream}: Non-IID FedAvg}\\
        \hline
 {\textbf{Method}} & \textbf{Acc} $\uparrow$ & \textbf{Variance} $\downarrow$ & \textbf{Acc} $\uparrow$ & \textbf{Variance} $\downarrow$\\
        \hline
 {Centralized} &87.02   $\pm$ 1.33  & 	15.92   $\pm$ 2.54 & 85.58   $\pm$ 1.95 &	50.93  $\pm$ 2.06 \\
\systemnott & \textbf{88.94}   $\pm$  1.45 &	\textbf{13.21}  $\pm$ 1.98  & \textbf{86.79}  $\pm$ 1.37  & \textbf{31.44}   $\pm$ 3.11   \\
\hline
     \end{tabular}}
        \caption{Results of downstream FL using Tiny-ImageNet dataset, initialized with a model pre-trained on ImageNet.}
    \end{subtable}
    \caption{Results with pre-training on a centrally stored public dataset. ImageNet is used for pre-training, while CIFAR-100 and Tiny-ImageNet are used for downstream FL. \label{tb:imagenet_cifar_tiny}}
\end{table*}

\clearpage
\end{document}